\documentclass{article}

\usepackage[english]{babel}
\usepackage{natbib}
\usepackage[letterpaper,top=2cm,bottom=2cm,left=3cm,right=3cm,marginparwidth=1.75cm]{geometry}

\usepackage[utf8]{inputenc} 
\usepackage[T1]{fontenc}    
\usepackage{url}            
\usepackage{booktabs}       
\usepackage{amsfonts}       
\usepackage{nicefrac}       
\usepackage{microtype}      
\usepackage{xcolor}         
\usepackage{comment}
\usepackage{bm}
\newtheorem{remark}{Remark}

\usepackage{algorithm}
\usepackage{algpseudocode}

\usepackage{amsmath}
\usepackage{graphicx}
\usepackage[colorlinks=true, allcolors=blue]{hyperref}

\title{Structure-Informed Deep Reinforcement Learning for Inventory Management}
\author{Alvaro Maggiar\footnote{corresponding author}\\ \texttt{maggiara@amazon.com}
  \and Sohrab Andaz\\ \texttt{sandaz@amazon.com}
  \and Akhil Bagaria\\ \texttt{akhilbg@amazon.com}
  \and Carson Eisenach \\ \texttt{ceisen@amazon.com}
  \and Dean Foster \\ \texttt{foster@amazon.com}
  \and Omer Gottesman \\ \texttt{omergott@amazon.com}
  \and Dominique Perrault-Joncas \\ \texttt{joncas@amazon.com}
  }
  
\date{July 29, 2025}  
  
\begin{document}
\maketitle

\begin{abstract}
This paper investigates the application of Deep Reinforcement Learning (DRL) to classical inventory management problems, with a focus on practical implementation considerations. We apply a DRL algorithm based on DirectBackprop to several fundamental inventory management scenarios including multi-period systems with lost sales (with and without lead times), perishable inventory management, dual sourcing, and joint inventory procurement and removal. The DRL approach learns policies across products using only historical information that would be available in practice, avoiding unrealistic assumptions about demand distributions or access to distribution parameters. We demonstrate that our generic DRL implementation performs competitively against or outperforms established benchmarks and heuristics across these diverse settings, while requiring minimal parameter tuning. Through examination of the learned policies, we show that the DRL approach naturally captures many known structural properties of optimal policies derived from traditional operations research methods. To further improve policy performance and interpretability, we propose a Structure-Informed Policy Network technique that explicitly incorporates analytically-derived characteristics of optimal policies into the learning process. This approach can help interpretability and add robustness to the policy in out-of-sample performance, as we demonstrate in an example with realistic demand data. Finally, we provide an illustrative application of DRL in a non-stationary setting. Our work bridges the gap between data-driven learning and analytical insights in inventory management while maintaining practical applicability.
\end{abstract}
\section{Introduction}

The field of inventory management has long been a cornerstone of operations research and supply chain management. Its aim is chiefly to optimize the tradeoffs between carrying too much or too little inventory, and depending on the setting within which a problem is formulated, the difficulty of the task spans a wide range. From simple analytical solutions such as the seminal newsvendor problem \cite{Petruzzi2011-ws} for a single-period problem with stochastic demand, or the Economic Order Quantity (EOQ) for a constant deterministic demand \cite{harris1990many}, problems become increasingly intricate as the model grows in the features it considers, which can include stochastic demand, lead times, perishability, multiple locations, or multiple sourcing options, to name a few. Even in a relatively simple settings such as one with stationary demand, lost sales and lead times, the problem is notoriously difficult \cite{zipkin2008old}, and only properties of the optimal policy can be derived.

Traditional approaches have relied on analytical methods and heuristics to address the complexities of inventory control systems (\cite{zipkin2000foundations,snyder2019fundamentals}). However, the advent of machine learning, particularly reinforcement learning (RL), has opened new avenues for tackling these challenges.  Inventory management systems, especially those involving multiple locations, product types, and complex constraints, present significant difficulties due to their high-dimensional state and action spaces, stochastic demand patterns, and intricate cost structures. While classical methods have provided valuable insights and solutions for simpler scenarios, they often struggle with the complexity and scale of real-world inventory systems. Deep Reinforcement Learning (DRL) offers a promising alternative, capable of learning effective policies directly from data without relying on explicit modeling of system dynamics.

Recent work in this domain has demonstrated the potential of DRL in addressing practical inventory control problems. Notably, \cite{Madeka2021} introduced the DirectBackprop algorithm, which leverages a differentiable formulation of inventory systems to achieve state-of-the-art practical performance. Their approach, which handles lost sales, correlated demand across time,  and stochastic lead times,  represents a significant step forward in applying DRL to practical inventory management scenarios.

Traditional inventory management problems studied in the literature assume very idealized, though unrealistic, settings, such as stationary environments and backlogging of missed sales. They nonetheless represent cases on which practical DRL approaches can be tested and compared to known solutions or heuristics to help validate the methodology, as well as gain a deeper understanding thereof.  In recent years,  a growing body of literature has explored the use of DRL to inventory management in such idealized settings in order to establish the benefits of RL and overcome the limitations of traditional methods in practical applications (e.g. \cite{vanvuchelen2020use,oroojlooyjadid2020,gijsbrechts2022can,liu2024multi}). Nonetheless, these approaches suffer from a number of shortcomings when it comes to representing practical alternatives.

A common approach among those pieces of research is to apply an RL algorithm, and learn a policy, at an instance level.  In other words fixing, say, the economic parameters of a product, and rolling out different realized paths of demand for that one and only product.  In such an approach, the policy is learned at the instance level by leveraging information across a multitude of realized scenarios for a single product. This is contrary to what would be done in practice, where one would only have access to single realized scenarios for individual products, and leverage information across a multitude of them.  Additionally, many studies allow the policy to consume, as part of the state, information about the demand distribution at hand in the form of distribution parameters. In practice, we would only have access to historical realized demand and would have to implicitly predict future demand from historical information.  Finally, most of these endeavors make use of RL algorithms that require careful fine-tuning and hyperparameter optimization. In contrast, the DirectBackprop algorithm of \cite{Madeka2021}, which leverages an exo-MDP \citep{sinclair2023hindsight} formulation, allows for the reduction of the problem to a supervised learning one where the only hyperparameter is essentially the learning rate.

This paper explores the application of  deep reinforcement learning to a range of classic inventory management problems with a concern towards practicality.  These inventory management problems can be formulated as exo-Markov Decision Processes, and we apply a practical DRL algorithm based on the DirectBackprop algorithm. Crucially, while we consider the idealized settings studied in the literature, we act from the agent perspective as if this information was not known. We assess the performance of the DRL algorithm with respect to benchmarks, consider whether the DRL algorithm is able to learn the structure of optimal policies, and finally propose a Structure Informed Neural Network to train policies in order to leverage analytically derived structural information of the optimal policies.  Our work aims to contribute to the literature and extend the application of DRL in inventory management in several  ways:

\begin{itemize}
	\item We apply DRL as it would be implemented in practice, mimicking state information that would be the one available to practitioners.  In particular, this implies learning policies across products, using only the historical information available for them.

    \item We apply the DirectBackprop algorithm to a broad range of classic inventory management problems, including multi-period systems with lost sales (both with and without lead times), perishable product management, dual sourcing, and joint inventory procurement and removal. This diverse approach allows us to demonstrate the versatility and effectiveness of a generic DRL algorithm across diverse scenarios.

    \item We compare the performance of the DRL approach against known optimal solutions or established heuristics in stationary settings. This benchmarking provides a solid evaluation of DRL's capabilities relative to traditional methods.

    \item We emphasize the robustness of the DirectBackprop approach by demonstrating strong performance with virtually no parameter tuning. This aspect is crucial for practical applications, where extensive fine-tuning may not be feasible.

    \item Through  examination of the learned policies, we show that our DRL approach captures the known structures of optimal policies derived from traditional operations research methods in order to bridge the gap between data-driven learning and analytical insights.

    \item We leverage the  above structural analysis to propose a technique that explicitly incorporates analytically-derived characteristics of optimal policies into the learning process. This ``targeted regularization'' approach not only improves policy performance but also enhances interpretability and generalization, and stabilizes training.
    
	\item We demonstrate on a small example with realistic demand the potential benefits of incorporating known structural information in the learning of a DRL policy. 
	
	\item We illustrate on this realistic data how an end-to-end DRL approach outperforms a ``predict-then-optimize'' one.
    
\end{itemize}

In what follows, we first give some background on the Exo-MDP framework used in our approach, as well as the use of DRL for inventory management in Section~\ref{sec:background}. We then present the formulation of the considered problems and the DRL implementation in Section~\ref{sec:formulation}, which we apply to a number of classical inventory management problems in Section~\ref{sec:experiments}. Based on the results and observations of these experiments, we propose in Section~\ref{sec:sinn} a Structured-Informed Neural Network to enhance the learned DRL policies by incorporating knowledge of the structural properties of the optimal policies. Finally, we illustrate in Section~\ref{sec:nonstationary} on realistic, non-stationary demand data how an end-to-end DRL approach that aims to directly predict the order quantity outperforms a more traditional ``predict-then-optimize'' one.

\section{Background}\label{sec:background}

\subsection{Reinforcement Learning for Inventory Management}

Inventory management tackles the problem of optimizing inventory levels in a supply chain. Typically, this entails trading off the costs of too high or too low inventory levels in the face of uncertainty and operational constraints. Too high an inventory level incurs higher holding costs, costs of capital, and potentially spoilage costs for perishable products, while too low an inventory level results in missed sales and loss of goodwill from the customers. The inventory management literature is vast and we refer to \cite{zipkin2000foundations,porteus2002foundations,snyder2019fundamentals} for broader references on the field.  

Inventory management problems often naturally lend themselves to formulations as Markov Decision Processes (MDP) \cite{puterman2014markov}, but from a practical perspective, the dimensions of their state space often render them intractable, running into the so-called ``curse of dimensionality''. As a result, much of the literature is concerned with circumventing the resulting difficulty by considering specialized formulations that admit optimal policies with simple structures, or for which heuristics can be derived.  Some of the classical inventory management problems involve multi-period inventory management and different variations thereof, including either lost sales or backlogging assumptions, lead times,  perishability, dual sourcing, or removals to name some of the more common. Often, these models are studied with an added assumption of stationarity.  In some of these settings, optimal policies can explicitly be derived, although in most cases, only structural information about the optimal policies can be derived, usually involving tools from convex analysis such as supermodularity, multimodularity or $L^\natural$-convexity (\cite{simchi2014logic,li2014multimodularity,chen2017convexity}). The resulting structural properties often involve monotonicity of the policy with respect to state variables. Typically, this would involve results such as buying fewer units as the inventory position increases.

These theoretical results are informative, and the resulting heuristics still the basis of many practical implementations. However, the growing intricacies and scale of modern supply chain networks, and the complex observed demand patterns, require solutions that can better adapt to real-world problems.  Reinforcement Learning offers an attractive alternative to traditional methods, especially because it can leverage and learn from the large amounts of data available in industrial settings.  RL, and especially DRL, provide with a means of directly learning policies from the data, limiting the assumptions made about the world.  

Reinforcement Learning has a long history \cite{sutton2018reinforcement}, but its combination with Deep Learning techniques, resulting in Deep Reinforcement Learning methods, has brought it to the fore in the past decade as a viable approach to solve high-dimensional Markov Decision Processes (MDP), notably thanks to its public successes in video and board games \cite{mnih2015human, silver2017mastering,vinyals2019grandmaster}. As a result, although there had been attempts to apply some forms of RL to inventory management problems, such as the SMART algorithm of \cite{das1999solving,giannoccaro2002inventory}, or even early  applications of neural networks to inventory management \cite{van1997neuro}, the advent of DRL has spurred the investigation of its applicability to inventory management, and more general logistics problems.

One of the first applications of deep neural networks to inventory management can be found in \cite{oroojlooyjadid2020} who combine forecasting and optimization steps in the single-period newsvendor problem to predict the optimal buying quantity. Using DRL, and more specifically a Proximal Policy Optimization (PPO) algorithm \citep{schulman2017proximal},  \cite{balaji2019orl} consider a stationary multi-period version of the newsvendor model with lost sales and fixed lead times, assuming uniformly distributed prices and costs, and Poisson-distributed demand, to show that DRL can beat a naive heuristic. \cite{oroojlooyjadid2022} apply a deep Q-network \citep{mnih2015human} to the so-called ``beer game'', an example of a serial supply chain network. \cite{gijsbrechts2022can} apply the A3C algorithm \citep{a3c} to multi-period inventory management problems with lost sales and lead times, dual sourcing, and multi-echelon problems and show that it performs on par with common heuristics on instances of those problems. \cite{de2022reward} apply a deep Q-network with reward shaping to instances of an inventory management problem with perishable goods. \cite{geevers2024multi} apply the PPO algorithm to instances of multi-echelon networks with backlogged demand. \cite{kaynov2024deep} also apply the PPO algorithm to instances of a distribution network with one warehouse serving multiple retailers. \cite{stranieri2023comparing} compare the performance of three different DRL algorithms, A3C, PPO, and the Vanilla Policy Gradient (VPG also known as REINFORCE) \citep{williams1992simple} on instances of a non-stationary two-echelon network. 

While these approaches implement model-free DRL algorithms such as PPO, A3C, deep Q-networks, or VPG, an other stream of work in \cite{Madeka2021,alvo2023neural} leverage the differentiability of many inventory models and express their multi-period inventory management problems as Exo-MDPs. This allows for the inclusion of complex features, and extensions that include general arrival dynamics \cite{andaz2023learning}, or capacity control \cite{eisenach2024neural}.

Additionally, a shortcoming of most of the above applications of DRL to inventory management (with the exception of \cite{Madeka2021}) is that they usually learn policies either at the instance level, by rolling out many paths for the same product and learning a policy specifically for that product, or by allowing for different demand traces, but fixing the economics to be the same across products. In contrast, we take a more realistic approach where we have a multitude of products with varying economic costs, including holding costs which are usually dependent on the cubic volume of the products, and only one demand trace per product, whose generating process is unknown to the DRL agent.

\subsection{Exo-MDP Framework and Differentiable Models}

Many inventory management problems can be fit into the Input-Driven MDP framework \cite{mao2018variance}, in which the state of the system can be split into two main components: exogenous and endogenous inputs. The exogenous inputs evolve independently of the agent's actions. For example, the (uncensored) customer demand process can be considered as an exogenous process, whose values are independent of the inventory levels.  The evolution of endogenous factors on the other hand is impacted by the agent's decisions. In the context of inventory management, an example of an endogenous variable could be the inventory level, or the amount of returnable inventory. Decisions on the number of units to bring in, or remove, affect these factors directly. If we further assume that the only thing that is unknown about the exogenous inputs are their future distributions, the Input-Driven MDP becomes an Exo-MDP \cite{sinclair2023hindsight}.

The formulation of the problem as an Exo-MDP offers many benefits. In particular, it allows for the use of historical data to evaluate any policy, and through the reduction of the problem to a supervised learning one, the optimization of the agent's policy. Because the dynamics of the problem are differentiable, the (offline) rollout of a parametrized policy using the historical exogenous data readily yields gradients of the reward function with respect to the policy's parameters, which can then be used in an optimization scheme such as a variant of stochastic gradient ascent. This approach has proved successful in practice \cite{Madeka2021}, and in more theoretical work \cite{alvo2023neural}.

\subsection{Practical Data-Driven Systems for Inventory Management}

The DRL approach allows for an end-to-end implementation that directly optimizes for the decision. The traditional way of implementing inventory management systems is one of ``predict, then optimize''. In inventory management, this usually involves a demand forecasting system, optimized for forecasting accuracy, whose output is then consumed by some inventory optimization system. The integration of operations research principles into machine learning models represents an emerging trend in the field, and a number of pieces of research have aimed at bridging the two pieces, by either jointly predicting and optimizing, or by directly optimizing and removing the prediction step (see \cite{qi2022integrating} for a discussion of these approaches). For example, the directed regression of \cite{kao2009directed} seeks to minimize a loss that consists of a combination of ordinary least squares and empirical optimization, and more recently \cite{elmachtoub2022smart} develop a ``Smart Predict-then-Optimize'' framework that combines prediction and optimization.

Alternatively, as with DRL, the approach of directly predicting inventory decisions from data is increasingly popular. The ``big data newsvendor'' of \cite{rudin2014big}  directly predicts the optimal solution of a newsvendor problem from data. \cite{qi2023practical} use deep learning to predict the output of an oracle policy in a multi-period inventory management problem with lead times. Similarly, \cite{van2024supervised} integrate forecasting and inventory control to directly produce inventory decisions in the cases of  lost sales, perishable goods, and dual-sourcing inventory problems.

These latter approaches implementing end-to-end decision systems have proved successful in practical applications and industrial settings. The work of \cite{qi2023practical} was implemented at JD.com, \cite{liu2023ai} describe a move towards end-to-end inventory systems using RL at Alibaba, and the DRL buying method of \cite{Madeka2021} was successful in experiments at Amazon.com.

In addition to the empirical successes of data-driven approaches, theoretical results also bolsters their pertinence, such as the learnability results of \cite{Madeka2021} and VC theory work of \cite{xie2024vc} in the context of inventory management.

\section{Formulation and Implementation}\label{sec:formulation}

\subsection{Exo-MDP Formulation of the Problems}

Similarly to \cite{Madeka2021} and \cite{alvo2023neural}, we consider the Input-Driven MDP framework \cite{mao2018variance}, and more specifically the  Exo-MDP \cite{sinclair2023hindsight} setting.  As any Markov Decision Process,  Exo-MDPs are characterized by their state, transition dynamics,  control process (policy), and reward function. We here describe these components at a high level, and will then expand on their specific forms in Section~\ref{sec:experiments} for each of the considered problems.

\paragraph{State}

In the Exo-MDP framework, the state of the problem $\mathbf{z}_t$ at time $t$ can be split into exogenous and endogenous variables. The exogenous variables are driven by external processes that are outside of the control of the agent, while the endogenous variables are impacted by their actions. Typically, in an inventory management setting,  exogenous variables would include the (uncensored) demand, cost, or price of a product (in the absence of pricing decisions). Endogenous variables on the other hand would include things like inventory levels that are directly affected by decisions such as buying quantities. To facilitate the exposition in later sections, we explicitly split the state information in three categories:
\begin{itemize}
\item exogenous time series variables $\mathbf{x}_t$,
\item exogenous static variables $\mathbf{s}$,
\item endogenous variables $\mathbf{y}_t$,
\end{itemize}
so that the state at time $t$ is given by:
\begin{align*}
\mathbf{z}_t&=\left(\mathbf{x}_t,\mathbf{s},\mathbf{y}_t\right).
\end{align*}
Note that the exogenous time series variables $\mathbf{x}_t$ are made of the observations of  past realizations of the time series features $\mathbf{x}_t=(\mathbf{o}_0,\ldots,\mathbf{o}_{t-1})$, where $\mathbf{o}_t$ represents the realization of the time series features at time $t$. For example, if the only exogenous times series variable is the realized demands $d_t$, then $\mathbf{x}_t=(d_0,\ldots,d_{t-1})$. In practice, only a finite number $H$ of previous observations would be kept in memory, so that for all intents and purposes, we can assume that $\mathbf{x}_t=(\mathbf{o}_{t-H},\ldots,\mathbf{o}_{t-1})$ for some window $H>0$.

\paragraph{Control Process}
The control process is generated from a policy class $\Pi=\left\{\pi_{\theta,t}|\theta_t \in \Theta, t \in[0,T]\right\}$ parametrized by a vector $\bm\theta$ that could represent, for example, the weights in a neural network. In each period $t$, the policy generates an action vector $\mathbf{a}_t=\pi_{\theta,t}(\mathbf{z}_t)$ that depends on the current state. In an inventory management setting, the actions could be single dimensional and specify, say, how many units to purchase, or multi-dimensional and represent the number of units to purchase as well as how many units to return in the case of a problem that includes removal decisions; or how many units to purchase from domestic and import vendors in the case of dual sourcing. 

\paragraph{Transition Function}
The endogenous part of the state $\mathbf{y}_t$ evolves according to a transition function $f$ that depends on the current state $\mathbf{z}_t$, the realization of the exogenous inputs $\mathbf{o}_t$, and the action $\mathbf{a}_t$, so that at time $t+1$, we have:
\begin{align*}
\mathbf{y}_{t+1}=f(\mathbf{z}_t, \mathbf{o}_t, \mathbf{a}_t).
\end{align*}
As an example, if we consider a simple inventory management problem with no lead times and lost sales where the endogenous variable is single dimensional, representing the inventory level, we might have $y_{t+1}=\max(y_t - d_t + q_t, 0)$, where $d_t$ is the observed demand in period $t$, and $q_t$ the number of purchased units, so that the inventory level in the next period is equal to the units left after demand has been drained from the inventory at the beginning of the period plus the purchased units.

\paragraph{Reward Function}
We let $R_t(\bm\theta;\mathbf{z}_t)$ be the reward received in period $t$. We may also specify a different terminal reward function $R_T$ in the final period $T$. Letting $\mathbb{P}$ denote the full joint probability distribution of the exogenous inputs,  the value function is then expressed as the expected discounted sum of period rewards over the problem horizon, using a discount factor $\gamma$:
\begin{align}
J_T(\bm\theta;\mathbf{y}_0) = \mathbb{E}_{\mathbb{P}}\left[\sum_{t=0}^{T_1} \gamma^t R_t(\bm\theta;\mathbf{z}_t)\middle|\mathbf{y}_0\right].\label{eq:reward_func}
\end{align}
Note that we explicitly expressed the dependence of the reward on the initial endogenous state, since we are at liberty of setting it. Alternatively, we may consider an average reward value function given by:
\begin{align*}
J(\bm\theta) = \lim_{T\to\infty} \mathbb{E}_{\mathbb{P}}\left[\frac{1}{T}  \sum_{t=0}^{T-1} R_t(\bm\theta;\mathbf{z}_t)\right].
\end{align*}
This latter objective is especially useful in a stationary setting, although in practice it would often be computed as a finite average $\bar{J}_{T,b} = \mathbb{E}_\mathbb{P} \left[ \frac{1}{T-b}\sum_{t=b}^{T-1} R_t(\bm\theta;\mathbf{z}_t)\right]$ for some "burn-in" number of periods $b$ that allow for the system to reach steady state.

\paragraph{Exo-MDP Formulation}
Putting all the above components together, we can express the Exo-MDP formulation of the optimization problem in the following succinct form:
\begin{align}
\max_{\bm\theta} ~&  \mathbb{E}_{\mathbf{Y}_0}\left[J_T(\bm\theta;\mathbf{Y}_0)\right] \label{eq:opt_problem}\\
s.t.~& \mathbf{a}_t=\pi_{\theta,t}(\mathbf{z}_t) \nonumber\\
 &  \mathbf{z}_{t+1}=f(\mathbf{z}_t, \mathbf{o}_t,\mathbf{a}_t)\nonumber
\end{align}
This formulation is especially advantageous because it allows for a direct optimization of the problem, using stochastic gradients of the objective function with respect to the policy parameters $\bm\theta$. In particular, if the policy is expressed as a neural network, these gradients are readily obtained through automatic differentiation in most modern deep learning libraries such as PyTorch \cite{paszke2019pytorchimperativestylehighperformance}. It is important to note that this also hinges on the model being differentiable, and in particular in the transition function being differentiable.

\subsection{Policy Representation}\label{sec:policy_architecture}

\begin{figure}[htbp]
\centering
\includegraphics[scale=0.6]{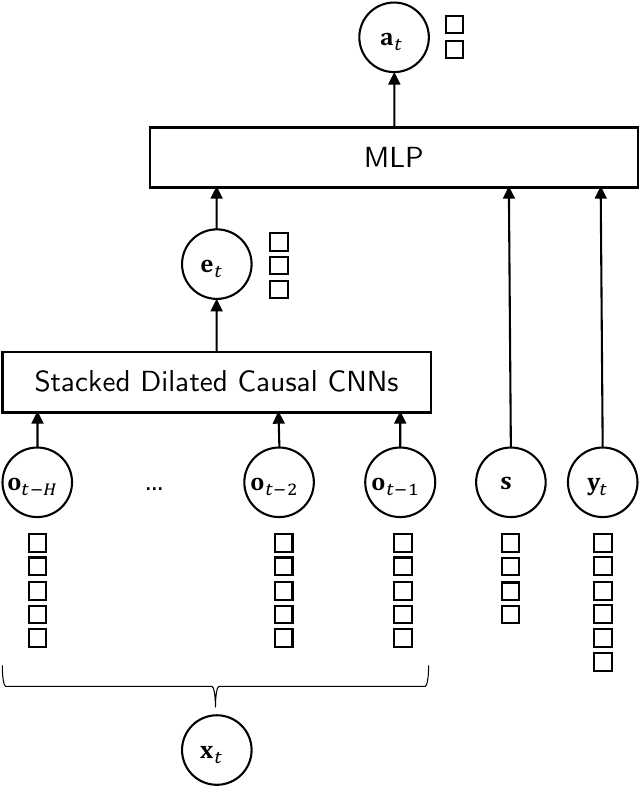}
\caption{Schematic representation of the policy architecture.}\label{fig:policy}
\end{figure}

We here describe the simple neural architecture used to represent all the policies trained in our paper. Recall that in the context of the Exo-MDP formulation,  we categorized the information in  three groups:
\begin{itemize}
\item exogenous time series variables,
\item exogenous static variables,
\item endogenous variables.
\end{itemize}
In our DRL approach, the policy is represented by a neural network whose weights are trained in order to learn an optimal policy.  The policy takes in the state $\mathbf{z}_t=(\mathbf{x}_t, \mathbf{s}, \mathbf{y}_t)$ and outputs the vector of actions $\mathbf{a}_t$.  There is a wide variety of possible architectures to represent the policy, and we here choose a very simple one. The policy neural network we consider is made of what is essentially a WaveNet \cite{van2016wavenet} encoder for the time series variables,  consisting of stacked causal dilated convolutions, the output of which is then combined in a Multi Layer Perceptron (MLP) with the static and endogenous variables.This setup is similar to what has already been successful in forecasting \cite{wen2017multihorizon} and RL for buying \cite{Madeka2021}.  Figure~\ref{fig:policy} presents a schematic representation of the policy architecture using the paper notation, and where $\mathbf{e}_t$ represents the encoded output of the the time series features.

Given this architectural design, the remaining choices to be made are:
\begin{itemize}
\item the size of the history window $H$ for the time series features,
\item the number of causal CNNs to be stacked and their dilation,
\item the number of layers and neurons per layer in the MLP,
\item the activation functions to be used.
\end{itemize}

We will use the same architecture for all the use-cases, unless otherwise stated. In particular,  for the time series features we will use $H=32$, meaning that we allow the agent to look at the 32 prior observations,  and 5 stacked causal CNNs with dilations of 1,2,4,8, and 16, respectively.  As to the  MLP, it consists of 2 hidden layers of 32 neurons.  We use ELU activation functions throughout.

We should point out that this is but one possible representation of the policy, one that is not particularly tailored to the usecases at hand. More involved, or more tailored representations could perform even better.

\subsection{DirectBackprop}\label{sec:directbp}

DirectBackprop \cite{Madeka2021}, also called Hindsight Differentiable Policy Optimization (HDPO) \cite{alvo2023neural}, is a stochastic gradient algorithm aiming to solve the optimization problem \eqref{eq:opt_problem}. This algorithm constructs a differentiable simulator of the inventory system, allowing for end-to-end training of the policy network. By leveraging historical data to create this simulator, DirectBackprop can capture complex dynamics such as stochastic vendor lead times, lost sales, correlated demand, and price matching. In  \cite{Madeka2021} the method outperforms model-free reinforcement learning and traditional newsvendor baselines in both simulations and real-world deployments, demonstrating its effectiveness in handling the complexities of modern inventory management systems. We refer to \cite{Madeka2021} for the description of the algorithm, which we also reproduce in Appendix~\ref{app:dbp}.

\section{Application of DRL to Classical Inventory Management Problems}\label{sec:experiments}

\subsection{Overview, Data, Training, and Evaluation}\label{sec:overview}

\paragraph{Overview}

We consider five classes of multi-period inventory management problems that have been extensively studied in the literature. These are:
\begin{itemize}
\item Multi-period inventory management with lost sales and null lead time (Section~\ref{sec:lost_sales}),
\item Multi-period inventory management with lost sales and positive lead time (Section~\ref{sec:lost_sales_lead_times}),
\item Multi-period inventory management with perishability (Section~\ref{sec:perishable}),
\item Multi-period inventory management with dual sourcing (Section~\ref{sec:dual_sourcing}),
\item Multi-period inventory management with removals (Section~\ref{sec:removals}).
\end{itemize}

Most inventory management problems have been studied under the assumptions of either backlogged or lost missed sales. The rationale being that while a lost sales assumption is much more realistic, the backlogging assumption allows for stronger theoretical results. Unfortunately, a backlogging assumption is most likely too restrictive in practice since research has shown that only about 17\% of customers would delay their purchase in an out-of-stock situation \cite{gruen2002retail}. Additionally, numerical studies further suggest that a backlogging assumption, in general, is a poor substitute for lost sales \cite{zipkin2008old}. As a result, we mostly focus on problem formulations involving lost sales, in particular also to demonstrate how DRL can shine on problems that prove intractable through traditional methods such as dynamic programming.

We use DirectBackprop to train policies represented by the simple neural architecture described in Section~\ref{sec:policy_architecture} with no additional tweaks such as normalization, and compare its results to some established benchmarks. Additionally, we pay attention to the structures of the learned policies to check whether they satisfy properties that are expected of optimal policies. This in turn motivates the regularization schemes proposed in Section~\ref{sec:sinn}. The training is done on simulated products, which consists of product economics, as well as single (uncensored) demand realizations to simulate the information that would be available to an agent in practice.

\paragraph{Data}

The data we generate includes product economics, as well as parameters of Gamma distributions that serve to model the demand, and from which we draw (single) path realizations for each product.  
The products are characterized by their price $p$, purchasing cost $c$, penalty for lost sale $k$, holding cost $h$, and mean demand $\mu$ and coefficient of variation $\nu$. Additionally, in the case of dual sourcing we will let $c$ be the purchasing cost from the domestic option, while we will introduce an import cost $c_i$\footnote{the expression for the import cost is justified in Section~\ref{sec:dual_sourcing}}; and in the case of inventory management with removals, we will let $r$ be the return value. The distributions used to draw these values are described in Table~\ref{tab:distributions}, where $Exp(\lambda)$ refers to an exponential distribution with mean $\lambda$, and $U_i$ denotes a uniform random variable between 0 and 1.

\begin{table}[htbp]
\centering
\caption{Product Distribution}\label{tab:distributions}
\begin{small}
\begin{tabular}{lcccccccc}
\toprule
 & $p$ & $c$ & $c_i$ & $r$ & $k$ & $h$ & $\mu$ & $\nu$\\
\midrule
Distribution & $Exp(p_m)$ & $p U_1$ & $c-\min\left(kU_6, U_2 c\right)$ & $c U_3$ & $k_m U_4$ & $Exp(h_m)$ & $Exp(\mu_m)$ & $U_5$\\
Values & $p_m=100$ & & & & $k_m=10$ & $h_m=5$ & $\mu_m=100$\\
\bottomrule
\end{tabular}
\end{small}
\end{table}

In particular, we use exponential distributions to draw the prices, holding costs, and mean demands, and use the drawn prices to set the (lower) purchasing costs, which in turn are used to set  even lower import costs (in the case of dual sourcing). The coefficient of variation of the demand distribution is randomly and uniformly sampled between 0 and 1, which can lead to distributions with narrow to large spreads. The drawn means and coefficients of variation are used to obtain the corresponding parameters for Gamma distributions from which we sample realizations of demand $\{D_t\}_{-H\leq t \leq T-1}$ for each product. The demands $D_t$ for $t=-H,\ldots,-1$ correspond to the historical observed demands in the $H$ periods prior to time 0, which serve to set the initial state, while the demands $D_t,~t=0,\ldots,T-1$ correspond to the observations that will gradually be revealed to the agent over the course of the rollouts over $T$ periods.

Our baseline population will contain 40,000 products drawn from this distribution, although we will also investigate the impact of the sample size on the performance of the algorithm in some of the examples.

\paragraph{Initialization} 
The initial values $\mathbf{x}_0$ and $\mathbf{s}$ of the exogenous parts of the product states are dictated by the data sampling. What remains to be initialized is the endogenous variables. Because in the exo-MDP framework we assume that we have knowledge of the exogenous variables, any choice for the intialization of the endogenous variables $\mathbf{y}_0$ is valid.  In all of our considered examples, the endogenous variables, while of different dimensions, all represent inventory units, either on-hand, or in transit. In \cite{Madeka2021}, where the data corresponded to real production data, three initializations were considered: 1) null initial values, 2) historical initial values, and 3) initial values inherited from where another policy left off. \cite{alvo2023neural} on the other hand also suggest using randomly sampled initial values. We opt for the latter, which allows for larger parts of the state space to be visited. This is especially important in cases such as inventory management with removals, where initializing at, say, null initial values, would result in a behavior where the algorithm never or seldom visits states that correspond to large overstock position, and would thus never be able to learn good removal policies.

\paragraph{Training}

Equipped with the training data generated according to the preceding description, we seek, for each usecase, to optimize the corresponding total expected reward given in Equation~\eqref{eq:reward_func} using a discount factor of 1 by rolling out the policy over $T=100$ periods (excluding the $H=32$ periods used to initialize the exogenous data)  and iteratively updating it using the Adam algorithm \cite{kingma2014adam} with a fixed learning rate of 0.001 over a fixed number of 1,000 epochs. The training hyperparameters used across examples are summarized in Table~\ref{tab:hyperparams}.

\begin{table}[htbp]
\begin{center}
\caption{Training hyperparameters}\label{tab:hyperparams}
\begin{tabular}{lc}
\toprule
Hyperparameter & Value \\
\midrule
$H$ & 32\\
$T$ & 100\\
$\gamma$ & 1\\
Causal CNN dilations & [1,2,4,8,16]\\
CNN channels & 8\\
MLP layers & 2\\
MLP neurons per layer & 32\\
Activation Functions & ELU\\
Optimization algorithm & Adam\\
Learning rate & 0.001\\
Sample size & 40,000\\
Batch size & 2,500\\
Epochs & 1,000\\
\bottomrule
\end{tabular}
\end{center}
\end{table}

\paragraph{Evaluation}

To evaluate the policies, and given that we consider stationary problem settings, we use an empirical average reward function as:
\begin{align*}
\bar{J}_{T,b} = \mathbb{E}_\mathbb{P} \left[ \frac{1}{T-b}\sum_{t=b}^{T-1} R_t(\bm\theta;\mathbf{z}_t)\right],
\end{align*}
computed over a sample of 100,000 independent products, and 500 periods excluding a burn-in of 20 periods (i.e. 520 periods in total). For the evaluation, we also initialize all endogenous variables to 0.

\subsection{Multi-Period Inventory Management with Lost Sales}\label{sec:lost_sales}

\subsubsection{Description}

The multi-period inventory management problem is one of the earliest ones to be studied in the literature. The model, initially known as the Arrow-Harris-Marschak \cite{kenneth1951optimal},  has been comprehensively studied both in its backlogging and lost sales variations. The models with null lead times and no fixed ordering costs offer especially simple settings where myopic policies are optimal \citep{karlin1960dynamic,veinott1965optimal}. As a result, we not only know the structure of the optimal policy, but they typically admit closed form solutions in the form of an order-up-to policy that can be explicitly computed as a quantile of the demand distribution.

In its simplest formulation, this model considers a single product that can be periodically purchased and whose demand in each of these periods is independent and identically distributed. It assumes that purchases are made at the beginning of the period, after which demand realizes and revenue is generated from sales. Any demand that exceeds the available inventory is lost and incurs a penalty that captures the loss of goodwill from the customer and their future business. On the other hand, any excess inventory incurs a linear holding cost. 

This is possibly the simplest setting for a multi-period inventory management problem, one for which everything is known. As a result, because of its simplicity, it is a convenient case on which to test a DRL approach, and also to get a sense of the impact of a few factors such as sample size or historical window, which we will investigate below.

\subsubsection{MDP Formulation}

The model formulation assuming lost sales takes the following form:
\paragraph{State} The endogenous state is simply given by the inventory level $y_t$, while the exogenous time series variables $\mathbf{x}_t$ consists of the previous $H$ demand realizations ($\mathbf{x}_t=(d_{t-H},\ldots,d_{t-1})$) and the exogenous static vector contains the economic parameters of a product, $\mathbf{s}=(p,c,h,b)$, where $p$ is the price, $c$ the purchasing cost, $h$ the holding cost, and $b$ the penalty for lost sale.

\paragraph{Action} The action $a_t=q_t$ at time $t$ is given by the number of units to purchase at time $t$, which will also arrive at the beginning of the period.

\paragraph{Transition Function} The transition function is simply given by $y_{t+1} = \max(y_t + q_t - d_t , 0)$, meaning that we sell up to $d_t$ units out of the $y_t+q_t$ units that are available for sale, and any unit left over is carried on to the next period.

\paragraph{Reward Function} In each period, we receive $p$ for each sold unit, and incur unit costs of $c$ for any purchased unit, $b$ for any missed sale, and $h$ for any leftover unit, yielding:
\begin{align*}
R_t &= p \min(d_t, y_t+q_t) - c q_t - b \max(d_t - y_t - q_t, 0) - h \max(y_t + q_t - d_t, 0).
\end{align*}

\subsubsection{Benchmark}

There is an obvious benchmark to this problem, which is the optimal myopic policy \citep{karlin1960dynamic,veinott1965optimal}. This policy is called myopic because it is the one obtained if one were to solve a single period problem, i.e. a newsvendor problem, using a full recovery value ($\gamma c$). The optimal policy in this scenario is an order-up-to  policy, where the optimal order-up-to level is given by a \emph{critical quantile} $y^*$ of the demand distribution expressed as:
\begin{align*}
F_D(y^*) = \frac{c_u}{c_u+c_o},
\end{align*}
where $F_D$ is the cumulative density function of the demand distribution $D$, and $c_u$ and $c_o$ are the underage and overage costs, defined in our case as $c_u=p-c+b$ and $c_o=h+(1-\gamma c)=h$. The optimal action is then to buy up to this level if we are below it, and nothing if we are above it: $q_t=\max(y^*-y_t,0)$. Note that $y_t\leq y^*$ implies that $y_s\leq y^*$ for any $s>t$.

\subsubsection{Experiments}

This example allows us to compare our learned policies to a known optimal policy. We also use the example to evaluate the impact of the training sample size $N$, as well as history window $H$ on the out-of-sample performance of the learned policies. We consider history windows $H$ of lengths 16, 32 (default), and 64; and a wide range of training sample sizes $N$, and compare the results of the DRL-learned policies to the optimal policies rolled on the same test set described in Section~\ref{sec:overview}.

\begin{remark}\label{rem:disadvantage}
It is important to note that the DRL agent is at a disadvantage compared to the optimal policy for a couple of reasons: 1) the DRL agent is not aware that the demand distributions are Gamma distributed, 2) the DRL agent only has access to the previous $H$ demand realizations. In setting the benchmark, we thus use information that is unavailable to the agent. 
\end{remark}

Because of Remark~\ref{rem:disadvantage}, we additionally consider a ``non-omniscient'' benchmark where the benchmark agent still knows that the data is Gamma-distributed and that the optimal policy is to buy up to the critical quantile, but only has access to the previous $H$ demand realizations. This agent then fits a Gamma distribution to these data points using the method of moments, i.e. computing empirical mean and standard deviation and deriving the shape and scale parameters corresponding to these moments.

We mentioned in Section~\ref{sec:overview} that we randomly initialized the endogenous variables, and in this example, in each epoch, we randomly set the initial inventory level $y_0$ of a product to be uniformly sampled between 0 and $2d_{-1}$. We further set the terminal condition to be $R_T(\mathbf{z}_T)=c y_T$.

\subsubsection{Results}

\paragraph{Evaluation Results}
The default configuration corresponds to $N=40,000$ and $H=32$. After training DirectBackprop using the default settings shown in Table~\ref{tab:hyperparams}, we evaluate it on the test set, which we recall consists of 100,000 randomly sampled products, over 520 periods of which we discard the first 20 as burn-in periods. We similarly run the benchmark policies on the same set (omniscient and non-omniscient benchmark policies). Table~\ref{tab:basic_default_results} shows the average reward estimates for both the learned and optimal policies. We observe that the DRL policy, even with only 40,000 products to train on and its information disadvantage with respect to the optimal policy (see Remark~\ref{rem:disadvantage}), gets within 0.5\% of the optimal policy, and is exactly on par with the non-omniscient optimal policy, which has access to the same amount of information, although it further knows 1) that the data originated from a Gamma distribution, and 2) that the optimal policy is base-stock with respect to the critical ratio. It is somewhat remarkable that the DRL agent was able to learn to generate a policy that mimics this behavior on its own.

\begin{table}[htbp]
\centering
\caption{Results of the learned and optimal policies on the test set for the default configuration on the Example~\ref{sec:lost_sales}.}\label{tab:basic_default_results}
\begin{tabular}{lccc}
\toprule
Agent & reward & gap \\
 \bottomrule
 DRL & 4,548.95 & -\%\\
  Non-Omniscient Optimal & 4,548.95  & 0.00\% \\
 Omniscient Optimal & 4,567.58  & -0.41\% \\
 \bottomrule
\end{tabular}
\end{table}

\begin{figure}[htbp]
\centering
\includegraphics[scale=0.6]{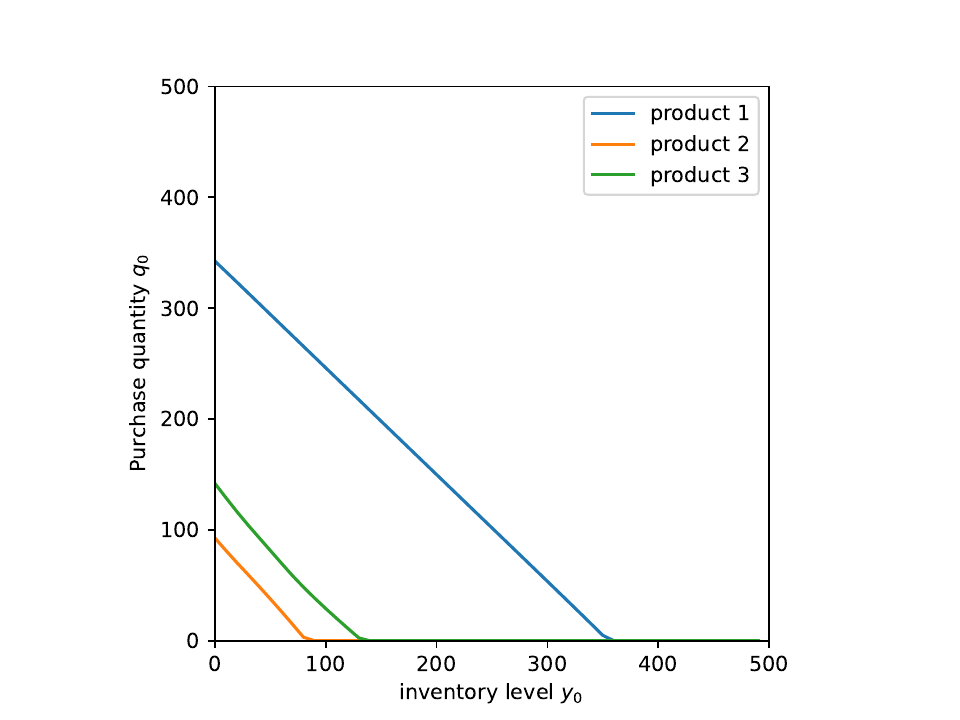}
\caption{Actions (purchase quantity) produced by the learned DRL policy as a function of the inventory level at a fixed time for three different products.}\label{fig:outl}
\end{figure}

\paragraph{Structure}
There are several things we might want to dive deeper into. The first is related to the structure of the learned policy. We know, and made use of the fact, that the optimal policy follows an order-up-to level (also known as base-stock), which is something we used when rolling out the optimal policy, but did not inform the DRL agent of in any way. A natural question is then whether the DRL agent is able to recover this (admittedly simple) structure. Figure~\ref{fig:outl} shows the actions produced at time 0 (purchase quantities $q_0$) for three random products, where we probed the agent for different values of the inventory level $y_0$. We observe that the obtained curves are near-perfect diagonal lines with slopes of -1, which corresponds exactly to the expected optimal behavior, which is given by $q_0^*=\max(y_0^*-y_0,0)$ for some optimal order-up-to level $y^*_0$.

\begin{figure}[htbp]
\centering
\includegraphics[scale=0.45]{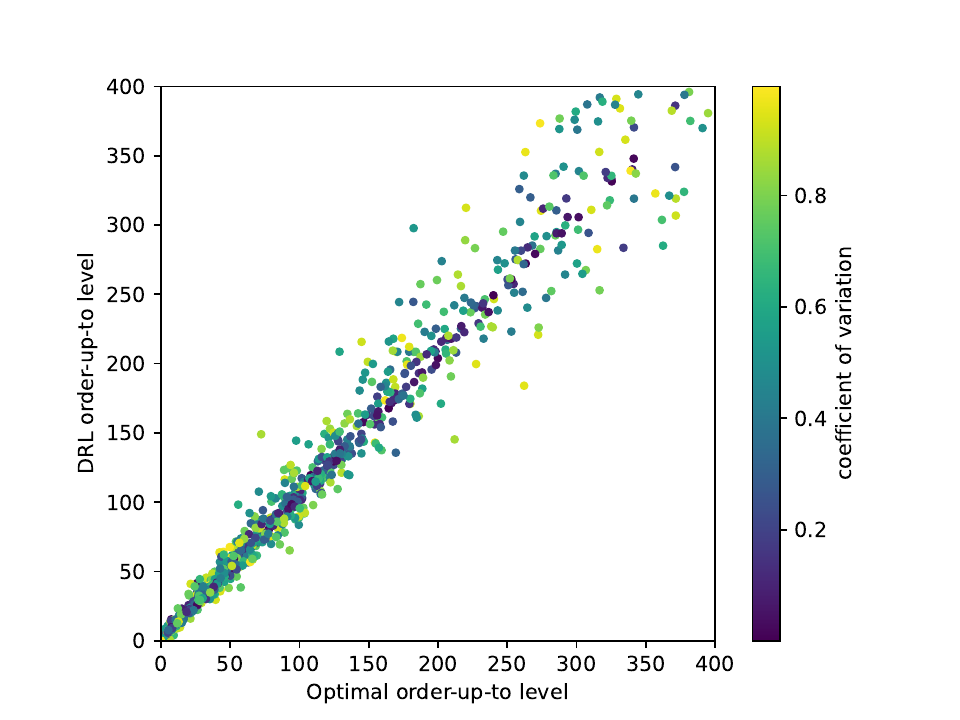}
\includegraphics[scale=0.45]{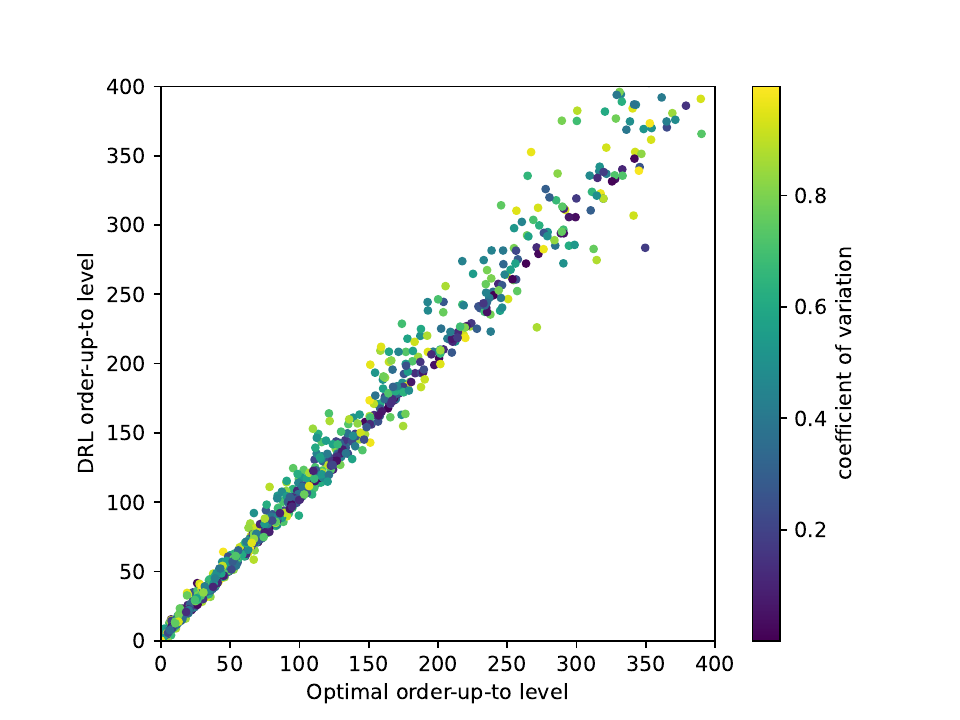}
\caption{Order-up-to level predicted by the DRL policy against the optimal order-up-to level for 1,000 random products, coded by coefficient of variation of the demand, for the omniscient benchmark agent (left), and non-omniscient benchmark agent (right).}\label{fig:tipvstip}
\end{figure}

Given that the DRL agent appropriately learns order-up-to policies, the next question is how close these are from the true optimal order-up-to levels. We recall once again that the sole  information the DRL agent has about the demand is through the previous $H=32$ realizations. As a result we also expect that it should perform better on products whose demands are less volatile, i.e. with lower coefficients of variation, and are thus easier to estimate. In order to approximate the learned order-up-to levels, we use the order quantity produced for an inventory level of 0: $\hat{y}_0^*=q_0$. We then plot $\hat{y}_0^*$ against $y_0^*$ for 1,000 random products in Figure~\ref{fig:tipvstip}, also color-coding the points with the products' coefficient of variation.

\begin{figure}[htbp]
\centering
\includegraphics[scale=0.4]{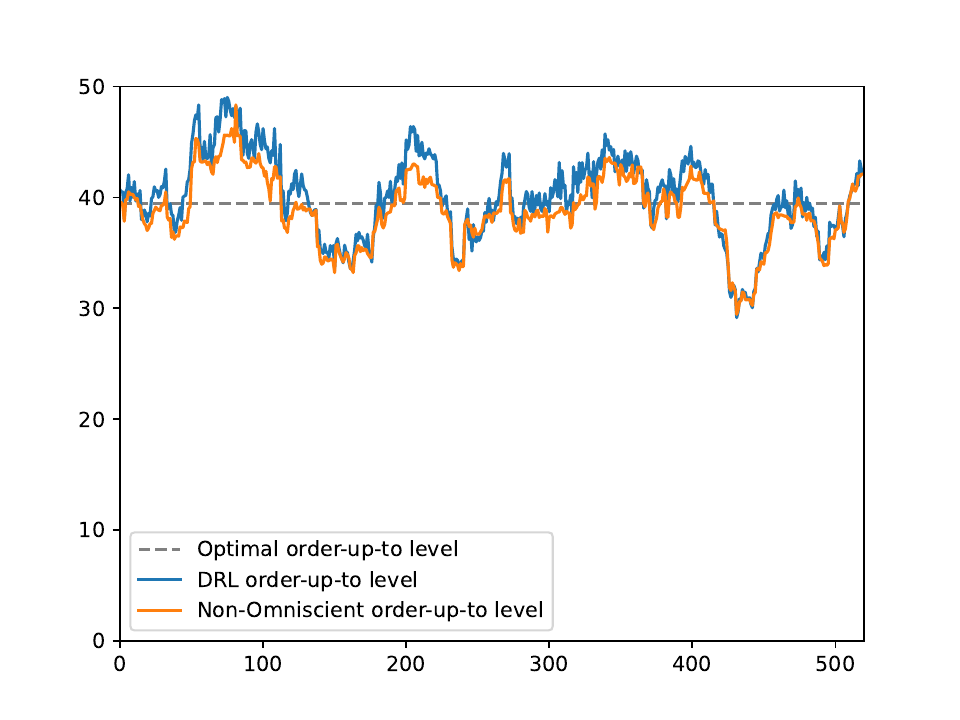}
\includegraphics[scale=0.4]{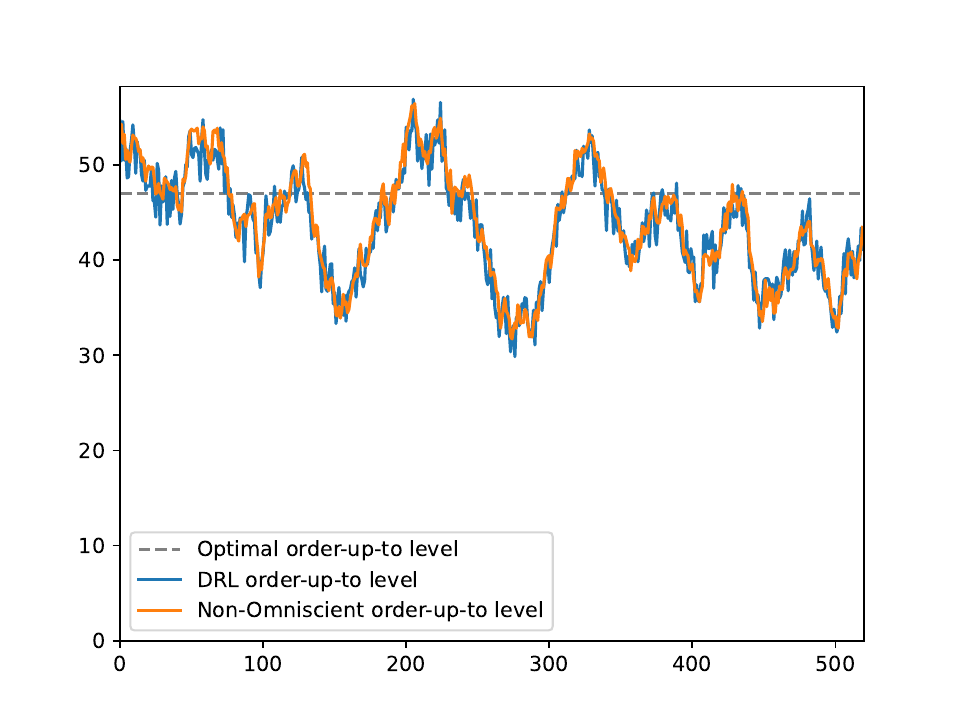}
\includegraphics[scale=0.4]{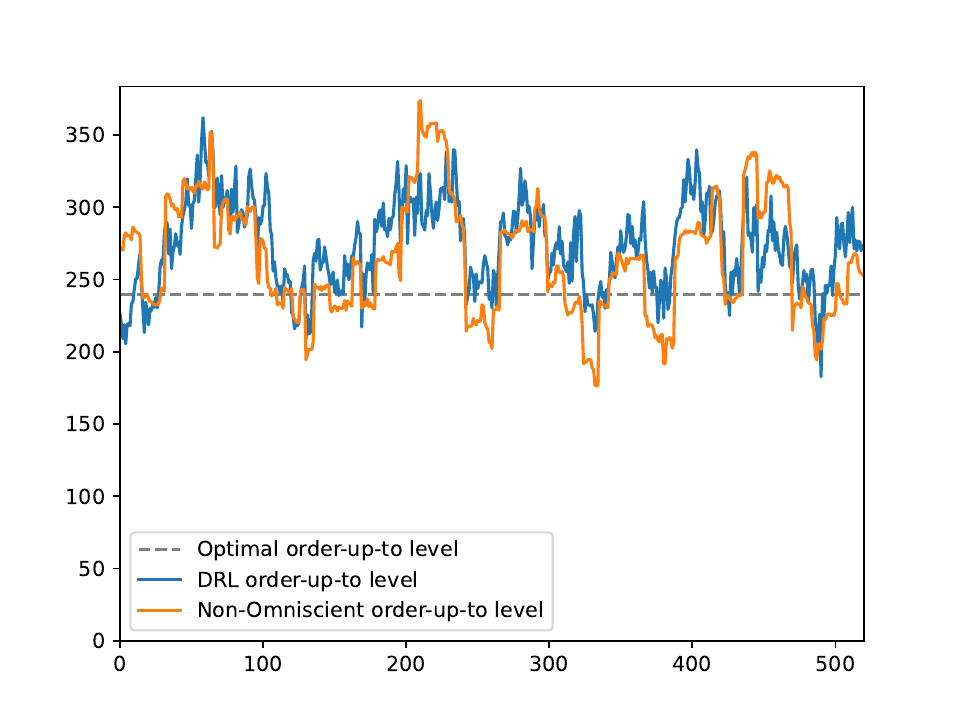}
\includegraphics[scale=0.4]{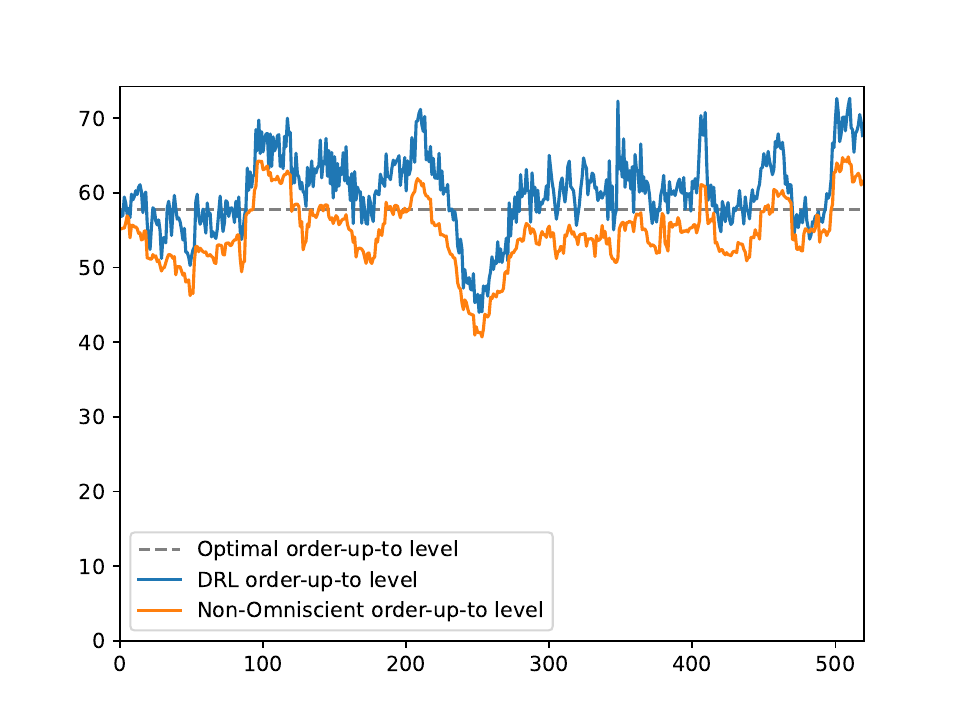}
\caption{Order-up-to levels used by the DRL agent and two benchmark agents for a few products over the course of the test rollout.}\label{fig:lost_sales_policies}
\end{figure}

Finally, we compare for a few products over the course of the test rollout, the order-up-to levels learned by the DRL agent with those of the omniscient benchmark agent (which is constant), and that of the non-omniscient benchmark agent. These are produced in Figure~\ref{fig:lost_sales_policies}. The latter is quite remarkable in that we observe that the DRL agent essentially learned on its own to predict and buy up to the critical quantile. In this manner, the end-to-end strategy of DRL is justified, as opposed to the usual two-stage approach of first predicting a demand distribution by minimizing some forms of quantile losses, and then using that distribution in a buying problem. We here directly optimize for the objective function at hand, instead of having to optimize intermediary proximal problems.

\begin{figure}[htbp]
\centering
\includegraphics[scale=0.5]{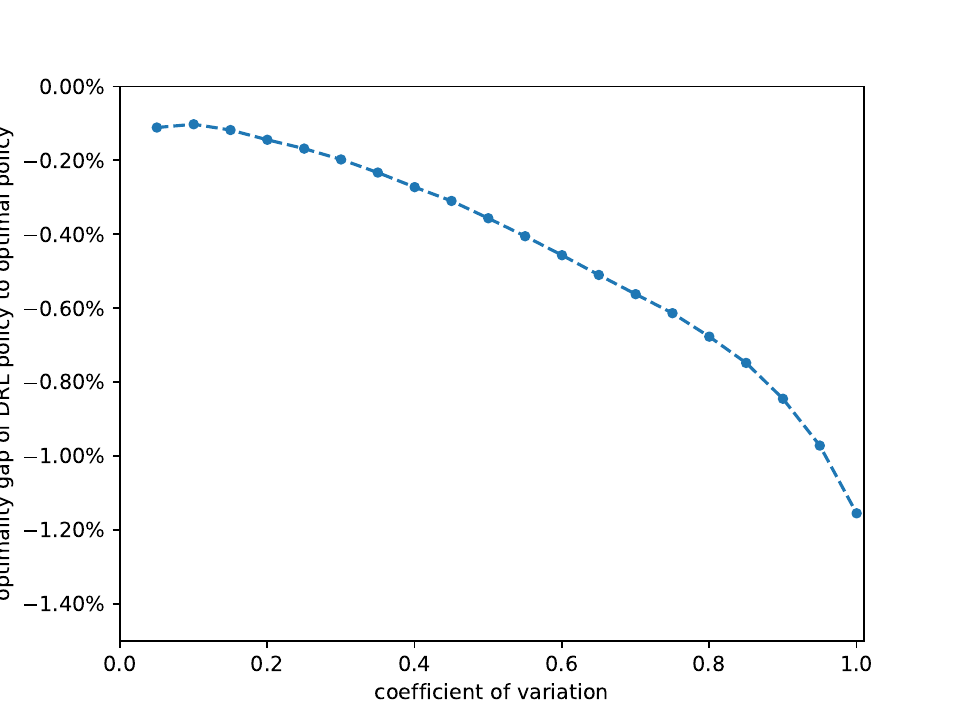}
\caption{Gap of the DRL policy to the optimal policy as a function of the demand coefficient of variation.}\label{fig:cov}
\end{figure}

\paragraph{Impact of the Demand Coefficient of Variation}
We observe that there is a good alignment between the predicted order-up-to levels and the optimal ones. We also see from the color-coding that it appears products with higher coefficients of variation are further away from the diagonal. We confirm the observation by generating sets of products with fixed coefficients of variation, but otherwise following the data generating process from Section~\ref{sec:overview}, and evaluate the performance of the DRL agent against the optimal policy on each set. Figure~\ref{fig:cov} shows that the gap to the optimal policy does indeed decrease as the coefficient of variation increases, which is expected since the agent is working with a fixed number of samples to learn about the demand distribution.

\begin{figure}[htbp]
\centering
\includegraphics[scale=0.45]{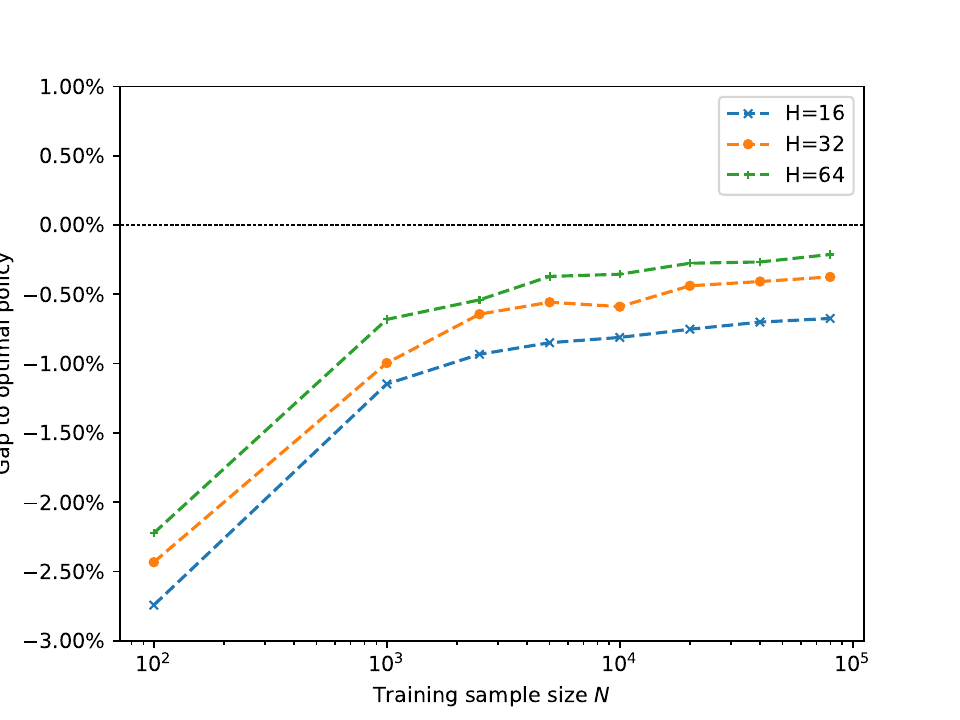}
\includegraphics[scale=0.45]{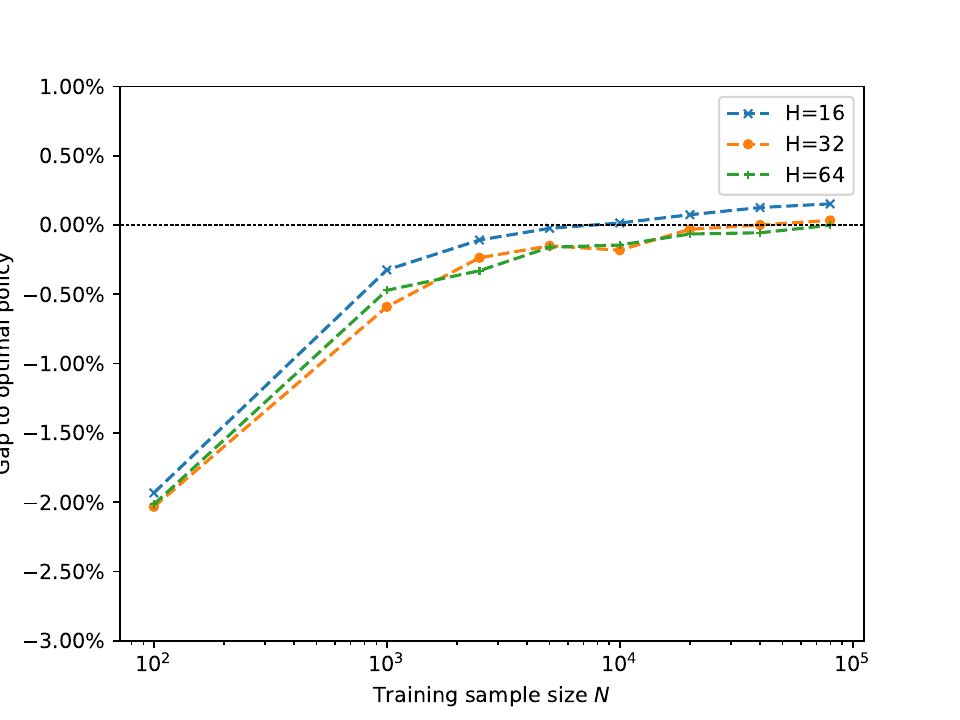}
\caption{Gap of the DRL policies to the optimal policy (left) and non-omniscient optimal policy (right) as a function of the training sample size $N$ for different values of the history window $H$.}\label{fig:reward_vs_N}
\end{figure}

\paragraph{Impact of Training Sample Size $N$ and History Window $H$}
The decrease in performance with increasingly volatile demand suggests that we might perform better with a longer history window $H$, as well as a bigger training set $N$. We here investigate the impact of those factors. We replicate the same experiment as for the default configurations using different values of $H$ and $N$. Note that when we change the history window, we also modify the Dilated CNN stack accordingly, adding a layer with dilation of 32 for $H=64$, and removing the layer with a dilation of 16 for $H=16$. We plot the gap to the benchmarks on the test set obtained for DRL policies learned using $H=16,32,64$ and a range of sizes of the training sample in Figure~\ref{fig:reward_vs_N}\footnote{We also extended the number of epochs to 2,000 for cases where the sample size was less than 5,000 to ensure a minimum number of gradient steps since we kept the batch size to 2,500 (or the sample size if it was less than 2,500)}. The figure highlights the fact that the performance of the DRL agents increases with the training sample size $N$, as well as with the length of the history window $H$. Notably, with sufficient training samples, the DRL learns to perform identically to the optimal non-omniscient policy.

We should also point out that the order-up-to level structure of the optimal policy was quickly learned by the DRL agent, and as few as 3 products were enough for it to pick up on this structure and produce what looks like and order-up-to policy, albeit not a very good one.

\subsection{Multi-Period Inventory Management with Lost Sales and Lead Times}\label{sec:lost_sales_lead_times}

\subsubsection{Overview}\label{sec:leadtime_overview}

The model explored in the previous section assumed that purchased units arrived instantaneously, and the consideration of such a stylized model led to a simple tractable optimal policy in the form of a myopic policy that can be written analytically. To add realism, the model has been extended to the inclusion of lead times, which characterize the lag between when an order is placed and when it is received. These models were first considered in \cite{karlin1958inventory} where the optimality of order-up-to policies was shown for models with backlogging. The paper also showed that no such simple policy holds for a model with lost sales, although some monotonicity and sensitivity results were demonstrated. Bounds on the optimal policies as well as heuristics were then derived in \cite{morton1969bounds,morton1971near}. In particular, \cite{morton1969bounds} suggests what has been called a vector base-stock policy. Other heuristics have later been proposed, in particular the dual-balancing policy \cite{levi20082} that gives a worst-case bound on its performance. \cite{huh2009asymptotic} showed that the best base-stock policy is asymptotically optimal as the lost-sales penalty grows, and \cite{bijvank2014robustness} extend the result to show that the base-stock policy for the backlogging case is also asymptotically optimal under some conditions.  Other asymptotic results include the asymptotic optimality of the constant-order policies introduced by \cite{reiman2004new} for large lead times \cite{goldberg2016asymptotic, xin2016optimality}, the asymptotic optimality of a sequence of capped base-stock policies \cite{xin2021understanding}, and more recent work demonstrates that a new policy, a \emph{projected inventory level} (PIL) policy is asymptotically optimal in two regimes \cite{van2024projected}. Newer structural results were derived in \cite{zipkin2008structure}, who introduced the use $L^\natural$-convexity to analyze the structure of optimal policies in inventory management, and proved additional bounds and structural properties of the optimal policy. 

Our goal in this section is to investigate the performance of the DRL agent on a problem that has long been considered challenging in the inventory management literature.

\subsubsection{MDP Formulation}

The model formulation of the problem with lost sales and lead times is very similar to the one with no lead time, except that we now must keep track of the pipeline inventory.

\paragraph{State} The endogenous state $\mathbf{y}_t$ is comprised of the on-hand inventory level $y^0_t$ as well as pipeline inventory levels $y^i_t$ for $i=1,\ldots,L-1$ corresponding to the units purchased $L-i$ periods earlier, and set to arrive in $i$ periods. The exogenous time series variables $\mathbf{x}_t$ consist in the previous $H$ demand realizations ($\mathbf{x}_t=(d_{t-H},\ldots,d_{t-1})$) and the exogenous static vector contains the economic parameters of a product, $\mathbf{s}=(p,c,h,b)$, where $p$ is the price, $c$ the purchasing cost, $h$ the holding cost, and $b$ the penalty for lost sale.

\paragraph{Action} The action $a_t=q_t$ at time $t$ is given by the number of units to purchase at time $t$, which will  arrive $L$ periods hence.

\paragraph{Transition Function} The transition function is given by:
\begin{align*}
y_{t+1}^k &= \begin{cases}
\max(y_t^0 - d_t , 0)+y_t^1,& k=0\\
y_t^{k+1},& 0<k<L-1\\
q_t,& k=L-1
\end{cases},
\end{align*}
reflecting the fact that any units left over after sales carry over to the next period, to which we add the $y_t^1$ units  to arrive in the next period. The other pipeline units are shifted down the pipeline, and the newly ordered units go to the back of the pipeline.

\paragraph{Reward Function} The reward function is almost identical to the one in the no-lead-time case. In each period, we receive $p$ for each sold unit, and incur unit costs of $c$ for any purchased unit, $b$ for any missed sale, and $h$ for any leftover unit, yielding:
\begin{align*}
R_t &= p \min(d_t, y_t^0) - c q_t - b \max(d_t - y_t^0, 0) - h \max(y_t^0 - d_t, 0).
\end{align*}

\subsubsection{Benchmarks}

We already mentioned a few benchmarks in Section~\ref{sec:leadtime_overview}, in addition to which we can mention the myopic policy \cite{morton1971near}, which aims to minimize the one period cost to be incurred in the period in which the newly ordered units are received. In spite of being heuristics, many of these heuristics still carry a high computational cost. For example, the myopic policy requires the evaluation of $L$ nested expectations for which there are no closed form expressions and is thus only practical for small lead times (2-3), otherwise requiring Monte-Carlo simulation or numerical integration.

As a result we consider two of the most commonly used and practical heuristics, the standard base-stock, and standard vector base-stock policies. In order to introduce them, it is useful to consider a different, although equivalent, representation of the endogenous state. Instead of considering the on-hand and pipeline inventory, $y_t^0$ and $y_t^k$ respectively, we consider the cumulative pipeline and on-hand inventory starting from the most recent orders. We let $u_{L}=0$, and $u_{l}=\sum_{k=l}^{L-1}y_l,~l=0,\ldots,L-1$, so that $u_{L-l}$ represents the sum of orders placed in the previous $l$ periods, and in particular $u_0$ is the inventory position, i.e. the sum of all on-hand and pipeline inventory. We similarly let $D_t^l$ represent the cumulative demand from periods $t+l$ to $t+L$ (inclusive): $D_t^{l}=\sum_{k=l}^L D_{t+l}$.

\begin{figure}[htbp]
\centering
\includegraphics[scale=0.65]{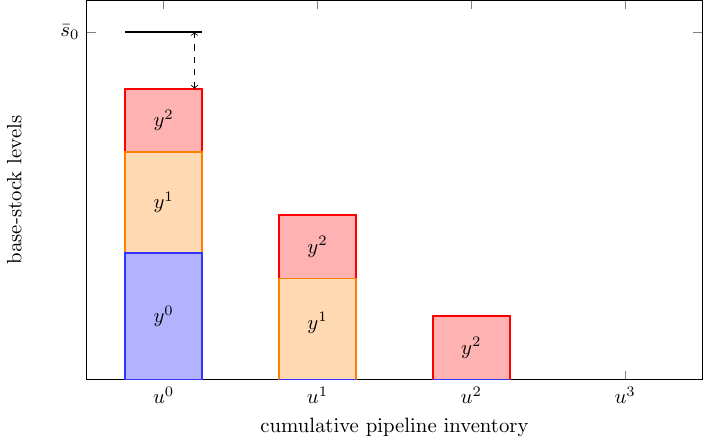}
\includegraphics[scale=0.65]{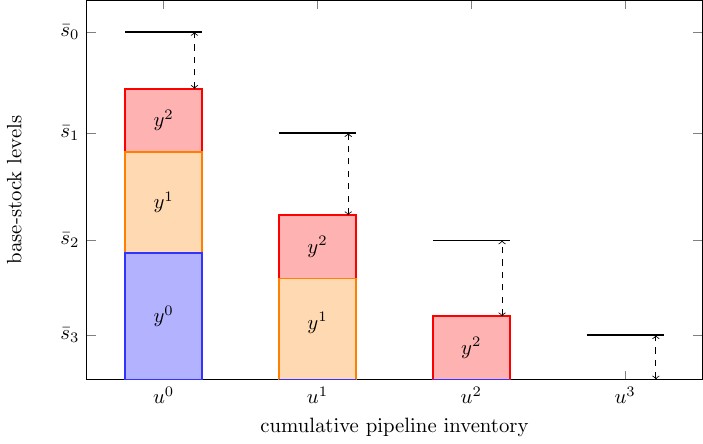}
\caption{Illustration of the base-stock (left) and vector base-stock (right) policies for an example with a lead time of $L=3$.}\label{fig:leadtime_benchmark_illustration}
\end{figure}

\paragraph{Standard base-stock policy} The standard base-stock policy applies the same logic to the case with lost sales as to the case with backlogging. In other words, the policy brings the inventory \emph{position}, i.e. the sum of on-hand and pipeline inventory, up to an order-up-to level $\bar{s}_L$ if the inventory position is under it, and does nothing if it is above. This order-up-to level is the one that is optimal in the backlogging case, and is given by $\bar{s}_L$ that satisfies:
\begin{align*}
F_{D_t^{L}}(\bar{s}_L)&= \frac{c_u}{c_u+c_o},
\end{align*}
where $c_u$ and $c_o$ are the same underage and overage costs as in the no-lead time case ($c_u=p-c+b$ and $c_o=h$), and $D_t^{L}$ is the demand over the periods $t$ through $t+L$ (inclusive). In other words, the purchase quantity is given by:
\begin{align*}
q_t = \max\left(\bar{s}_L - u_t^0,0\right).
\end{align*}
The policy is illustrated in Figure~\ref{fig:leadtime_benchmark_illustration} (left) for a lead time of $L=3$.

\paragraph{Standard Vector base-stock policy}
The standard vector base-stock applies a similar logic to the simple base-stock policy, except that it applies an order-up-to level for each of the components $u_t^k$, and not just $u_t^0$. It thus considers a vector of order-up-to level $\mathbf{\bar{s}}=(\bar{s}_0,\ldots,\bar{s}_L)$, and purchases the minimum quantity that allows for each component to remain under its base-stock level, and nothing if one of the components is already above it. The purchase order quantity is thus given by:
\begin{align*}
q_t &= \max\left(\min\left\{\bar{s}_l-u_t^l,~l=0,\ldots,L\right\},0\right).
\end{align*}
The vector of base-stock levels is found by considering the same quantile condition as in the case of the simple base-stock policy, but applying it now for each cumulative demand $D_t^l$:
\begin{align*}
F_{D_t^{l}}(\bar{s}_l)&= \frac{c_u}{c_u+c_o}.
\end{align*}
This policy is motivated by the fact that \cite{morton1969bounds} proved that for the optimal policy, the bounds $u_t^l + q_t\leq \bar{s}_l$ held for each $l=0,\ldots,L$. The policy is illustrated in Figure~\ref{fig:leadtime_benchmark_illustration} (right) for a lead time of $L=3$.

\subsubsection{Experiments}

We trained a DRL policy using the default values of Section~\ref{sec:overview} for lead times $L=2,3,4,5,6,7$, and then evaluated those policies, along with the two benchmarks on the same test set of 100,000 simulated products.

\subsubsection{Results}

\begin{figure}[htbp]
\centering
\includegraphics[scale=0.45]{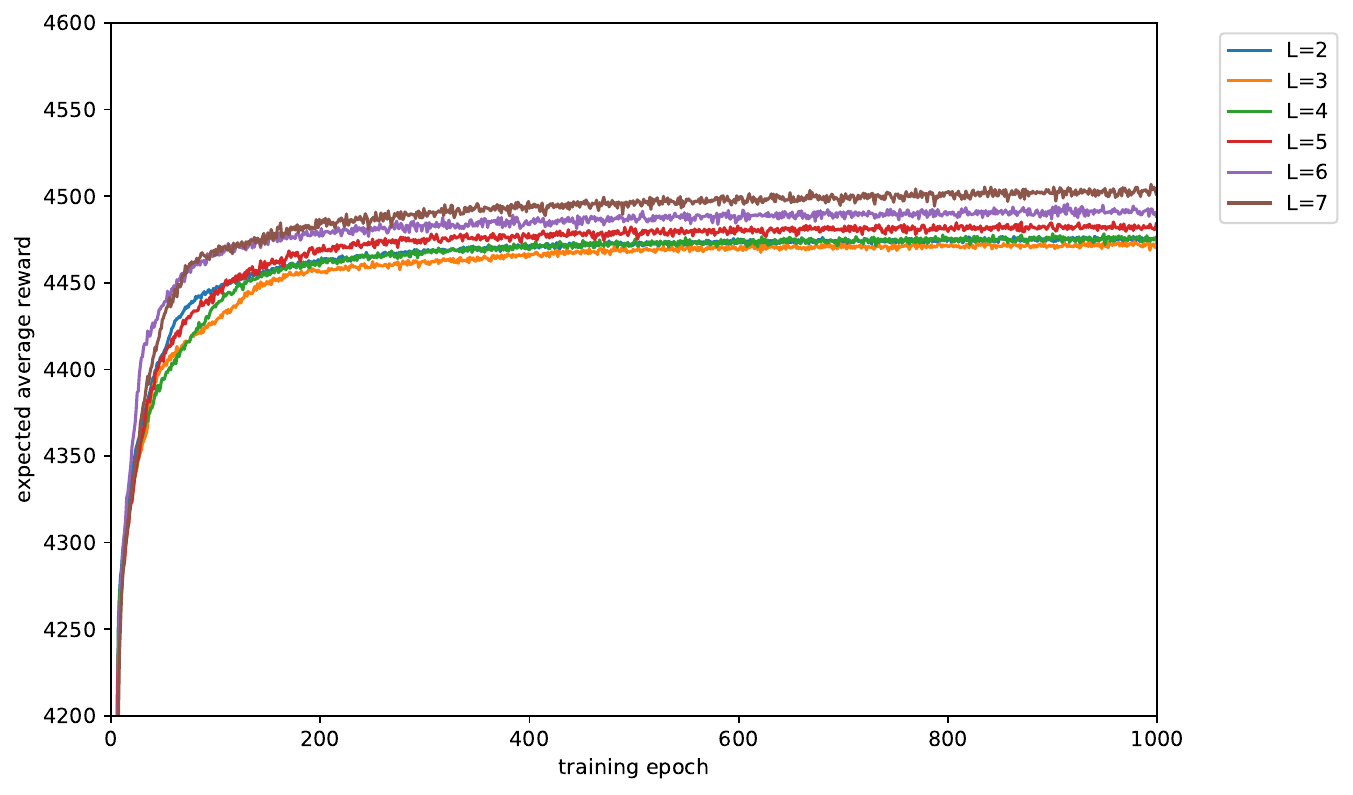}
\caption{Training curves of the DRL policy for different values of the leadtime.}\label{fig:leadtime_training}
\end{figure}

Figure~\ref{fig:leadtime_training} shows the expected average rewards during the training of the policies for the different values of the leadtime.

\begin{figure}[htbp]
\centering
\includegraphics[scale=0.5]{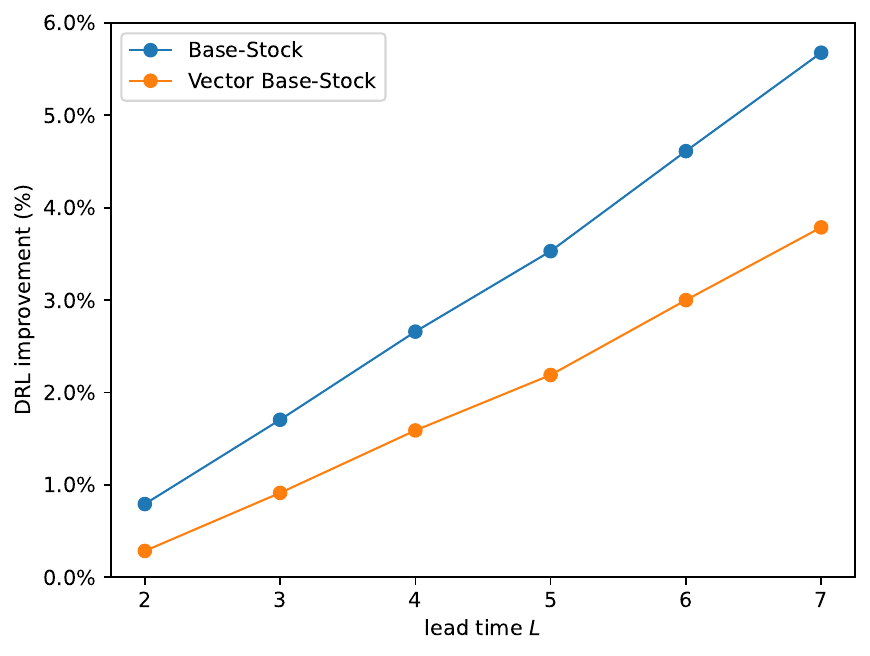}
\caption{Relative improvement of the DRL policies over the benchmarks in the case of lost sales with lead times, as a function of the lead time.}\label{fig:leadtimes_results}
\end{figure}

The results of the policies, DRL-learned and benchmarks, are presented in Table~\ref{tab:leadtimes_results} and Figure~\ref{fig:leadtimes_results}, using the vector base-stock policy as the reference. We observe that the DRL policies outperforms the heuristics, which we recall have access to more information since they are being set with the knowledge that the demand is stationary, Gamma-distributed, and has access to the parameters of each product's demand distribution. The benefits of the DRL approach also seem to increase with the dimension of the problem, i.e. as the lead time increases.

\begin{table}[htbp]
\begin{small}
\centering
\caption{Results of the DRL and benchmark policies on the case with lost sales and lead times for different values of the lead times.}\label{tab:leadtimes_results}
\begin{tabular}{llcccccc}
\toprule
\textbf{Policy} & & 2 & 3 & 4 & 5 & 6 & 7 \\
\midrule
\textbf{DRL} & reward &   4,418.50   &   4,385.43   &   4,360.48  &   4,336.16   &   4,324.03   &   4,312.96   \\
\midrule
\textbf{Vector Base-Stock} & reward & 4,405.93 & 4,345.74 & 4,292.26 & 4,243.25 & 4,198.09 & 4,155.59 \\
 & DRL vs BS (\%) & 0.29\%	&	0.91\%	&	1.59\%	&	2.19\%	&	3.00\%	&	3.79\% \\
\midrule
\textbf{Base-Stock} & reward & 4,383.73 & 4,311.92 & 4,247.55 & 4,188.32 & 4,133.38 & 4,081.25 \\
 & DRL vs BS (\%) & 0.79\%	&	1.71\%	&	2.66\%	&	3.53\%	&	4.61\%	&	5.68\% \\
\bottomrule
\end{tabular}
\end{small}
\end{table}

\begin{figure}[htbp]
\centering
\includegraphics[scale=0.42]{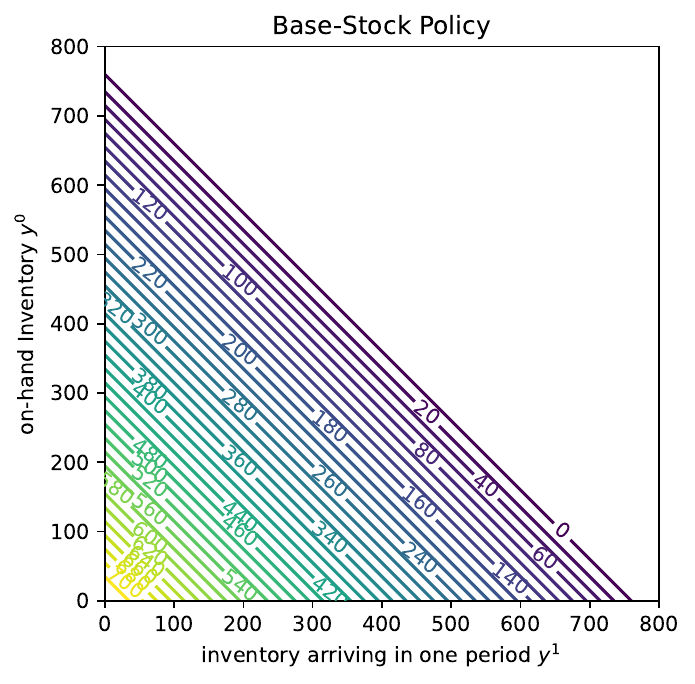}
\includegraphics[scale=0.42]{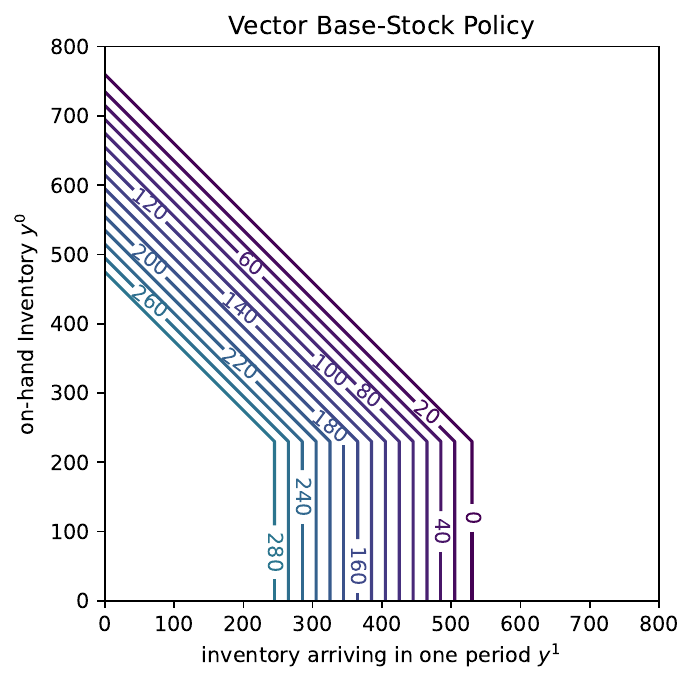}
\includegraphics[scale=0.42]{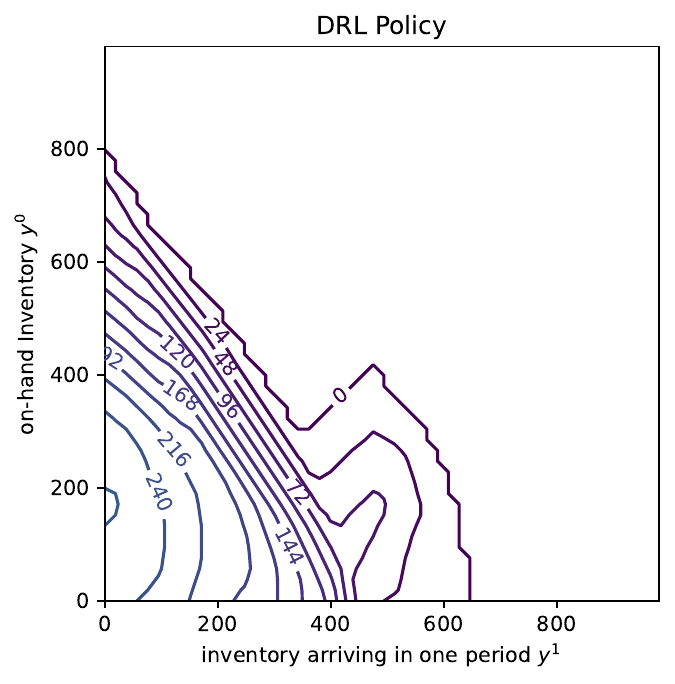}
\caption{Contour plots of the DRL and benchmark policies for a given product in the case of $L=2$ as a function of the endogenous state $\mathbf{y}=(y^0,y^1)$ at time $t=0$.}\label{fig:leadtime_policies}
\end{figure}

The results demonstrate that even in this case that is considered difficult, a simple unadorned RL agent is able to outperform established heuristics. We may then wonder what sort of policy the RL agent is learning, and in particular whether it satisfies the theoretical structural properties known of the optimal policy. Figure~\ref{fig:leadtime_policies} considers a single product in the case of a lead time of $L=2$. Its endogenous state is thus given by $\mathbf{y}_t=(y_t^0,y_t^1)$ where $y_t^0$ is the on-hand inventory at time $t$, and $y_t^1$ is the inventory to arrive in the next period $t+1$. For this product, we fix $t=0$, vary the values of $y_0^0$ and $y_0^1$, and record the ordering quantities produced by the different policies at each combination of $y_0^0$ and $y_0^1$. This allows us to draw contour plots of the policies, and observe how they evolve as a function of the inventory state.

The profiles of the contour plots of the heuristics ensue pretty naturally from their definitions. The simple base-stock policy does not differentiate between on-hand or pipeline units, which the vector base-stock policy does, providing tighter bounds on the ordering quantities. We recall that the vector base-stock policy represents an upper bound on the optimal policy, and this does seem to hold for the DRL policy, which is roughly encased within the contours of the base-stock one. Nonetheless, while the general shape of the policy is what we would expect of the optimal policy (similar plots of an optimal policy obtained throug Dynamic Programming can be found in \cite{maggiar2017joint}), it violates known theoretical structural properties of the optimal policy, which have been derived in \cite{morton1969bounds,zipkin2008structure}.  In particular, the order quantity should be monotonically decreasing as a function of both $y_t^0$ and $y_t^1$, which is not true everywhere in the graph.

\begin{figure}[htbp]
\centering
\includegraphics[scale=0.42]{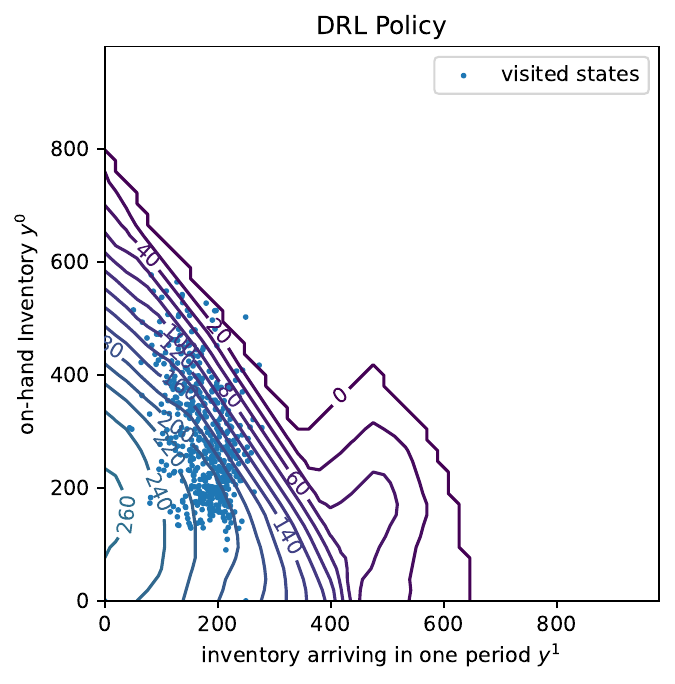}
\caption{Contour plot of the DRL policy overlaid with the visited states in the test rollout for the same example used in Figure~\ref{fig:leadtime_policies}.}\label{fig:leadtime_scatter}
\end{figure}

The fact that the DRL policy is somewhat lacking in some regions of the state space can partially be explained by the states it visits. We illustrate this in Figure~\ref{fig:leadtime_scatter} where we reproduce the same DRL policy as in Figure~\ref{fig:leadtime_policies}, but overlay the states visited during the test rollout. We observe, as would be expected in this stationary setting, that the visited states cluster around a particular region of the state space. As a result, in its training, the RL agent focuses on this region, and has little information outside of it where it need not generate good decisions since those states are seldom if ever visited. 

While it is not necessary in the stationary setting to learn good policies outside of those regions that are visited, it leaves us vulnerable to generating suboptimal decisions if we were ever to find ourselves outside of the stationary regime region. In a practical, possibly non-stationary setting, this could hurt the generalization of the learned policy. We return to this issue in Section~\ref{sec:sinn} where we propose methods to improve the structural properties and generalization properties of the DRL policies.

\begin{figure}[htbp]
\centering
\includegraphics[scale=0.33]{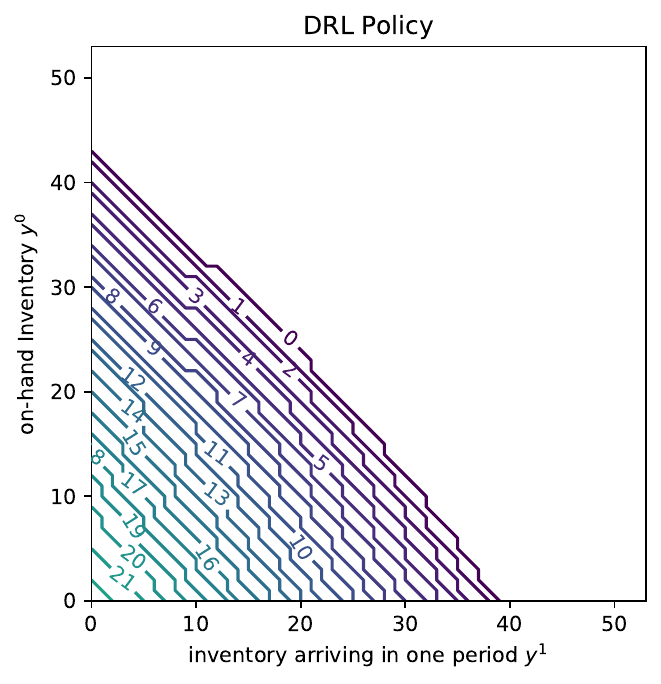}
\includegraphics[scale=0.33]{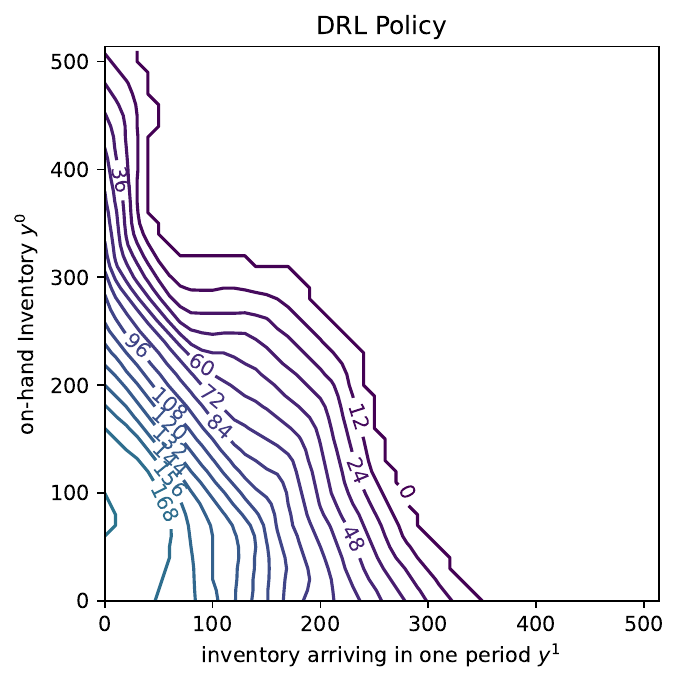}
\includegraphics[scale=0.33]{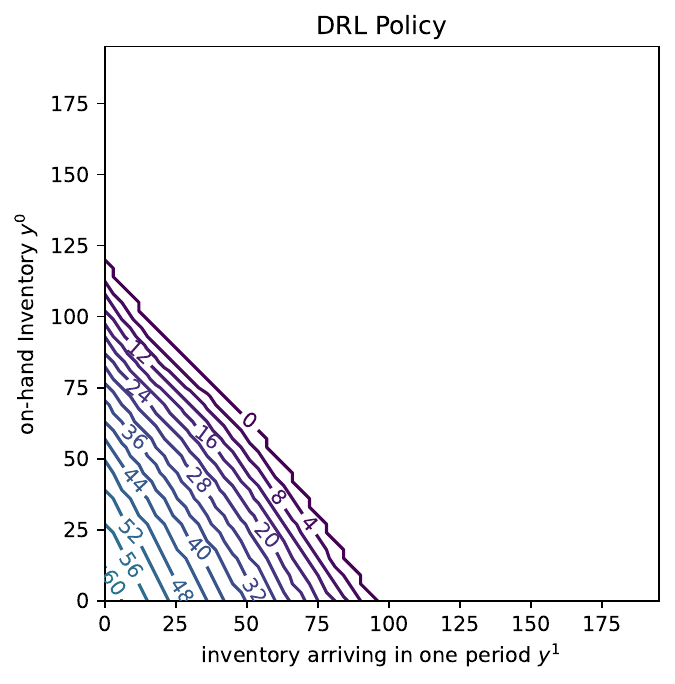}
\includegraphics[scale=0.33]{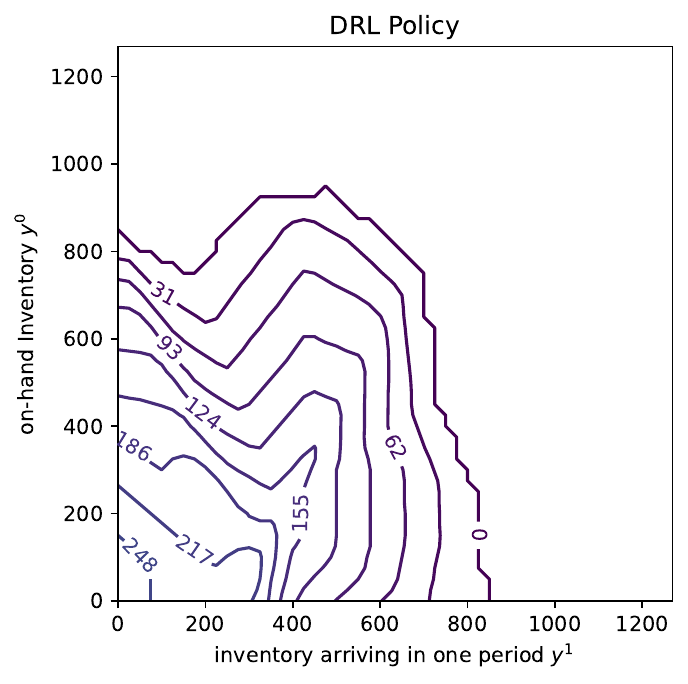}
\caption{Example contour plots of the DRL policy for a few given product in the case of $L=2$ as a function of the endogenous state $\mathbf{y}=(y^0,y^1)$ at time $t=0$.}\label{fig:leadtime_drl_examples}
\end{figure}

We show some additional contour plots of the DRL policy in Figure~\ref{fig:leadtime_drl_examples} for a few other products for $L=2$.

\begin{figure}[htbp]
\centering
\includegraphics[scale=0.33]{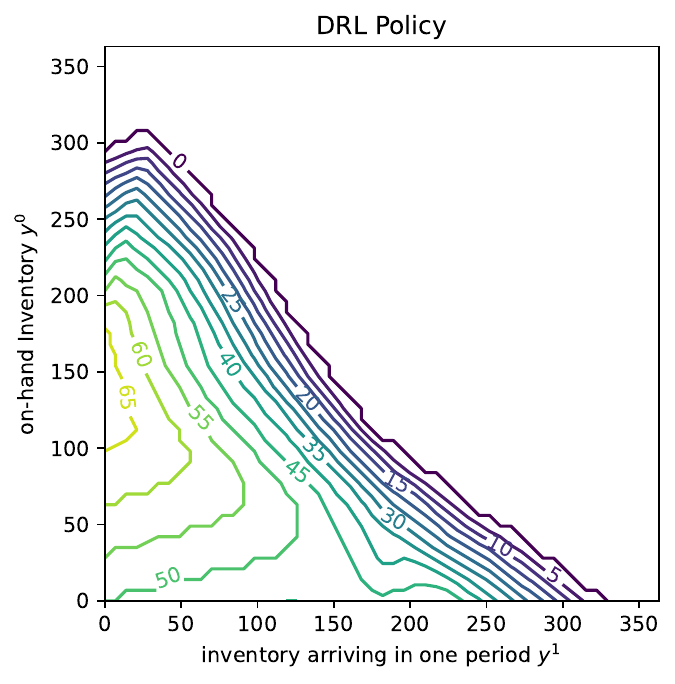}
\includegraphics[scale=0.33]{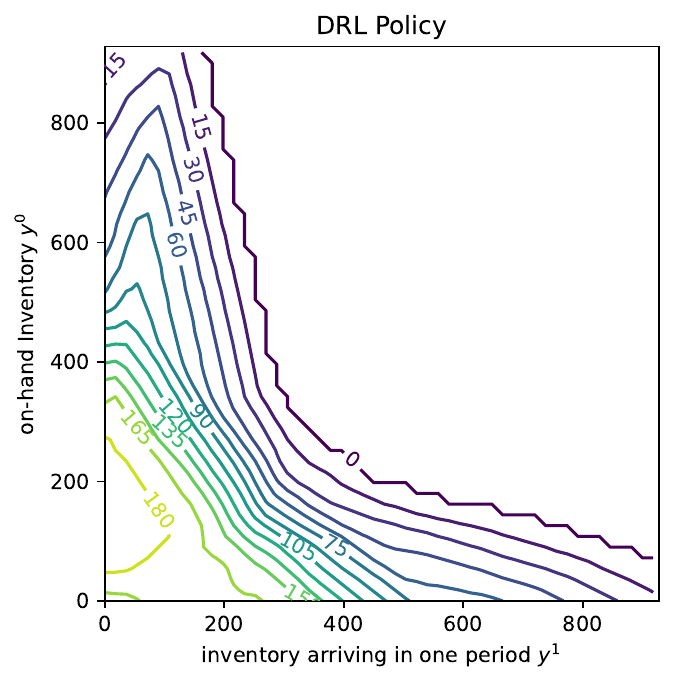}
\includegraphics[scale=0.33]{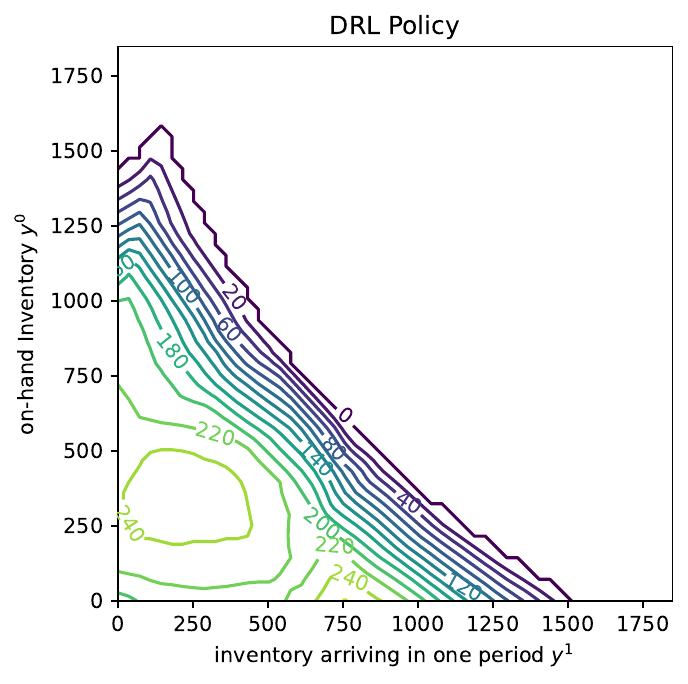}
\includegraphics[scale=0.33]{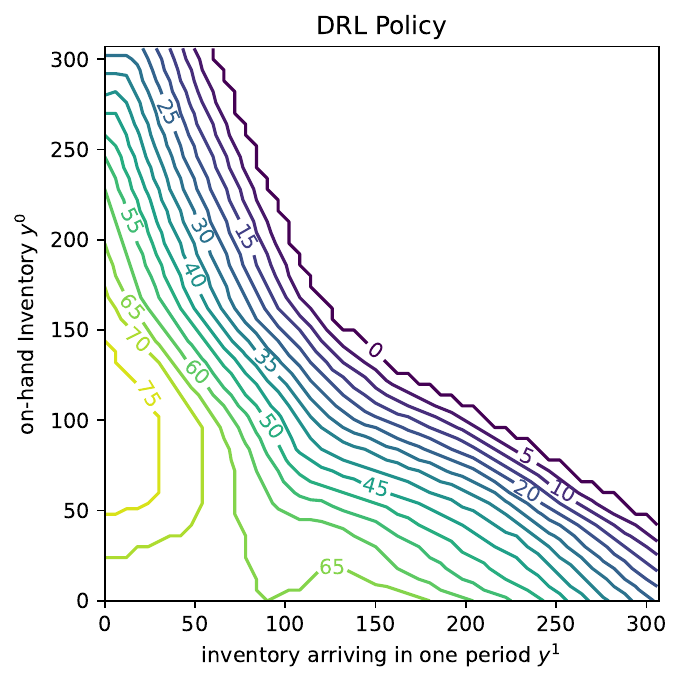}
\caption{Example contour plots of the DRL policy for a few given product in the case of $L=5$ as a function of the endogenous state $\mathbf{y}=(y^0,y^1,0,0,0)$ at time $t=0$.}\label{fig:leadtime_drl_examples_L_5}
\end{figure}

We used $L=2$ in all the examples above because we can plot the contour of of the policies across the entire endogenous state space since it has two dimensions. We can nonetheless plot similar graphs for larger lead times by fixing all but two dimensions of the state space. For example, we plot in Figure~\ref{fig:leadtime_drl_examples_L_5} contour plots of the DRL policy where we still vary $y^0$ and $y^1$ while fixing all other states to 0, i.e. we consider states of the form $\mathbf{y}=(y^0,y^1,0,0,0)$. This corresponds to a two-dimensional ``slice'' of the optimal policy, one that is likely to be outside of the region visited by the policy in the stationary regime. As a result, spurious behaviors are more frequent, further motivating the investigation of regularization techniques.

\subsection{Multi-Period Inventory Management with Perishable Products}\label{sec:perishable}

\subsubsection{Overview}\label{sec:perishable_overview}

A class of products that has received a sizeable amount of scrutiny is that of perishable products, i.e. products whose value vanishes at some expiration date, since managing such products is of significant importance in a number of industries, such as grocery, chemical,  pharmaceutical, or medical. The complexity in dealing with a finite shelf life of, say, $m$ periods, is that one needs to keep track of how many units are to expire in 0,\ldots,$m-1$ periods. The difficulty is compounded if we further include lead times, as above, or lost sales, or complex issuance policies. As a result, much of the literature and heuristics are devoted to the simpler case of null lead times,  backlogged demand, and FIFO issuance. It should nonetheless be noted that even though the complexity increases with longer shelf lives as the dimension of the problem grows, the problem gets closer and closer to the one with infinite shelf life, which only requires one dimensional endogenous state, the inventory level, leading most authors to focus on the cases $m=2,3$, and considering that larger shelf lives can essentially be treated as infinite\footnote{Another reason is that cases with $m=2,3$ remain solvable through a Dynamic Program, allowing for optimal solutions to be derived.} (\cite{nahmias1975comparison,nandakumar1993near}). The asymptotic optimality of base-stock policies has been proved not only for long shelf lives, but also in regimes where the penalty for large penalties for lost sales \cite{bu2023asymptotic}. 

The earlier work on perishable inventory management is due to \cite{nahmias1973optimal,nahmias1975comparison,nahmias1975optimal} for the backlogging case, and  \cite{fries1975optimal} in the case of lost sales, both highlighting the fact that an optimal policy depends on the full $m-1$ dimensional endogenous state space, and is thus numerically challenging to tackle. Much work has thus been done on devising heuristics and asymptotic results, just as in the case of multi-period management with lost sales. In those early studies, base-stock policies were shown to perform extremely well, and most heuristics adopt either a base-stock, or modified base-stock approach \cite{nahmias1975comparison,cohen1976analysis,nahmias1977higher,nandakumar1993near}.  These base-stock policies consider the total inventory position, while other heuristic differentiate between older and newer units, such as \cite{brodheim1975evaluation} who consider a policy that takes into account the amount of new inventory in the system. More intricate heuristics for the non-stationary case have also been proposed \cite{chao2015approximation}.

From a structural point of view, early monotonicity and sensitivity results were derived in \cite{nahmias1975optimal,fries1975optimal}, and then strengthened in \cite{li2014multimodularity,chen2014coordinating} using multimodularity and $L^\natural$-convexity.

Finally, we refer to \cite{Karaesmen2011} and \cite{nahmias2011perishable} for extensive treatments of inventory management of perishable products.

\subsubsection{MDP Formulation}

We consider a stationary multi-period problem with lost sales where products have a fixed shelf life of $m$ periods, at the end of which they perish (at no cost for ease of exposition). Similarly to the previous examples, we assume a revenue of $p$ per sale, a purchasing cost of $c$, holding cost $h$, and penalty for lost sale $b$. 

\paragraph{State}
The most obvious endogenous state would be one that tracks how many units have remaining shelf lives of 1, 2,..., $m-1$ periods, so that $\mathbf{y}_t=(y_t^1,\ldots,y_t^{m-1})$ where $y_t^k$ represents the number of units with $k$ periods of shelf life left. It is nonetheless equivalent and mathematically more convenient to introduce $\mathbf{w}_t=(w_t^1,\ldots,w_t^{m-1})$ and  let $w_t^k$ be the cumulative inventory with shelf life less than or equal to $k$, so that $w_t^{m-1}$ is the inventory level (sum of all units, regardless of remaining shelf life), and $w^{1}$ are the units to perish at the end of period $t$.  The number of units with shelf life $k$ can be found as $y_t^k=w_t^{k}-w_t^{k-1}$ (with the convention that $w_t^0=0$). The exogenous variables are then identical to the ones in the previous examples: the exogenous time series variables $\mathbf{x}_t$ consist in the previous $H$ demand realizations ($\mathbf{x}_t=(d_{t-H},\ldots,d_{t-1})$) and the exogenous static vector contains the economic parameters of a product, $\mathbf{s}=(p,c,h,b)$, where $p$ is the price, $c$ the purchasing cost, $h$ the holding cost, and $b$ the penalty for lost sale.

\paragraph{Action} The action is here again identical to the previous examples: the action $a_t=q_t$ at time $t$ is given by the number of units to purchase at time $t$, which will  arrive $L$ periods hence.

\paragraph{Transition Function}  In any period, the number of sold units $u_t$ is given by the minimum of on-hand inventory $w_t^{m-1}+q_t$ (to account for the newly purchased units) and realized demand $d_t$, i.e. $u_t=\min(w_t^{m-1}+q_t, d_t)$; while the number of units that perish $v_t$ is obtained as the number of units remaining from the first bucket $w_t^1$ after sales, if any: $v_t=\max(w_t^1-d_t,0)$. The transition from one period to the next then shifts all remaining units down the endogenous state vector:
\begin{align*}
w_{t+1}^k&= \begin{cases}
w_{t}^{k+1}-u_t-v_t,&k=1,\ldots,m-2\\
w_{t}^{m-1}+q_t-u_t-v_t,& k=m-1
\end{cases}.
\end{align*}

\paragraph{Reward Function}
The reward is here again similar to the ones we have seen previously. We receive a revenue $p$ per sale, and incur costs $c$ per purchased unit, $b$ per lost sale, and a holding cost $h$ per unit left over at the end of the period, including those that perished during the period.
\begin{align*}
R_t &= p \min(d_t, w_t^{m-1}+q_t) - c q_t - b \max(d_t - w_t^{m-1} - q_t, 0) - h \max(w_t^{m-1} + q_t - d_t , 0).
\end{align*}

\subsubsection{Benchmarks}\label{sec:perishable_benchmark}

We mentioned a number of heuristics in Section~\ref{sec:perishable_overview}, the vast majority of which take the form of base-stock policies where the order-up-to level is adjusted to reflect the risk of the units perishing. However, these heuristics are often unwieldy as they require that some numerical equation be solved. As a result, we consider two base-stock benchmarks that prove easier to implement: 1) a standard base-stock policy that uses the same order-up to level as in the non-perishing case, and 2) the best base-stock policy for the products in the test set. The latter is obtained by running a golden-section search \citep[ch.8]{luenberger2016linear} on each product to maximize its own expected average return and find its best order-up-to level, a process that is easily vectorized, using 0 and the standard base-stock level as initial lower and upper bounds.

We note that the first benchmark is asymptotically optimal as the shelf life increases, and in particular, this benchmark policy should converge to the expected reward found in Section~\ref{sec:lost_sales}. The second policy, while not technically selected among the a set of policies that include the optimal one, should perform very well since all numerical results in the literature point to base-stock policies as being close to optimal in the perishable case, and we are choosing the best among them. We thus expect to outperform the first, naive, policy for short shelf lives, but underperform it for larger shelf lives as the problem becomes similar to the regular multi-period problem with lost sales. Similarly, the DRL policy is likely to underperform the second benchmark, which will be close to optimal.

\subsubsection{Experiments}

\begin{figure}[htbp]
\centering
\includegraphics[scale=0.4]{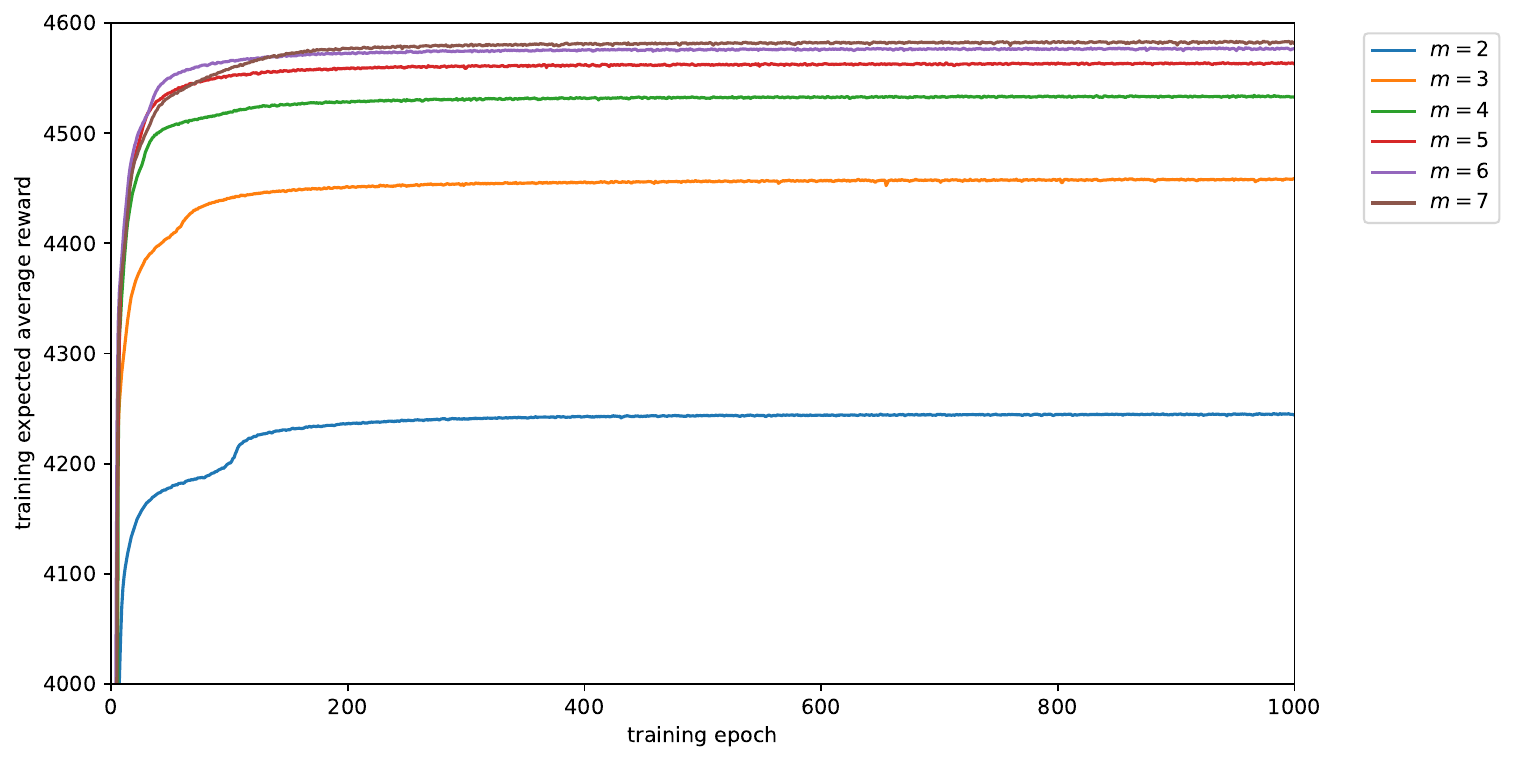}
\caption{Training curves of the DRL policy for different values of the shelf life.}\label{fig:perishable_training}
\end{figure}

We trained DRL policies for shelf lives ranging from 2 to 7 using the default settings presented in Table~\ref{tab:hyperparams}. Figure~\ref{fig:perishable_training} shows the expected average reward during training for the different values of the shelf life, and we observe how the marginal difference of increasing the shelf life quickly decreases so that there is little impact beyond a shelf life of about 7 periods.

\subsubsection{Results}

\begin{figure}[htbp]
\centering
\includegraphics[scale=0.5]{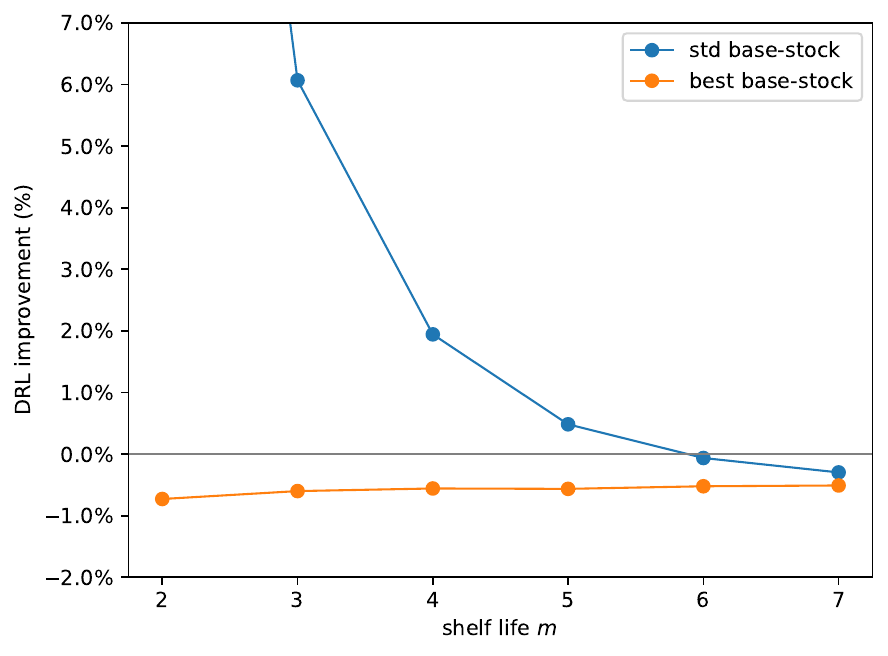}
\caption{Relative improvement of the DRL policies over the benchmarks in the perishable case, as a function of the shelf life $m$.}\label{fig:perishable_results}
\end{figure}

The results of the policies, DRL-learned and benchmarks, on the test set are presented in Table~\ref{tab:perishable_results} and Figure~\ref{fig:perishable_results}. The results follow the intuition, with the DRL policies vastly outperforming the naive standard base-stock for short shelf lives, but losing its edge for longer shelf lives, where the behavior of the system becomes closer and closer to that of the infinite shelf life case.  We nonetheless also see that the DRL policy's performance is very close to that of the best base-stock policy for all shelf lives, and that the gap quickly converges to the one observed in the case of the infinite shelf life in Table~\ref{tab:basic_default_results}.

\begin{table}[htbp]
\centering
\caption{Results of the DRL and benchmark policies on the case with perishable units for different shelf lives. The percentages are the relative improvements of the DRL policy against the benchmarks.}\label{tab:perishable_results}
\begin{tabular}{crrrrr}
\toprule
 & \multicolumn{1}{c}{DRL} & \multicolumn{2}{c}{std base-stock} & \multicolumn{2}{c}{best base-stock} \\
 \cmidrule(lr){2-2} \cmidrule(lr){3-4} \cmidrule(lr){5-6}
$m$ & reward & reward &  & reward & \multicolumn{1}{c}{} \\
\midrule
2 & 4,177.26 & 3,392.30 & 23.14\% & 4,207.92 & -0.73\% \\
3 & 4,397.65 & 4,146.07 & 6.07\% & 4,424.21 & -0.60\% \\
4 & 4,481.19 & 4,395.73 & 1.94\% & 4,506.33 & -0.56\% \\
5 & 4,515.29 & 4,493.55 & 0.48\% & 4,540.90 & -0.56\% \\
6 & 4,531.99 & 4,534.85 & -0.06\% & 4,555.77 & -0.52\% \\
7 & 4,539.30 & 4,552.84 & -0.30\% & 4,562.53 & -0.51\% \\
\bottomrule
\end{tabular}
\end{table}

\begin{figure}[htbp]
\centering
\includegraphics[scale=0.6]{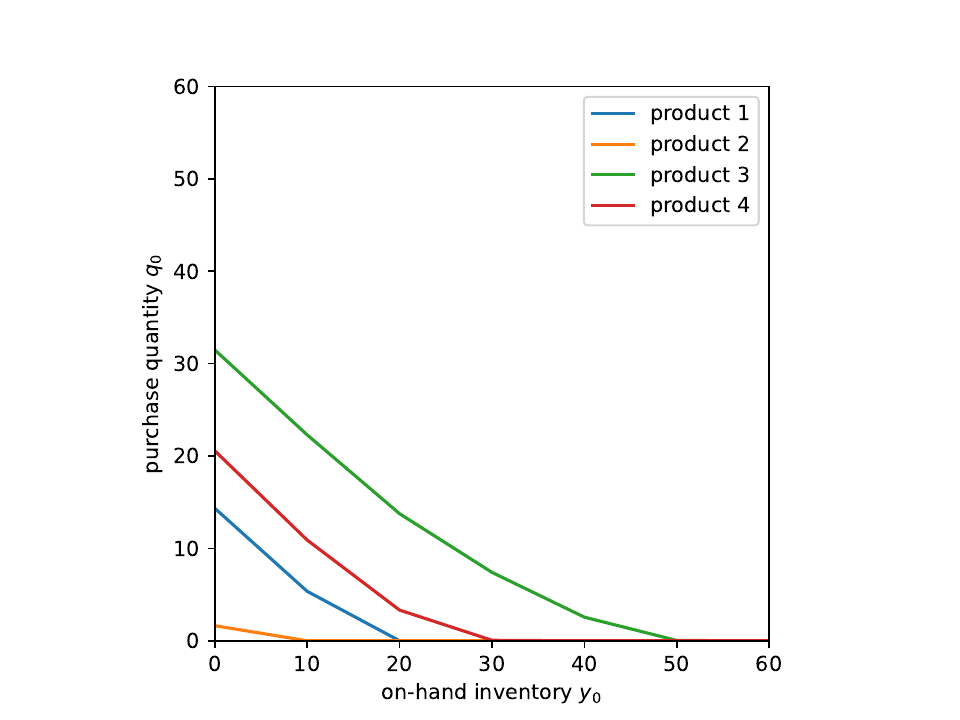}
\caption{Actions (purchase quantity) learned by the DRL policy as a function of the inventory level at time $0$ for four different products for a shelf life of $m=2$.}\label{fig:perishable_structure_m_2}
\end{figure}

We can next consider the structure of the optimal  policies. We recall that even in the case of a shelf life of $m=2$, where the problem is single-dimensional, there is no simple closed-form solution to the optimal policy, which is known to not be of the base-stock type, even in the backlogging case. We plot in Figure~\ref{fig:perishable_structure_m_2} the purchase order quantity for a few products at time 0 as  function of the on-hand inventory (which is to perish at the end of the period). We observe that the policy learned by the DRL agent is indeed not of the base-stock type since it is not a simple line with slope of -1. It does however seem to satisfy the requirement that the purchase quantity is monotonically decreasing as a function of the on-hand inventory with a slope between -1 and 0, which is one of the known structural results.

\begin{figure}[htbp]
\centering
\includegraphics[scale=0.33]{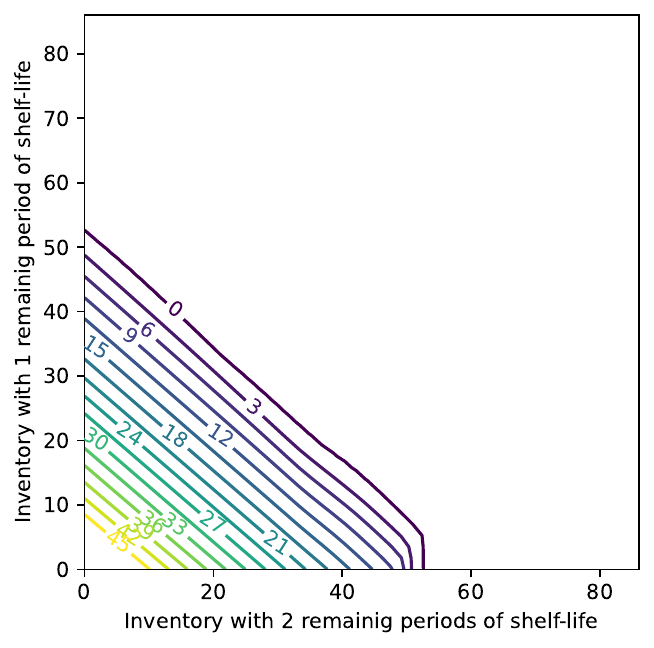}
\includegraphics[scale=0.33]{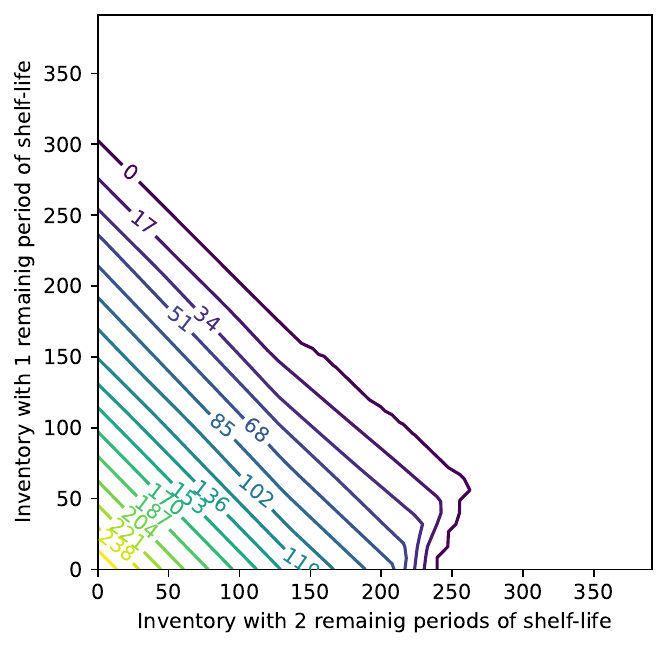}
\includegraphics[scale=0.33]{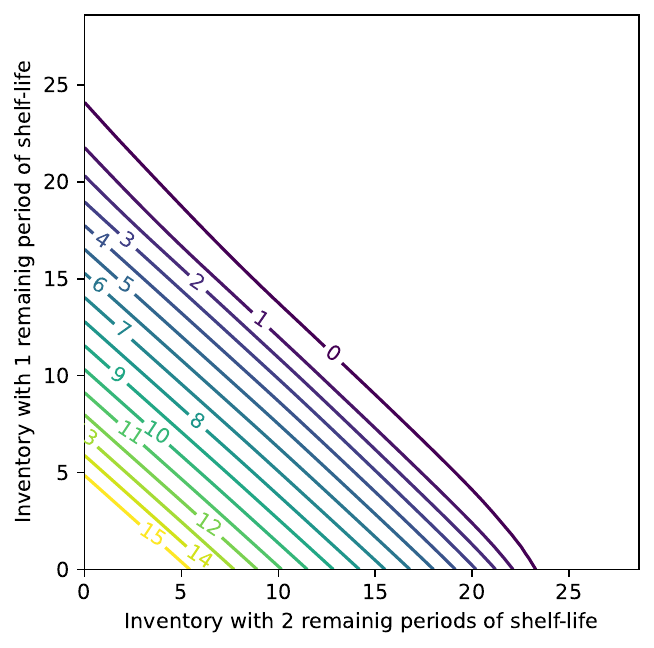}
\includegraphics[scale=0.33]{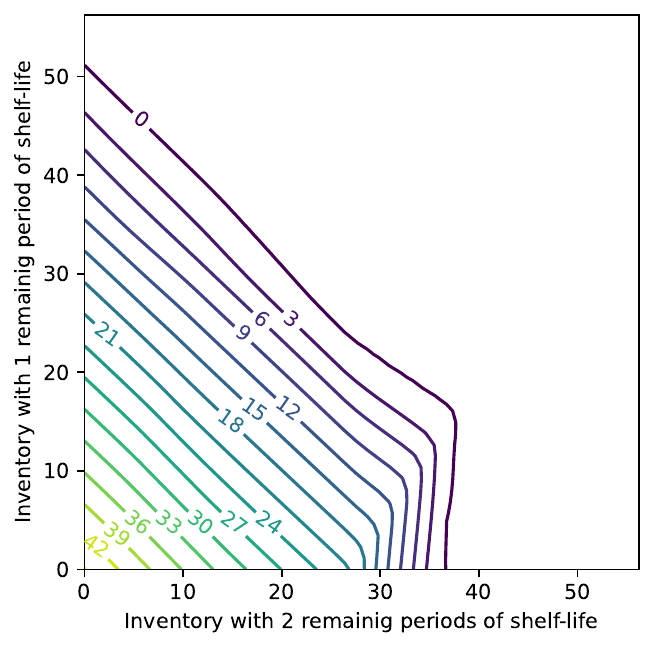}
\caption{Example contour plots of the DRL policy for a few given product in the case of $m=3$ as a function of the number of on-hand units set to perish at the end of the period, and at the end of the following period, at time $t=0$.}\label{fig:perishable_structure_m_3}
\end{figure}

We similarly show in Figure~\ref{fig:perishable_structure_m_3} the contour plots of the purchase quantity in the case of $m=3$, which leads to a two-dimensional endogenous state space. Here we display the contour plot as a function of the number of on-hand units set to perish at the end of the period, and at the end of the following period.  We observe here again that the policies seem to abide by the structural properties expected of the optimal policy. In particular, we also note that the policy does seem to be close to a base stock policy in much of the endogenous state space. However, just as in the case of lost sales with lead times in Section~\ref{sec:lost_sales_lead_times}, in the case of some products, the learned policy displays some behavior at odds with the optimal policy, such as a lack of monotonicity through the endogenous state space.

\subsection{Multi-Period Inventory Management with Dual Sourcing}\label{sec:dual_sourcing}

\subsubsection{Overview}\label{sec:dual_overview}

Dual sourcing is a strategy that involves procuring inventory for the same product from two suppliers who differ in their lead times and costs. Typically, these suppliers are labeled as regular and expedited, the regular source being cheaper but requiring a longer lead time, while the expedited source is able to deliver the inventory in a shorter lead time, but incurs a premium.  It is an interesting problem from the perspective of testing our DRL approach since, as opposed to the previous examples, it now involves a two-dimensional action in the form of both regular and expedited order quantities. There is an interplay between these two decision that is further complicated by our focus on a lost sales formulation, which is already considered as especially difficult in the single source case.

The study of this tradeoff between speed and cost has a long history in the inventory management literature, with some of the earliest work in the multi-period setting proposed by \cite{daniel1961delivery}. Assuming backlogging of missed sales, \cite{fukuda1964optimal} considers the special case of an expedited lead time of 0, and a regular lead time of 1, and proves in this case that the optimal policies are base-stock. The case of arbitrary fixed lead time differences between the expedited and regular lead times is studied in \cite{whittemore1977optimal} who show that for non-consecutive lead times, the optimal policy ceases to have a simple base-stock structure and depends on the full state of pipeline inventory. \cite{whittemore1977optimal} also extend the results of \cite{fukuda1964optimal} on consecutive lead times by allowing the expedited source to have an arbitrary  lead time of $L_e>0$ (still letting the regular lead time be $L_r=L_e+1$). These base-stock results on consecutive lead times have been extended to more than two sources in \cite{feng2006base}. A connection between dual sourcing systems with backlogging, and single-sourcing systems with lead times and lost sales is established in \cite{sheopuri2010new}, and additional structural properties of the optimal policy are derived in \cite{li2014multimodularity} using multimodularity, and  \cite{hua2015structural} using $L^\natural$-convexity. We should note that all the above references consider the case of a backlogged system, as opposed to our focus on lost sales systems.

Given the difficult nature of the dual sourcing problem, even in the backlogged case, much work has focused on the derivation of heuristics. The main heuristics in dual sourcing include the Single-Index policy of \cite{scheller2007effective}, the Dual-Index policy \cite{veeraraghavan2008now}, Tailored Base-Surge policy (TBS) \cite{allon2010global}, Capped Dual-Index (CDI) policy \cite{sun2019robust}, Best Weighted Bounds (BWB) policy \cite{hua2015structural}, and a more recent Projected Expedited Inventory Position (PEIP) policy \cite{drent2022effective}. These all assume backlogged sales. Their asymptotic properties and optimality have also been the subject of research, notably in \cite{xin2018asymptotic} who analyze the asymptotic behavior of the TBS policy as the lead time of the regular source grows large, and \cite{drent2022effective} who consider the optimal asymptotic behavior of the PEIP policy as the shortage cost and cost premium of the expedited source grow large.

We refer to \cite{xin2023dual} for a recent review of dual sourcing models.

\subsubsection{MDP Formulation}

We consider a stationary multi-period problem with lost sales where in each period, purchase orders are placed to the expedited and regular sources, to arrive $L_e$ and $L_r$ periods hence ($L_e<L_r$), respectively, at costs $c_e$ and $c_r$ ($c_e>c_r$). Similarly to the previous examples, we assume a revenue of $p$ per sale, a holding cost $h$, and penalty for lost sale $b$. 

\paragraph{State} 
The endogenous state $\mathbf{y}_t$ tracks the inventory in the system. $y_t^0$ represents the inventory on-hand, while $y_t^k$ for $k=0,\ldots,L_r$ represents the number of units to arrive in $i$ periods. As in earlier examples, the exogenous time series variables $\mathbf{x}_t$ consists of the previous $H$ demand realizations ($\mathbf{x}_t=(d_{t-H},\ldots,d_{t-1})$), and the exogenous static vector contains the economic parameters of a product, $\mathbf{s}=(p,c_e,c_r,h,b)$, where $p$ is the price, $c_e$ and $c_r$ the expedited and regular purchasing cost, $h$ the holding cost, and $b$ the penalty for lost sale.

\paragraph{Action} The action produced by the policy is here two dimensional $\mathbf{a}_t=(q_t^e, q_t^r)$ and represents the purchase quantities associated with the expedited ($q_t^e$) and regular ($q_t^r$) sources placed in period $t$.

\paragraph{Transition Function} 
The purchase orders are placed at the beginning of the period, following which the demand realization is observed, and units left over after sales carry over to the next periods. To facilitate the expression of the transition function, we first define $\mathbf{\tilde{y}}_t$, the updated endogenous state after the expedited order is placed:
\begin{align*}
\tilde{y}_t^k &=
\begin{cases}
y_t^k,&k\neq L_e\\
y_t^k+q_t^e,&k= L_e
\end{cases}.
\end{align*}
Then, we can express the transition function as:
\begin{align*}
y_t^k &=
\begin{cases}
\max(\tilde{y}_t^0-d_t,0)+y_t^1,&k=0\\
\tilde{y}_t^{k+1},&0<k<L_R-1\\
q_t^r,& k=L_R-1
\end{cases}.
\end{align*}
The transition function describes the dynamics where pipeline units move towards the facility, while inserting the newly purchased units in the right place.

\paragraph{Reward}
The reward is similar to the ones used in the previous examples, with the main difference being that we now need to differentiate between orders placed to the expedited and regular sources. Otherwise, we still receive a revenue $p$ per sale, and incur a penalty $b$ per lost sale, and holding cost $h$ per unit left over at the end of the period.
\begin{align*}
R_t &= p \min(d_t, \tilde{y}_t^0) - c_r q_t^r - c_e q_t^e - b \max(d_t - \tilde{y}_t^0, 0) - h \max(\tilde{y}_t^0 - d_t , 0).
\end{align*}

\subsubsection{Benchmark}

\begin{figure}[htbp]
\centering
\includegraphics[scale=0.8]{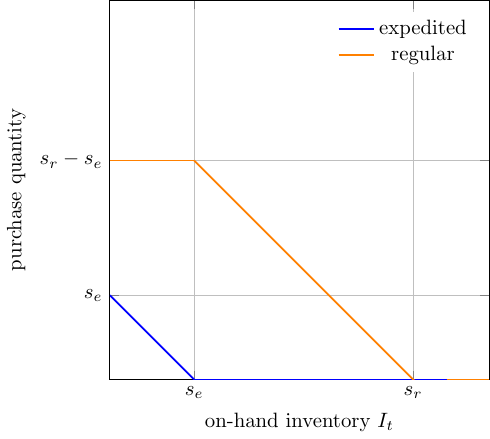}
\caption{Illustration of the single-index, dual base stock policy.}\label{fig:single_index}
\end{figure}

We mentioned a number of heuristics in Section~\ref{sec:dual_overview}, including Single Index (SI), Dual Index (DI), Capped Dual-Index (CDI), Tailored Base-Surge (TBS), Best Weighted Bounds (BWB), and Projected Expedited Inventory Position (PEIP). We recall that these were derived for systems assuming backlogged sales, while we assume lost sales, and many of these can also be cumbersome to implement. We can still devise simple and sensible benchmark policies by taking inspiration from those. In particular, we choose to implement a Single Index Dual Base-Stock policy, which is optimal in the case of consecutive lead times ($L_R=L_E + 1$) studied in \cite{fukuda1964optimal,whittemore1977optimal}. This policy tracks a single index, namely the inventory position of the system given by the sum of on-hand and all pipeline inventory, $I_t = \sum_{k=0}^{L_R-1} y_t^k$, and uses two base-stock levels $s_e$ and $s_r$ for the expedited and regular sources, respectively. It first applies the expedited base-stock policy, raising the inventory position to $s_e$ if necessary, and then applies the regular base-stock level policy, raising the inventory position resulting from the expedited base-stock policy to $s_r$. In other words, the purchase order quantities $q_t^e$ and $q_t^r$ from the expedited and regular sources in period $t$ are given by:
\begin{align*}
q_t^e&= \max(s_e - I_t, 0),& q_t^r&= \max(s_r - (I_t + q_t^e), 0).
\end{align*}
The policy is illustrated in Figure~\ref{fig:single_index}, where we see that for inventory levels $I_t$ below the expedited base-stock $s_e$, we purchase just enough to raise the inventory position to $s_e$, and then order a constant number of units $s_r-s_e$ in order to raise the resulting inventory position to $s_r$. For values of the inventory levels between $s_e$ and $s_r$, we purchase enough units to raise the inventory position to $s_r$. For values of $I_t$ greater than or equal to $s_r$, we do nothing.

Additionally, In the case of consecutive lead times with backlogging, the order-up to level for the expedited source admits a closed-form expression as \cite{fukuda1964optimal,whittemore1977optimal,xin2023dual}:
\begin{align}
s_e &= F_{L_{E}+1}^{-1}\left(\frac{c_u}{c_u+c_o}\right),\label{eq:dual_outl}
\end{align}
where $F_{l}$ is the cumulative distribution function of the demand over $l$ periods, $c_u$ is the underage cost with  $c_u=b-(c_e-c_r)$, and $c_o$ is the underage cost given by $c_o=h$. Note that this underage cost is the reason why we generated the regular cost the way we did, so as to guarantee that $c_u\geq 0$, a condition highlighted in \cite{whittemore1977optimal}. Additionally, in the case of lost sales, the underage cost is redefined as $c_u=p-c_e + b - (c_e-c_r)$.

We will implement a single index dual base-stock policy using Equation~\eqref{eq:dual_outl} to set the base-stock level $s_e$ of the expedited source, and will then find the optimal regular source base-stock level $s_r$, given this expedited base-stock level, using a golden section search as we did in the case of the best base-stock level for perishable units in Section~\ref{sec:perishable_benchmark}.

\subsubsection{Experiments}

We conduct two sets of experiments for the dual sourcing problem. In the first one, we depart momentarily from the lost sales setting on which we have focused so far, and consider the case of consecutive lead times with $L_E=0$ and $L_R=1$ under a backlogging assumption. The motivation to study this specific example is that it admits a known optimal policy in the form of a single index, dual base-stock policy \cite{fukuda1964optimal,whittemore1977optimal,xin2023dual}. This allows us to compare both the expected average reward obtained by the DRL policy to the true optimal, but also to investigate whether the DRL policy possesses the expected structural properties.

In the second set of experiments, we return to a lost sales assumption and apply DirectBackprop to learn DRL policies for several choices of expedited and regular lead times, with a fixed expedited lead time of 2, and regular lead times ranging from 4 to 9.

\subsubsection{Results}

\paragraph{Consecutive Lead Times ($L_E=0$, $L_R=1$)}
We trained a DRL policy in the consecutive lead time setting with $L_E=0$, $L_R=1$, and assuming backlogged sales. As mentioned above, our benchmark is optimal for this usecase. The results of the DRL and the optimal policy are presented in Table~\ref{tab:dual_consecutive_results}, and we observe that the DRL policy is able to learn a policy that is competitive with the optimal policy, which we recall is derived with additional knowledge than that which the DRL policy has access to.

\begin{table}[htbp]
\centering
\caption{Results of the DRL and the optimal benchmark single-index, dual base-stock policies on the consecutive lead times ($L_E=0$, $L_R=1$) case.}\label{tab:dual_consecutive_results}
\begin{tabular}{lcc}
\toprule
 & reward & gap \\
\midrule
Optimal & 5,049.75 & - \\
DRL & 5,031.66 & -0.36\% \\
\bottomrule
\end{tabular}
\end{table}

\begin{figure}[htbp]
\centering
\includegraphics[scale=0.33]{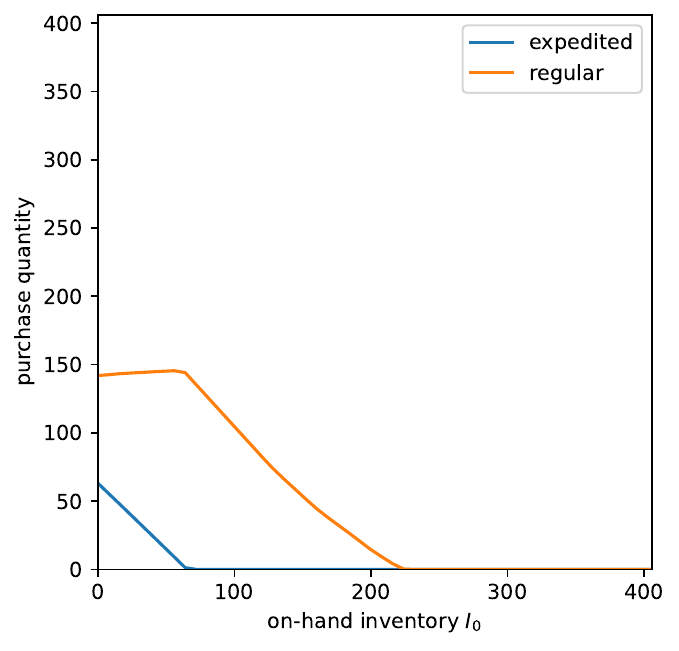}
\includegraphics[scale=0.33]{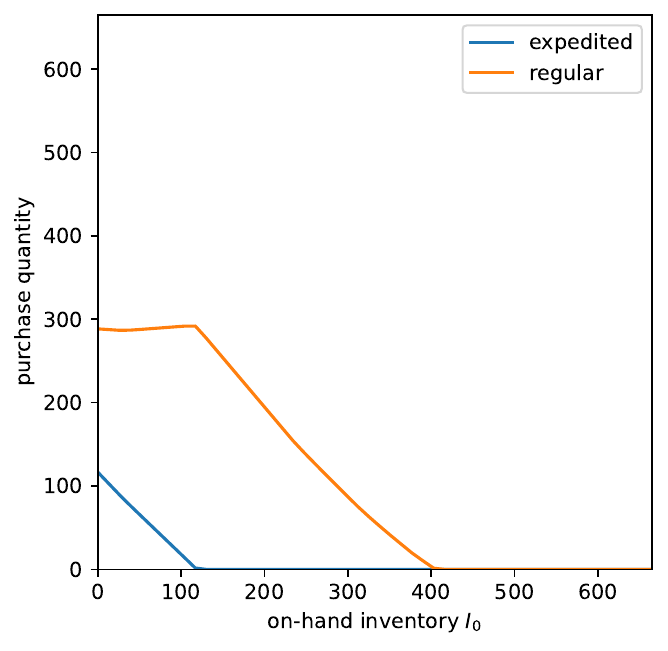}
\includegraphics[scale=0.33]{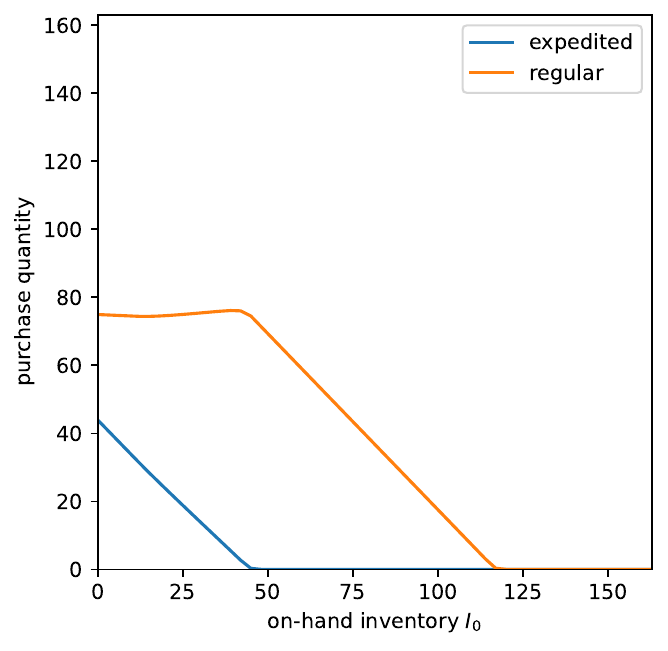}
\includegraphics[scale=0.33]{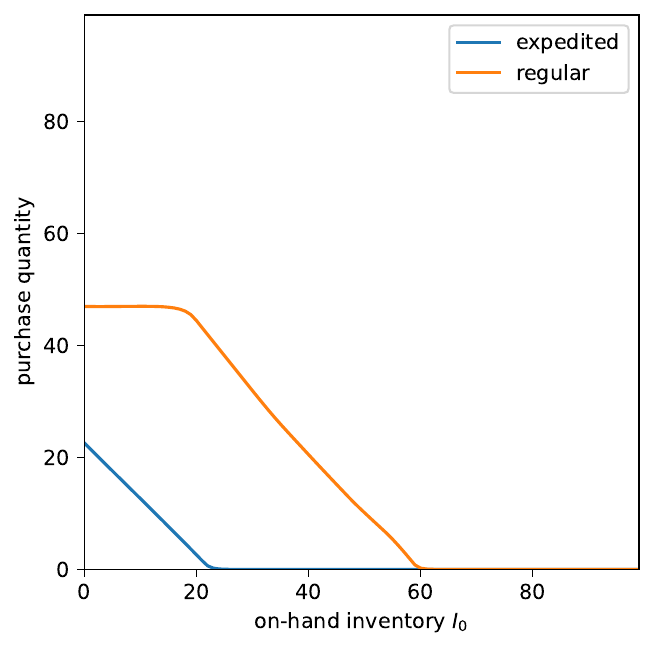}
\caption{Example plots of the expedited and regular orders produced by the DRL policy for a few given products in the case of consecutive lead times ($L_E=0$, $L_R=1$) as a function of the initial on-hand inventory level.}\label{fig:dual_consecutive_policy}
\end{figure}

A secondary question is whether, and how much, the DRL policy is able to learn the structural properties of the optimal policy. Figure~\ref{fig:dual_consecutive_policy} shows the actions produced by the DRL policy for a few products as a function of the endogenous state $y_t^0$ at some arbitrary period ($t=0$). Remarkably, the DRL agent is able to learn a structure that is very close to the optimal one presented in Figure~\ref{fig:single_index}, and is able to pick up the interactions between expedited and regular orders.

\paragraph{Non-Consecutive Lead Times and Lost Sales}

We now restore the lost sales assumption and consider settings where the expedited lead time is of 2 periods, while the regular lead time ranges from 4 to 9. These are high-dimensional problems that cannot possibly be tackled with traditional methods such as Dynamic Programming. We present in Table~\ref{tab:dual_sourcing_results} the results obtained by the learnt RL policies, and the ``Fukuda'' benchmark. We recall once again that while the heuristic is not optimal in these settings, it nonetheless has access to more information since it knows the underlying demand distributions, while the RL agent only has access to the previous 32 observations in each period.

\begin{table}[htbp]
\centering
\caption{Expected average reward for the benchmark and DRL policies for a fixed expedited lead time of 2 and varying regular lead times.}\label{tab:dual_sourcing_results}
\begin{tabular}{lrrrrrr}
\toprule
& \multicolumn{6}{c}{$L_r$}\\
 & 4 & 5 & 6 & 7 & 8 & 9 \\
\midrule
Benchmark & 4,607.45 & 4,592.14 & 4,562.46 & 4,534.27 & 4,514.41 & 4,499.73 \\
DRL  & 4,591.16 & 4,584.27 & 4,583.39 & 4,576.69 & 4,579.15 & 4,575.35 \\
\midrule
DRL Improvement & -0.35\% & -0.17\% & 0.46\% & 0.94\% & 1.43\% & 1.68\% \\
\bottomrule
\end{tabular}
\end{table}

\begin{figure}[htbp]
\centering
\includegraphics[scale=0.5]{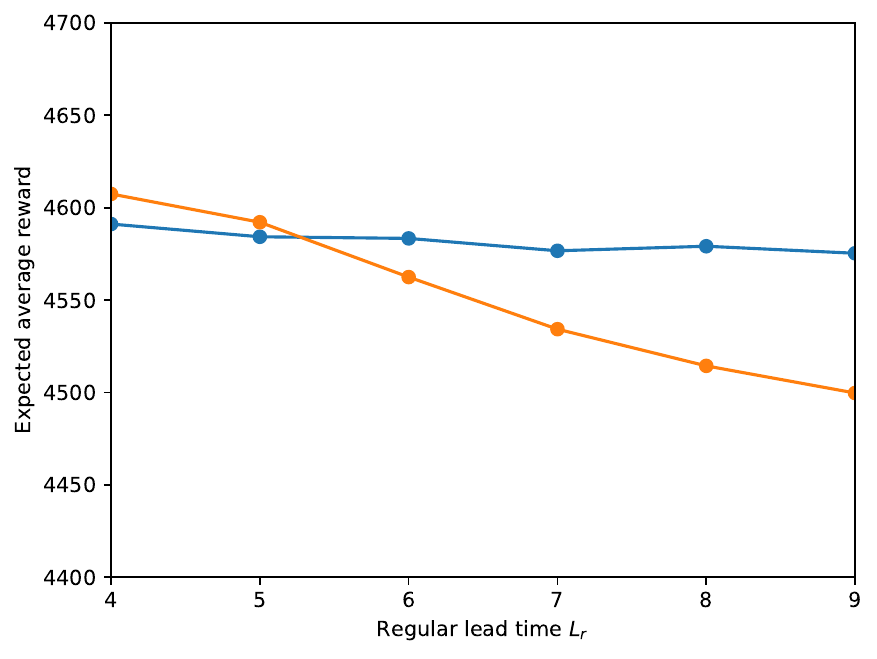}
\includegraphics[scale=0.5]{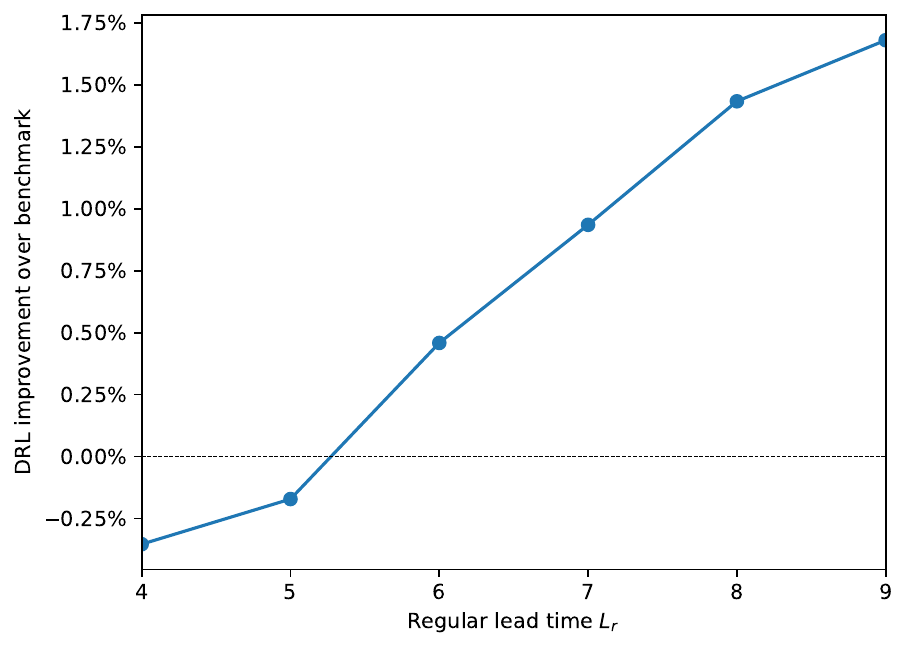}
\caption{Comparison of the performance of the DRL policies against the Single Index Dual Base-Stock benchmark for  $L_e=2$ and varying values of $L_r$.}\label{fig:dual_sourcing_results}
\end{figure}

The results are further plotted in Figure~\ref{fig:dual_sourcing_results}, where we observe that the DRL policy quickly outperforms the single-index dual base-stock benchmark as the regular lead time increases.

\begin{figure}[htbp]
\centering
\includegraphics[scale=0.45]{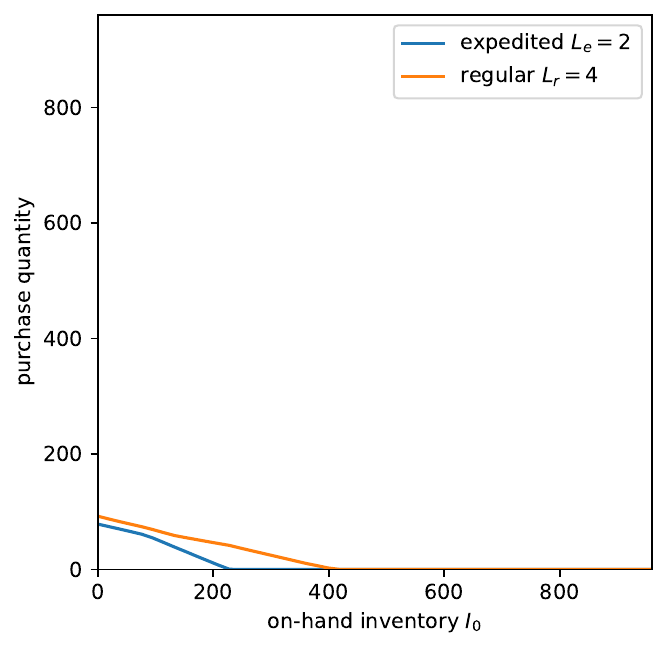}
\includegraphics[scale=0.45]{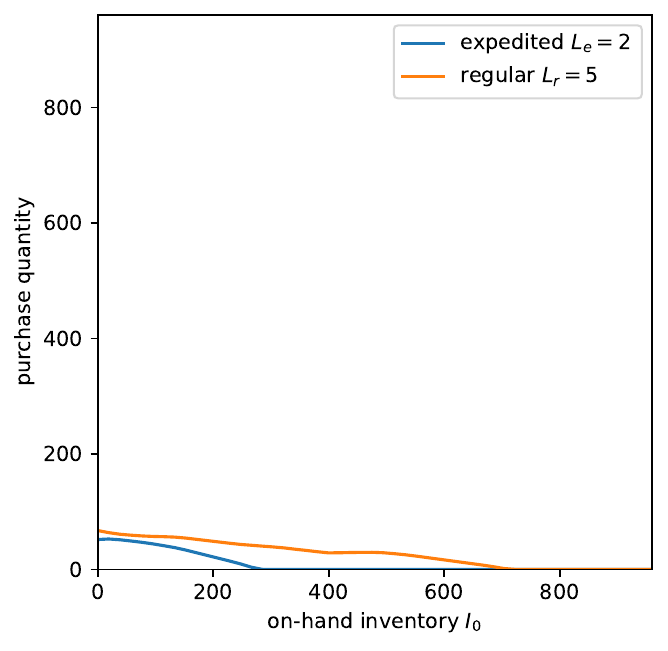}
\includegraphics[scale=0.45]{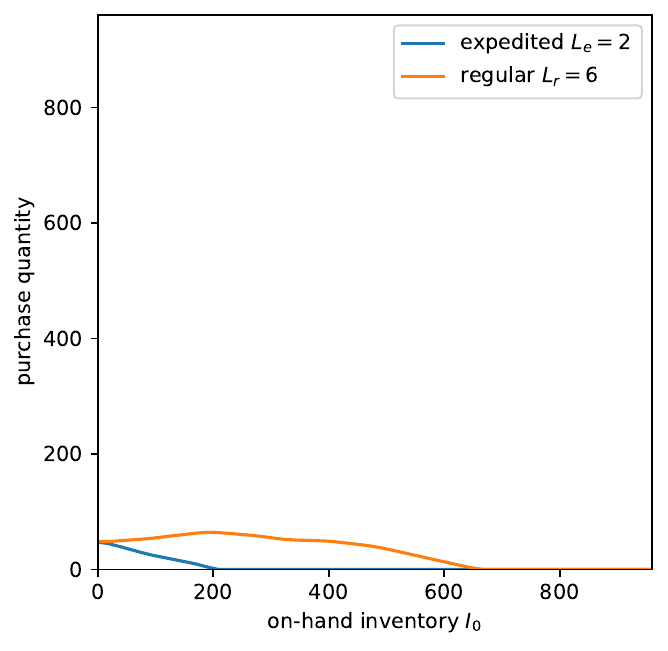}
\includegraphics[scale=0.45]{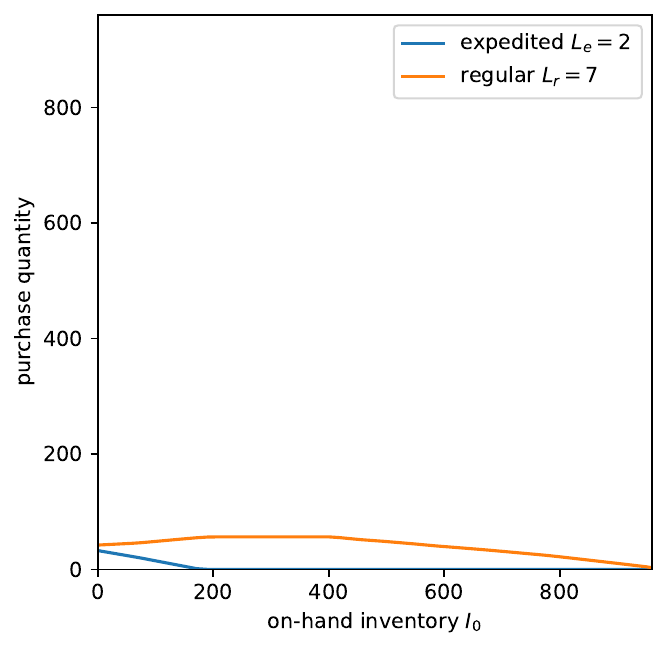}
\includegraphics[scale=0.45]{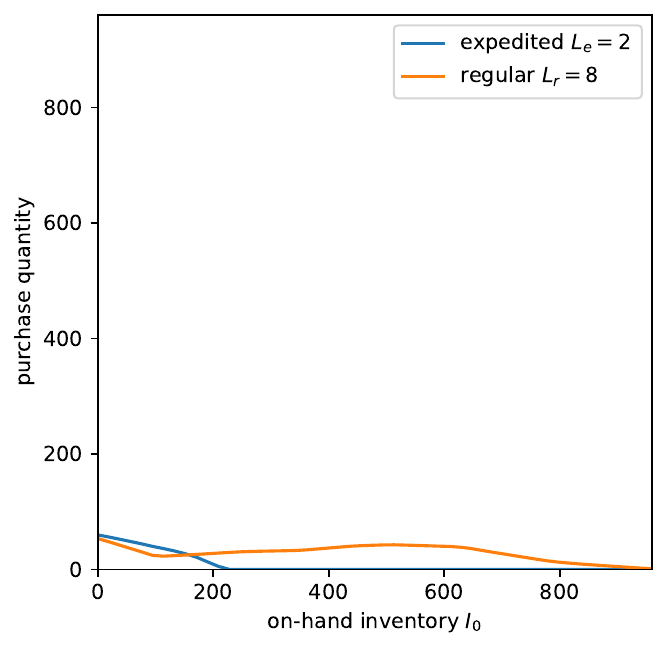}
\includegraphics[scale=0.45]{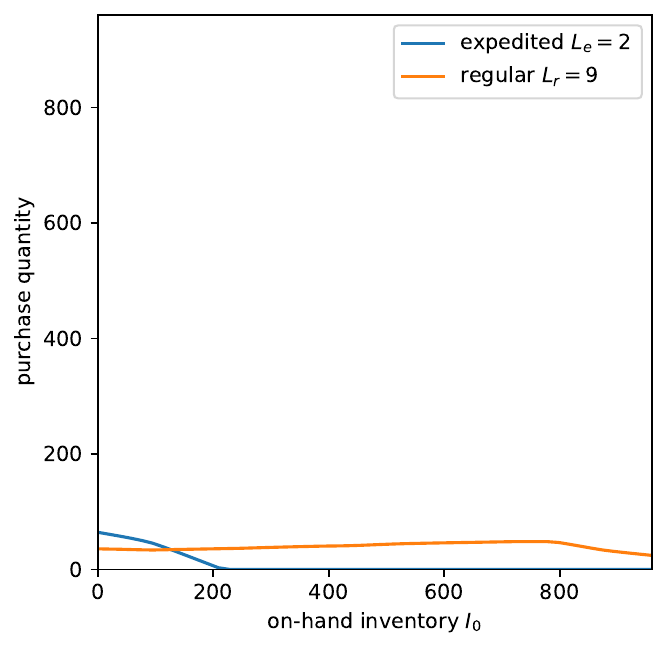}
\caption{Ordering policy for a given product at a given time, for endogenous state variables of the form $\mathbf{y}_t=(y_t^0,0,\ldots,0)$, for various values of $L_r$.}\label{fig:dual_sourcing_example}
\end{figure}

Given the high-dimensionality of the problem, it is not easy to visualize the policies learnt by the DRL agent. To circumvent this difficulty, we consider a given product in a given time period, say 0. We thus fix its static and time series variables $\mathbf{s}$ and $\mathbf{x_t}$, respectively, and consider endogenous variables $\mathbf{y}_t$ of the form $\mathbf{y}_t=(y_t^0,0,\ldots,0)$. In other words, we consider an inventory state where we vary the amount of on-hand inventory, and fix the rest of the pipeline to 0. We then plot the resulting expedited and regular orders as a function of $y_t^0$. We plot the resulting policy for increasing values of $L_r$ and a fixed value of $L_e=2$. Note that these are unlikely states that are commonly visited during training, and thus states where the learned policies might not be very well tuned. Figure~\ref{fig:dual_sourcing_example}

We observe that the expedited ordering policy is little changed as a function of $L_r$, while the regular ordering policy appears to be converging to one where the agent places a constant order from the regular source, independent of the inventory level, consistant with the result that a Tailored Base-Surge policy is optimal as the regular lead time goes to infinity \cite{xin2018asymptotic}.

\subsection{Multi-Period Inventory Management with Returns}\label{sec:removals}

\subsubsection{Overview}

\begin{figure}[htbp]
\centering
\includegraphics[scale=0.8]{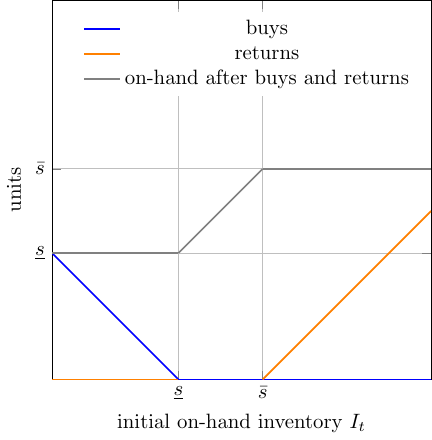}
\caption{Illustration of the optimal interval-stock policy for inventory management with returns.}\label{fig:interval_stock}
\end{figure}

Another twist on the basic multi-period inventory management problem is one where units may be returned to the vendor for a return value that is less than or equal to its original cost. Such return options are common in the retail world, and are especially important for seasonal products. The design of contracts including return rights has been considered in \cite{pasternack1985optimal} and \cite{cachon2003supply} for example, and its use in inventory management has been studied in \cite{maggiar2017joint}, where the role of returns in managing inventory in the face of capacity constraints is highlighted. In particular, it is demonstrated that assuming a fixed return value for all purchased units, the optimal policy is an \emph{interval-stock} one, where in addition to an order-up-to level $\underline{s}$, there also exists a somewhat symmetric remove-down-to level $\bar{s}$, such that if the inventory level $y$ is below $\underline{s}$, we raise the inventory up to $\underline{s}$, if the inventory is above $\bar{s}$, we remove units to bring the inventory down to $\bar{s}$, and we do nothing if $\underline{s}\leq y \leq \bar{s}$, as illustrated in Figure~\ref{fig:interval_stock}.

Note that in the stationary setting we are considering, once the inventory has been brought down to $\bar{s}$ (if it was necessary at all), only the order-up-to part of the policy will ever be needed from then on, so that we would only need to remove in the first period. Additionally, since our evaluation relies on measuring the average expected reward, we would be measuring the exact same thing we were measuring in the setting of the basic multi-period inventory management problem of Section~\ref{sec:lost_sales}. We will thus simply focus on in this section on a problem that involves a bi-dimensional action (buys and returns), and the ability of the DRL agent to recover the known optimal policy.

\subsubsection{MDP Formulation}

The model formulation assuming lost sales and returns is very similar to the problem of Section~\ref{sec:lost_sales}.

\paragraph{State} The endogenous state is simply given by the inventory level $y_t$, while the exogenous time series variables $\mathbf{x}_t$ consists of the previous $H$ demand realizations ($\mathbf{x}_t=(d_{t-H},\ldots,d_{t-1})$) and the exogenous static vector contains the economic parameters of a product, $\mathbf{s}=(p,c,h,b,r)$, where $p$ is the price, $c$ the purchasing cost, $h$ the holding cost,  $b$ the penalty for lost sale, and $r$ the value at which we can return inventory, with $r \leq c$.

\paragraph{Action} The action is here two-dimensional with $\mathbf{a}_t=(q_t, q_t^r)$, where $q_t$ is the number of units to purchase at time $t$, and $q_t^r$ is the number of units to return. Note that both purchases and returns happen instantaneously.

\paragraph{Transition Function} The transition function is simply given by $y_{t+1} = \max(y_t + q_t - q_t^r - d_t , 0)$, meaning that we sell up to $d_t$ units out of the $y_t+q_t-q_t^r$ units that are available for sale, and any unit left over is carried on to the next period.

\paragraph{Reward Function} In each period, we receive $p$ for each sold unit and $r$ for each unit returned, and incur unit costs of $c$ for any purchased unit, $b$ for any missed sale, and $h$ for any leftover unit, yielding:
\begin{align*}
R_t &= p \min(d_t, y_t+q_t- q_t^r) + r q_t^r - c q_t - b \max(d_t - y_t - q_t, 0) - h \max(y_t + q_t - q_t^r - d_t, 0).
\end{align*}

\subsubsection{Experiments}

We trained a DRL policy using the default values of Section~\ref{sec:overview}, and then evaluated the learned policy on a test set of 100,000 simulated products. Because the steady state of this problem is the same as that of Section~\ref{sec:lost_sales}, we do not present the measured average expected reward, which is essentially identical to what we obtained there, and focus exclusively on observing the structure of the learned policy. Note that in order to learn a return policy, we must create artificially high initial inventory levels, and we thus sampled during training initial inventory levels $y_0$ so that $y_0\sim 20 \mu U$, where $\mu$ is the mean demand of the product, and $U$ is a uniform random variable over [0,1].

\subsubsection{Results}

\begin{figure}[htbp]
\centering
\includegraphics[scale=0.4]{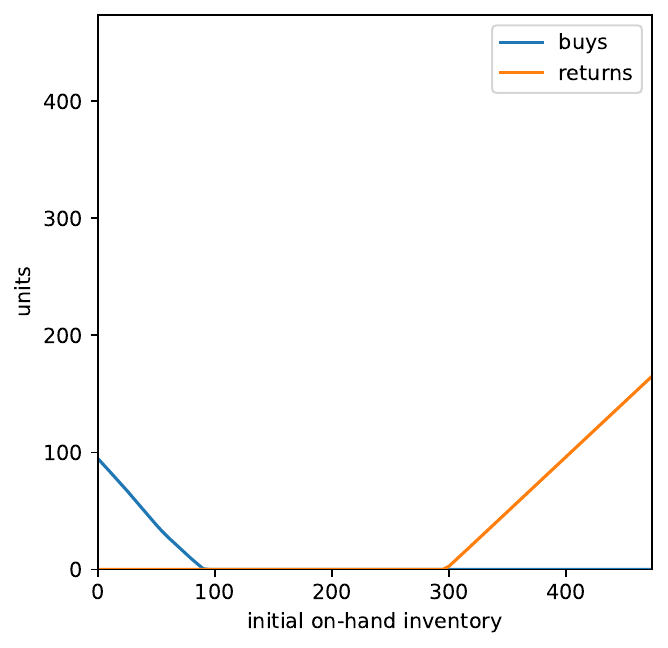}
\includegraphics[scale=0.4]{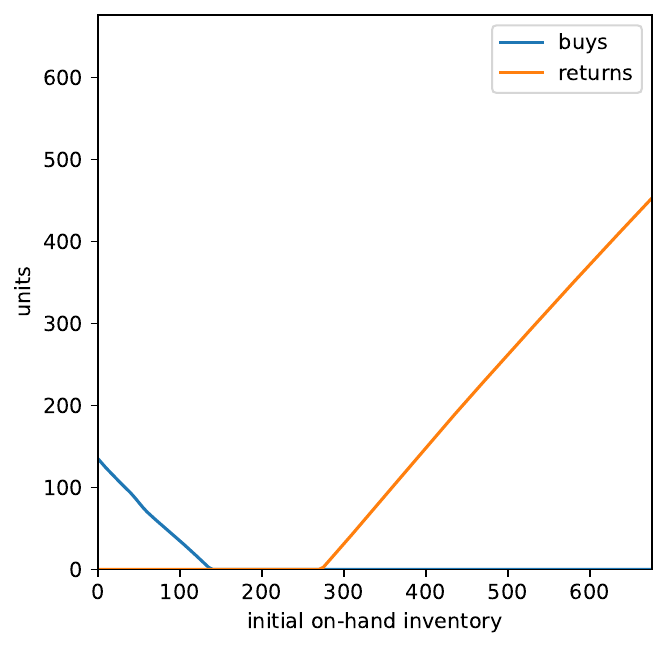}
\includegraphics[scale=0.4]{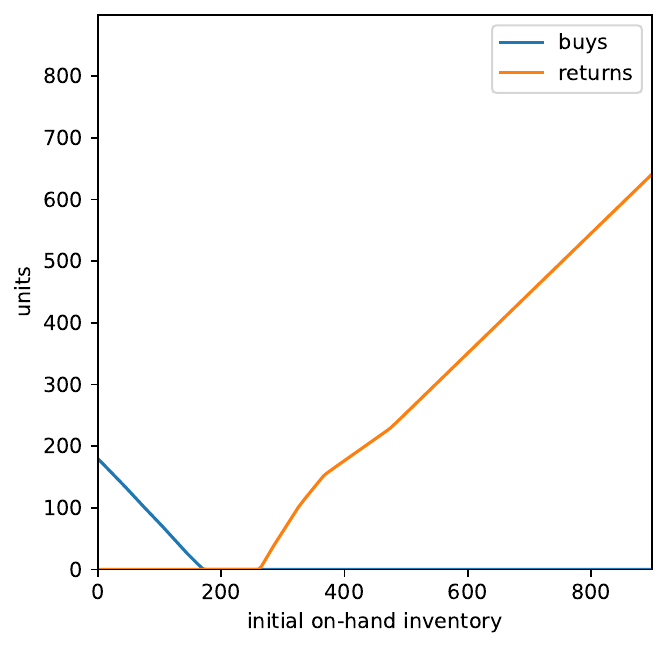}
\includegraphics[scale=0.4]{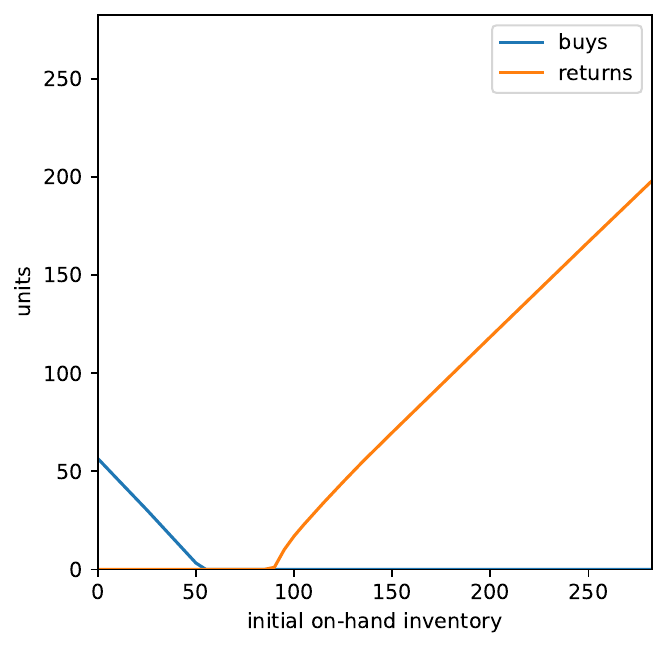}
\includegraphics[scale=0.4]{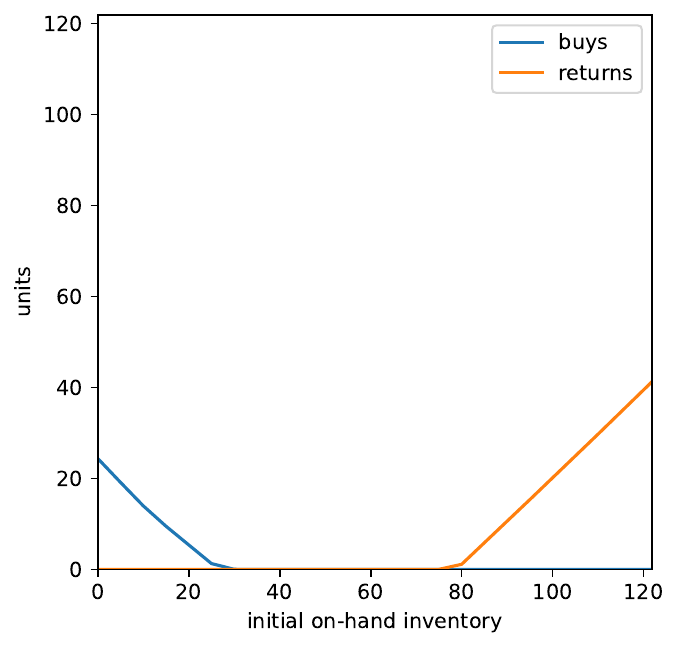}
\includegraphics[scale=0.4]{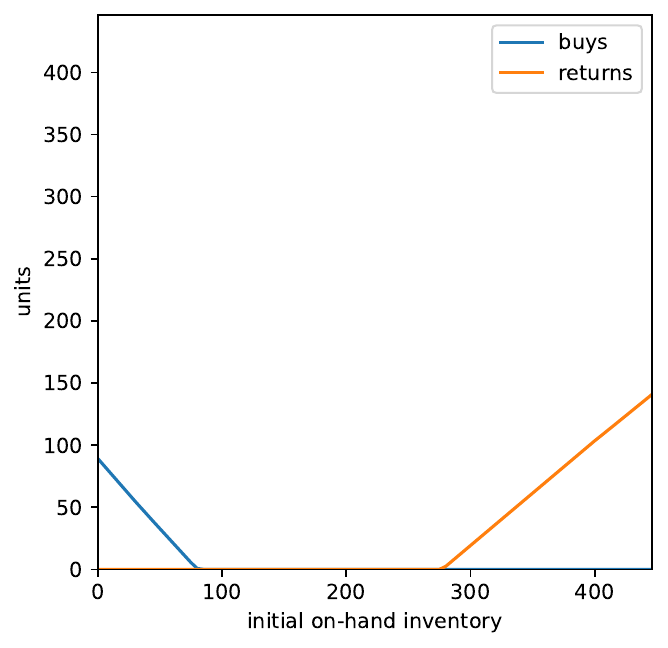}
\caption{Example plots of the buying and returning decisions produced by the DRL policy for a few given products in the case of inventory management with returns as a function of the initial on-hand inventory level.}\label{fig:returns_examples}
\end{figure}

As mentioned above, we merely focus here on observing whether the learned DRL policy learns the structural properties we expect from the optimal policy, namely the interval-stock behavior illustrated in Figure~\ref{fig:interval_stock}. We present in Figure~\ref{fig:returns_examples} the buying and returning decisions produced by the learned DRL policy for a few products on the test set at some specific time $t$, by varying their initial on-hand inventory. We observe that the DRL agent seems to be able to learn the right structural properties of the optimal policy, and this in spite of only having 1 period (the first) in each training rollout where it needs to remove inventory given the stationary nature of the problem.

\section{Enhancing Policies with Structural Results: Structure Informed Policy Network}\label{sec:sinn}

\subsection{Motivation}

The experiments and results presented in Section~\ref{sec:experiments} demonstrate the relevance of Deep Reinforcement Learning for inventory management. A plain agent was able to learn policies that perform on par with, or even better than, established benchmarks and heuristics in the literature, which further highlights its potential for practical applications given that it can be directly applied to less stylized settings than the ones typically studied in theoretical research. We further observed that while the DRL policy learns much of the known structures of the optimal policies, it can fail to display those characteristics throughout the state space. We saw this for example in the case of the example with lost sales and lead times in Section~\ref{sec:lost_sales_lead_times} where the policies plotted in Figures~\ref{fig:leadtime_drl_examples} and \ref{fig:leadtime_drl_examples_L_5} do not exhibit the monotonicity we would expect of the optimal policy everywhere. We already mentioned a reason for these spurious effects in the restricted region of the state space that is visited during training, and another risk would be to overfit the training data.

The lack of global structural conformism with that of the optimal policy was not detrimental in our experiments since a suboptimal policy in states that are not or seldom visited has little to no consequence on the reward. Nonetheless, when implementing a DRL agent in an industrial setting, such properties become more important for a couple of reasons:
\begin{itemize}
\item The training data might be either limited, or the distribution of the data subject to drift over time. In the context of a retailer, its growth, or evolving supply chain, affect different aspects of the economics and demands for the products, leading to test distributions that will be slightly off from the one seen in training,
\item  The adoption of the agent within a company requires convincing stakeholders and earning their trust that the solution is sound and robust. A policy that fails to produce actions that abide by simple rules expected of them, e.g. buying less if the inventory is higher, will garner hesitation from business partners who will hesitate in adopting it, especially in areas that are  customer or vendor facing.
\end{itemize}

In this section, we aim to develop mechanisms that would allow the DRL agent to learn policies that satisfy properties expected of them, with the dual goal of yielding more sensible policies, as well as hopefully policies that generalize better.

\subsection{Structural Properties through Partial Derivatives}\label{sec:partial_derivatives}

We first describe the mechanism through which structural information can be conveyed. A recurrent theme in inventory management is the characterization of optimal policies through properties of the partial derivatives of the actions with respect to its inputs (assuming a continuous setting). Take the simple lost sales example in the absence of lead time whose optimal policy is a base-stock one. The optimal buying quantity $q^*(y)$ decreases at a rate of 1 for inventory levels below the order-up-to level $y^*$ (we buy one fewer unit for each additional unit we have on-hand), and is identically null for inventory levels above it. We thus have $\frac{\partial q^*}{\partial y}=-1$ for $y< y^*$, and $\frac{\partial q^*}{\partial y}=0$ for $y\geq y^*$. A looser formulation of these conditions is: $-1\leq \frac{\partial q^*}{\partial y} \leq 0,~\forall y$. This type of properties of the optimal policy are abundant in inventory management, and we can return to our previously considered examples to illustrate this point:
\begin{description}
\item[Lead Times with Lost Sales:] The structure of the optimal policy in the case of inventory management with lead times and lost sales was first considered in \cite{morton1969bounds}, and revisited and strengthened in \cite{zipkin2008structure} using $L^\natural$-convexity. Using the notation we used in Section~\ref{sec:lost_sales_lead_times}, they proved the following bounds:
\begin{align*}
-1\leq \frac{\partial q^*}{\partial y^{L-1}} \leq \frac{\partial q^*}{\partial y^{L-2}}\leq \ldots \leq \frac{\partial q^*}{\partial y^{0}} \leq 0.
\end{align*}
In other words, the optimal order quantity is monotonically decreasing in all the pipeline variables, but not by more than one unit per unit increase in any pipeline variable, and additionally, the order quantity is more sensitive to more recent orders.
\item[Perishable Products:] In the case of perishable products, sensitivity results of optimal policy were first derived in \cite{nahmias1975optimal,fries1975optimal}, and then rederived and strengthened in \cite{li2014multimodularity} using multimodularity.  In particular, they prove the following string of inequalities:
\begin{align*}
-1\leq \frac{\partial q^*}{\partial y^{m-1}} \leq \frac{\partial q^*}{\partial y^{m-2}}\leq \ldots \leq \frac{\partial q^*}{\partial y^{1}} \leq 0.
\end{align*}
The interpretation here is similar: the optimal order quantity is monotonically decreasing in any inventory, regardless of remaining shelf life, but not by more than one unit per unit increase in any inventory. Additionally, the order quantity is more sensitive to newer inventory than older one.
\end{description}
In general, inventory management decisions possess a great deal of monotonicity properties, which are a focus of the study of many inventory problems, be it in terms of inventory decisions such as purchases or returns as in the examples we considered, or in terms of other related actions such as prices \cite{federgruen1999combined,hu2019joint,chen2023joint}. 

We just listed some monotonicity properties with respect to the endogenous variables in the examples we considered, but other similar monotonicity properties also hold with respect to static or series exogenous variables. For example in the case of the simple lost sales model, it is easy to see that the optimal order-up-to level is monotonically increasing with respect to the price, or monotonically decreasing with respect to the cost, so that $\frac{\partial q^*}{\partial p}\geq 0$ and $\frac{\partial q^*}{\partial c}\leq 0$.

\subsection{Approaches to Leverage Structural Properties}

\subsubsection{Fully Characterized Optimal Structures}
There are a few cases in inventory management where the optimal policy and its structure are fully known, and we mentioned a couple of them, notably in Section~\ref{sec:lost_sales} where the optimal policy is a simple base-stock policy, and in Section~\ref{sec:dual_sourcing} in the case of dual sourcing with backlogging where the lead times of the regular and expedited vendors were only one period apart, in which case a dual index policy is optimal. We saw in those cases that the DRL agent did learn most of the structure, even though that information had not been conveyed in any way. In such instances, one could design policies that specifically leverage this knowledge and form of the optimal policies. For example, in the case of the simple lost sales problem without lead time, we could remove the endogenous variable (inventory level) altogether from the inputs to the neural network, and modify the action to be the order-up-to level, instead of the buying quantity. Such modification does indeed help improve the performance of the learned policy, but situations where optimal policies can be fully described are rare in practice, and usually arise in largely stylized models. Additionally such a tailored implementation would be fragile, as minor changes to the underlying dynamics could quickly render it obsolete if these properties do not carry over.

\subsubsection{Monotonic Networks}

Given the prevalent presence of monotonicity properties in the structure of optimal inventory policies, one idea for the DRL agent would be to construct the neural architecture in a way that enforces monotonicity, or partial monotonicity only with respect to the relevant inputs. Such an approach has already been considered in domains where monotonicity is desired. For example, in probabilistic multi-horizon forecasting, where the quantile function should be monotone with respect to the input quantile levels, \cite{kan2022multivariate} represent the quantile function as the gradient of a convex function and leverage (Partially) Input Convex Neural Networks ((P)ICNN) \cite{amos2017input} to generate a monotone quantile function.

We could similarly enforce the (partial) monotonicity of the inventory policies by the use of (Partially) Monotonic Networks \cite{sill1997monotonic,daniels2010monotone}. However, there are several drawbacks to this approach. The first one is that (partial) monotonicity is not the only property satisfied by the policy, since for example in the case of lost-sales with lead times, we also had bounds on the magnitude of the sensitivity of the action ($\left|\frac{\partial q^*}{\partial y^i}\right|\leq 1,\forall i$), and on the relative sensitivities of the action with respect to the pipeline inventories ($\frac{\partial q^*}{\partial y^j}\leq \frac{\partial q^*}{\partial y^i},\forall j<i$). These properties would not be captured by a monotonic network, and we would still need to find ways to incorporate them. A second issue with monotonic networks lies in their expressiveness. The monotonicity of the network is guaranteed through constraints on the weights and activation functions for those connections and neurons that are used by the inputs with respect to which we want to enforce monotonicity. These either reduce the expressiveness of the network, or are difficult to train and do not work well in practice.

\subsubsection{Calibration and Certification Metrics}\label{sec:calibration_metrics}

A different perspective on the partial derivatives through which we express the structural properties is that they represent a metric that characterizes how ``calibrated" the policy is. A policy  would be fully ``calibrated" if it doesn't violate the structural differential equations. There are nonetheless additional choices to be made as to how to express such metrics.

As a means of illustration, consider the case of lead times and lost sales. Some of the structural equations are (see Section~\ref{sec:partial_derivatives}):
\begin{align}
\bar{g}_l(\bm\theta;\mathbf{z})&:=  - \frac{\partial q_{\bm\theta}^*}{\partial y^{l}} \leq 0, ~\forall l=0,\ldots,L-1,\label{eq:structural_inequality_1}\\
\underline{g}_l(\bm\theta;\mathbf{z})&:= \frac{\partial q_{\bm\theta}^*}{\partial y^{l}} + 1 \leq 0, ~\forall l=0,\ldots,L-1,\label{eq:structural_inequality_2}\\
g_{i,j}(\bm\theta;\mathbf{z})&:= \frac{\partial q_{\bm\theta}^*}{\partial y^{i}} - \frac{\partial q_{\bm\theta}^*}{\partial y^{j}}  \leq 0,~ 0\leq j\leq i\leq L-1.\label{eq:structural_inequality_3}
\end{align}
Here we emphasize that these are the actions produced by a policy parametrized by $\bm\theta$.

We may then apply to these inequalities a penalty function $h$  that is null if the inequality is satisfied, and is monotonically increasing in the violation otherwise. For example, we could let $h$ be a quadratic penalty function $h(x):=\max(0,x)^2$. The expected value of this violation penalty over the state space can then be interpreted, for each structural equation, how much it fails to conform to that particular property.

Recall that we let $\mathbb{P}$ be the joint distribution of the exogenous variables. For a given structural inequality $g$ and value $\mathbf{y}$ of the endogenous variables, we define the calibration metric $C(g;\bm\theta,\mathbf{y})$ as:
\begin{align*}
C(g;\bm\theta,\mathbf{y}) := \mathbb{E}_\mathbb{P}\left[h(g(\bm\theta,\mathbf{z}))\mid \mathbf{y}\right].
\end{align*}
Abusing notation, we then let $C(g;\bm\theta,\mathbf{Y})$ be the expected value of the metric over \emph{some} distribution over the endogenous state:
\begin{align*}
C(g;\bm\theta,\mathbf{Y}) := \mathbb{E}_\mathbf{Y}\left[C(g;\bm\theta,\mathbf{Y})\right].
\end{align*}
This calibration metric informs us on how well  the policy abides by the structural inequality condition over the specified distribution of endogenous variables. The idea is then to use this metric in the reward function so as to help shape the policy towards satisfying the corresponding structural property. To that end the objective function used in training balances reward maximization and satisfaction of the structural properties:
\begin{align*}
\mathbb{E}_{\mathbf{Y}_0}\left[J_T(\bm\theta;\mathbf{Y}_0)\right] - \lambda \sum_{g\in\mathcal{C}} \sum_{t=0}^{T-1} \gamma^t C(g;\bm\theta,\tilde{\mathbf{Y}}_t),
\end{align*}
where $\lambda >0$ is some parameter that places more or less weight on the structural penalties\footnote{The penalty parameter $\lambda$ can also be increased from iteration to iteration during training to gradually enforce the properties.}, $\mathcal{C}$ is the set of structural inequalities, and $\tilde{\mathbf{Y}}_t$ are arbitrarily distributed random variables in each period $t$. 

An obvious and convenient choice for the distributions of the $\tilde{\mathbf{Y}}_t$ is to let them be the distributions obtained from  the rolling out the policy parametrized by $\bm\theta$ as this allows us to write the objective function in a concise manner as:
\begin{align}
\mathbb{E}_{\mathbb{P},\mathbf{Y}_0}\left[\sum_{t=0}^{T-1}\gamma^t\underbrace{\left(R_t(\bm\theta,\mathbf{z}_t)-\lambda\sum_{g\in\mathcal{C}} h(g(\bm\theta,\mathbf{z}_t))\right)}_{R_t'(\bm\theta,\mathbf{z})}\middle|\mathbf{Y}_0\right].\label{eq:penalized_reward}
\end{align}
In doing so, we benefit from two advantages:
\begin{itemize}
\item It only requires augmenting the reward function with simple penalty terms, and thus represents only a minor and versatile approach where structural penalties can be easily plugged or removed.
\item It is especially convenient in the case of a DRL approach, given that we expressed the structural properties as differential inequalities. The partial derivatives at the visited states are readily available during training thanks to automatic differentiation \cite{paszke2017automatic}, and as a result such an implementation only has a minor numerical cost.
\end{itemize}
A drawback of using these ``natural" distributions is that they do not offer guarantees that the structural properties will hold everywhere since we only apply the constraints to the visited states, hoping that the properties will spread to other regions.

Returning to the special case of monotonic properties, both approaches can be found in the literature to overcome the issues mentioned above about monotonic networks, and instead of baking the monotonicity directly into the architecture, trying to enforce it through regularization. An early attempt at incorporating a penalty against non-monotonic function values was presented in \cite{sill1996monotonicity}. This was reintroduced in the context of neural networks in \cite{gupta2019incorporate}, where a penalty based on the divergence of the output function with respect to the set of features in which the function is monotonous, but still using the ``natural" distributions of the states. \cite{liu2020certified} introduce a penalty whose expectation is taken over points sampled uniformly in the domain (which is implicitly assumed to be bounded) in order to provide ``certificates'' on how well the monotonicity property is satisfied throughout the domain.

\subsubsection{Physics Informed Neural Network (PINN)}

Another perspective on Equation~\ref{eq:penalized_reward}, where we penalize the reward function based on the deviation from a partial differential equation, is that it is reminiscent of Physics Informed Neural Networks (PINN) \cite{raissi2019physics}, where a penalty is applied on deviations from physical laws governing the system being modeled. We can interpret the structural differential equations as governing the evolution of the policy over the state space. The generalization properties of PINNs has been studied, and some results can be found in \cite{shin2020convergence,mishra2022estimates}.

\subsection{Structure-Informed Network}\label{sec:sinn_implementation}

\subsubsection{Implementation}

We now describe the mechanism used to infuse the policy network with known structural information. We essentially use the process described in Section~\ref{sec:calibration_metrics}, and in particular Equation~\ref{eq:penalized_reward}, where we simply add a penalty term to the reward function, which is evaluated at the points visited during training. We assume that the structural differential equations are expressed as inequalities, such that the corresponding equations must be less than or equal to 0, as in Equations~\ref{eq:structural_inequality_1}-\ref{eq:structural_inequality_3}. For simplicity, we use a quadratic penalty of the form $h(x)=\max(0,x)^2$, yielding the following reward function in training:
\begin{align}
\mathbb{E}_{\mathbb{P},\mathbf{Y}_0}\left[\sum_{t=0}^{T-1}\gamma^t\left(R_t(\bm\theta,\mathbf{z}_t)-\lambda\sum_{g\in\mathcal{C}} \max(0,g(\bm\theta,\mathbf{z}_t))^2\right)\middle|\mathbf{Y}_0\right],\label{eq:penalized_reward_quadratic}
\end{align}
where we recall that $\mathcal{C}$ is the set of structural differential equations, and $\lambda$ a hyperparameter that balances reward and structural loss.

This formulation is quite versatile and results in a low marginal computational  expense for penalties on structural equations can easily be added or removed, and their computation is straightforward since they are based on partial derivatives of the policy with respect to (some of) its inputs, which are readily available through automatic differentiation. We illustrate the impact of the scheme on the case with lead times and lost sales in the next section.

\subsubsection{Illustration}

We consider the case of inventory management with lead times and lost sales studied in Section~\ref{sec:lost_sales_lead_times}, where we had observed that the learned policies failed to fully satisfy the structural properties expected of the optimal policy. We retrained the policies using the penalized objective function Equation~\eqref{eq:penalized_reward_quadratic} instead of the original \eqref{eq:opt_problem}, for different values for the penalty parameter $\lambda$. We included the structural constraints \eqref{eq:structural_inequality_1}-\eqref{eq:structural_inequality_3}, which are only a subset of all the valid structural constraints.

We should not necessarily anticipate an improvement in the expected average reward since it is computed on a ``steady-state" that the unpenalized policy can tailor itself to. Thus, unfortunately, our original setup is not very conducive to estimating the extrapolation benefits of the structure-informed policies, and we use a more appropriate setting in Section~\ref{sec:favorita} below. Nonetheless, we can obtain some insight by: 1) inspecting the learned policies and compare them to the ones of the unpenalized policies, and 2) observing the training curves to assess the impact of the regularization on training.

\paragraph{Expected Average Reward}
Table~\ref{tab:penalized_evaluation_results} shows the impact of the penalty on the expected average reward for all the lead times $L$. We observe that there is little impact to the expected average reward. There is some slight degradation for higher values of the penalty multiplier $\lambda$ and longer lead times, which can partly be explained  by the fact that it somewhat slows down the learning, and thus the truncation after 1,000 epochs doesn't allow the penalized policy to fully converge.

\begin{table}[htbp]
\centering
\caption{Impact of the penalty on the expected average reward for the model with lead times and lost sales.}\label{tab:penalized_evaluation_results}
\begin{tabular}{lcccccc}
\toprule
 & \multicolumn{5}{c}{$\lambda$}\\
 $L$ & 0 & 100 & 1,000 & 10,000 & 100,000 & 1,000,000\\
 \midrule
2 & - & 0.00\% & 0.02\% & 0.01\% & 0.01\% & 0.00\% \\
3 & - & 0.00\% & 0.00\% & 0.01\% & 0.02\% & 0.01\% \\
4 & - & 0.01\% & -0.01\% & -0.04\% & -0.07\% & -0.05\% \\
5 & -& 0.02\% & 0.07\% & 0.07\% & 0.06\% & 0.05\% \\
6 & - & 0.00\% & -0.03\% & -0.04\% & -0.07\% & -0.10\% \\
7 & -& -0.13\% & -0.12\% & -0.18\% & -0.13\% & -0.26\% \\
\bottomrule
\end{tabular}
\end{table}

\begin{figure}[htbp]
\centering
\includegraphics[scale=0.5]{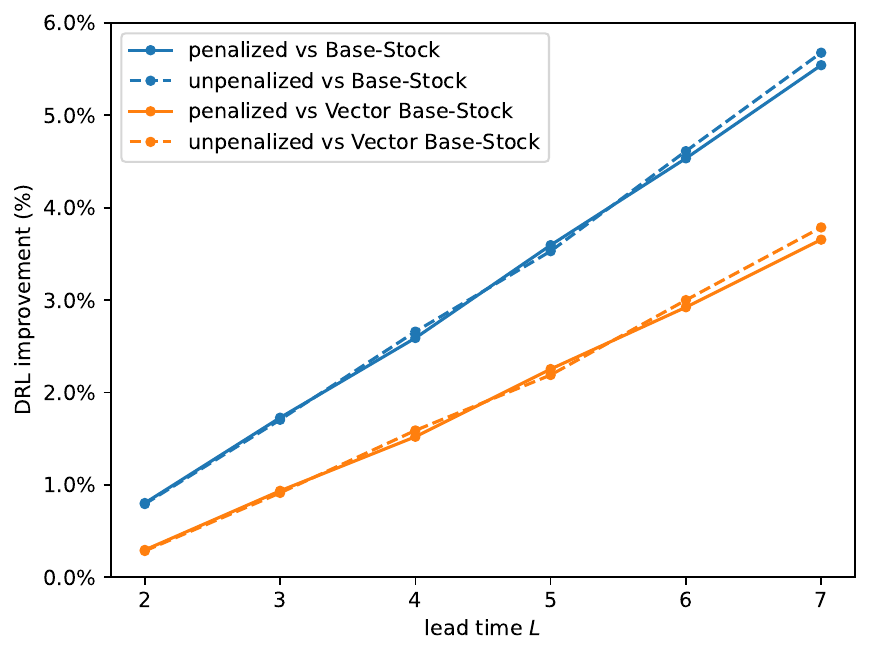}
\caption{Relative improvement of the unpenalized and penalized ($\lambda=1e5$) DRL policies over the benchmarks in the case of lost sales with lead times, as a function of the lead time.}\label{fig:penalized_results}
\end{figure}

Figure~\ref{fig:penalized_results} presents the relative improvements of the penalized DRL policies over the benchmarks considered in Section~\ref{sec:lost_sales_lead_times}, reproducing the results of the unpenalized policies of Figure~\ref{fig:leadtimes_results}. We observe little impact on the performance.

\begin{figure}[htbp]
\centering
\includegraphics[scale=0.35]{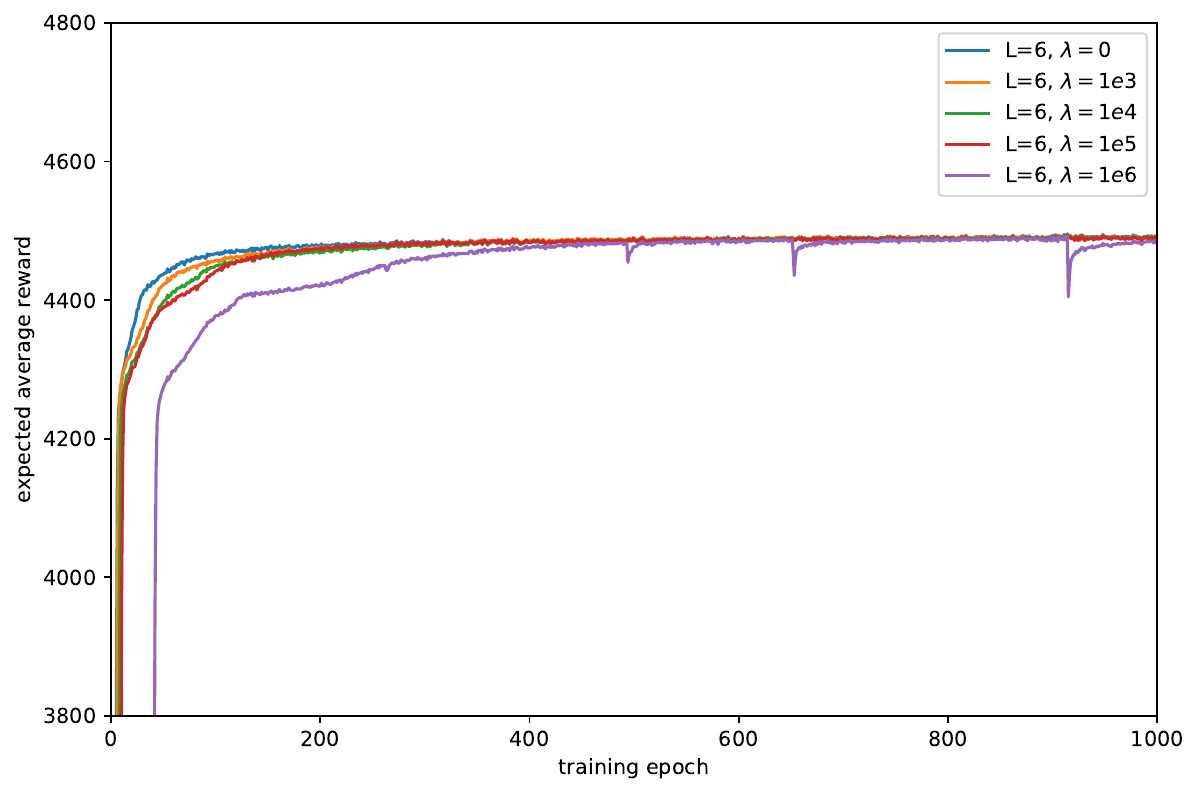}
\includegraphics[scale=0.35]{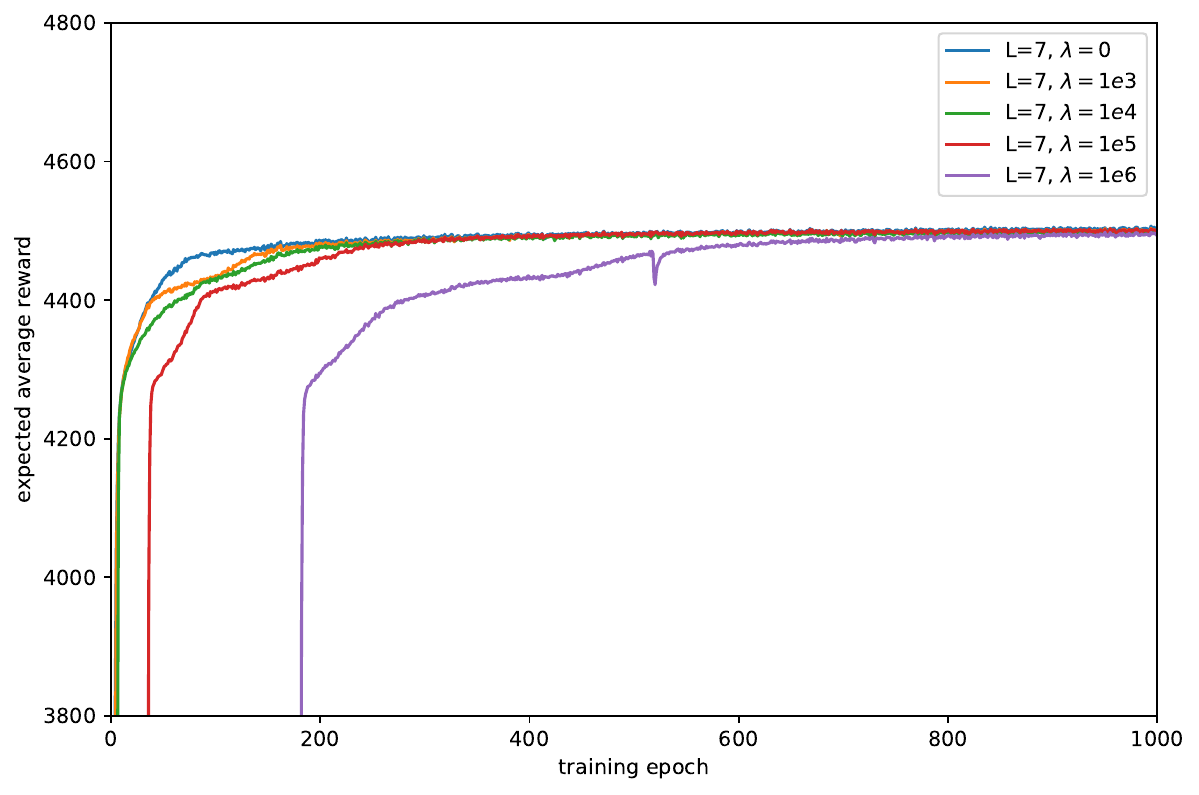}
\caption{Training curves of the DRL policies for lead  times $L=6,7$ for different values of the penalty parameter $\lambda$.}\label{fig:regularized_training_curves}
\end{figure}

\paragraph{Training Curves}
Figure~\ref{fig:regularized_training_curves} shows the training curves of the DRL policies for lead times $L=6,7$ for different values of the penalty parameter $\lambda$.  We observe that the addition of the penalty term can slow down convergence at higher values of $\lambda$.

\begin{figure}[htbp]
\centering
\includegraphics[scale=0.33]{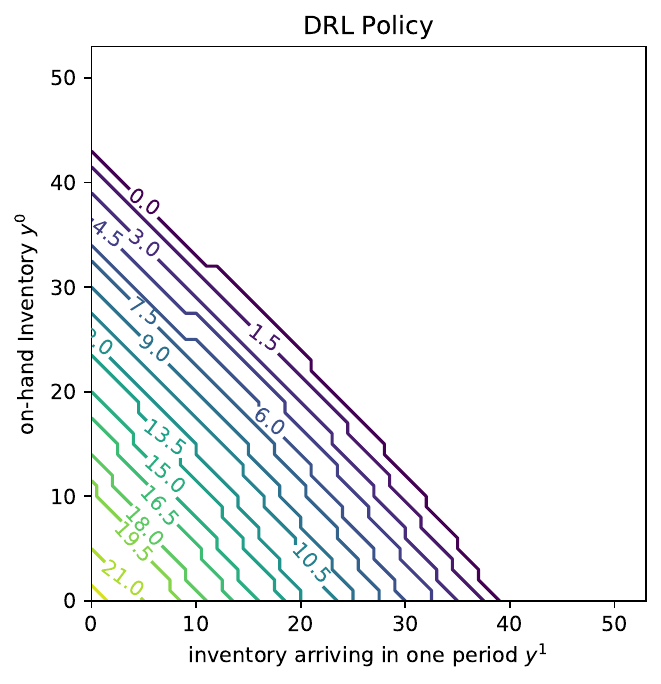}
\includegraphics[scale=0.33]{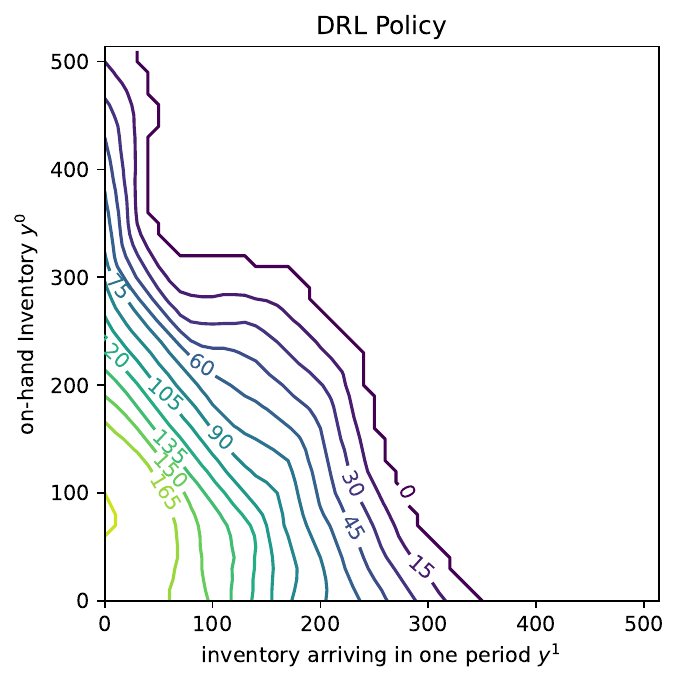}
\includegraphics[scale=0.33]{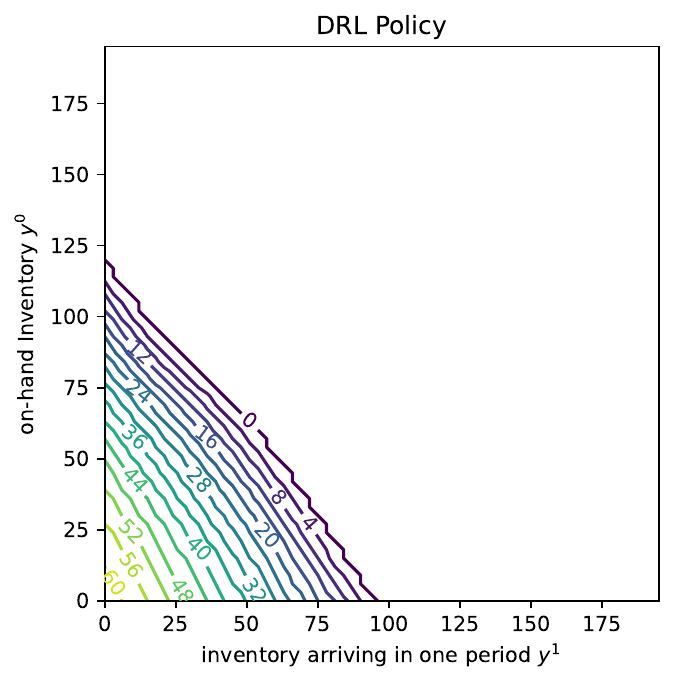}
\includegraphics[scale=0.33]{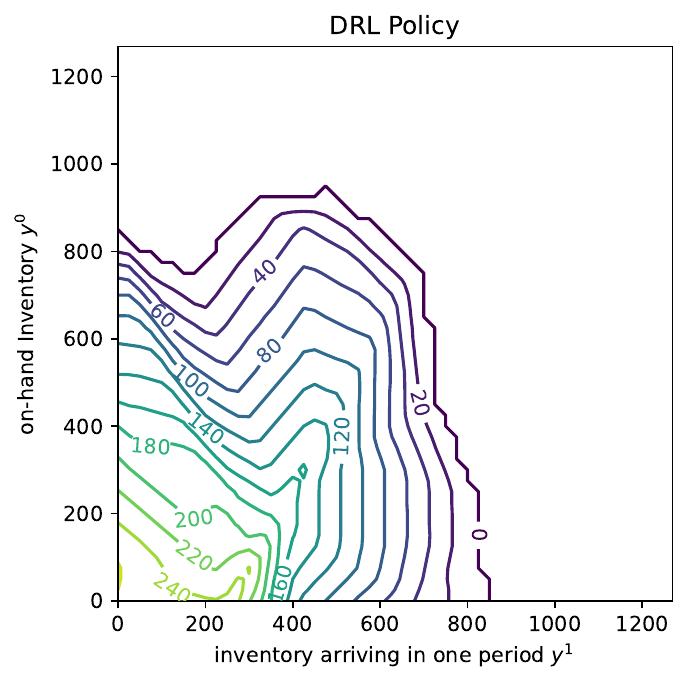}

\includegraphics[scale=0.33]{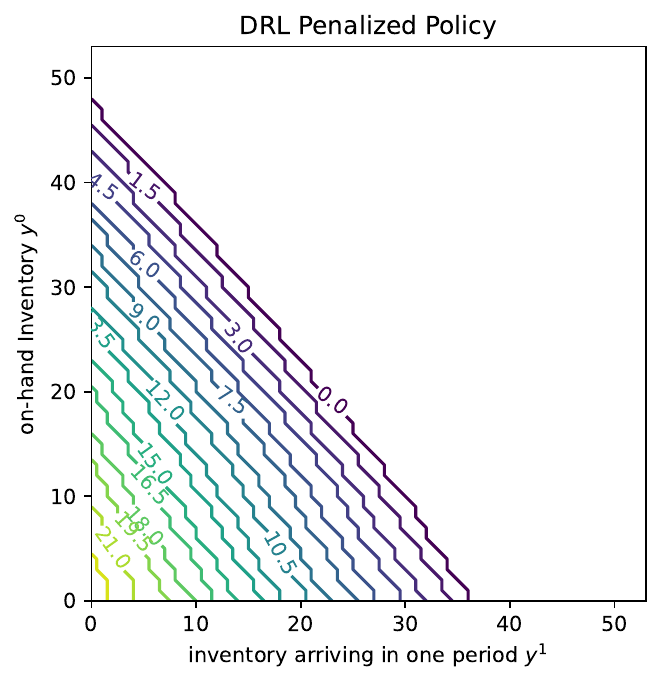}
\includegraphics[scale=0.33]{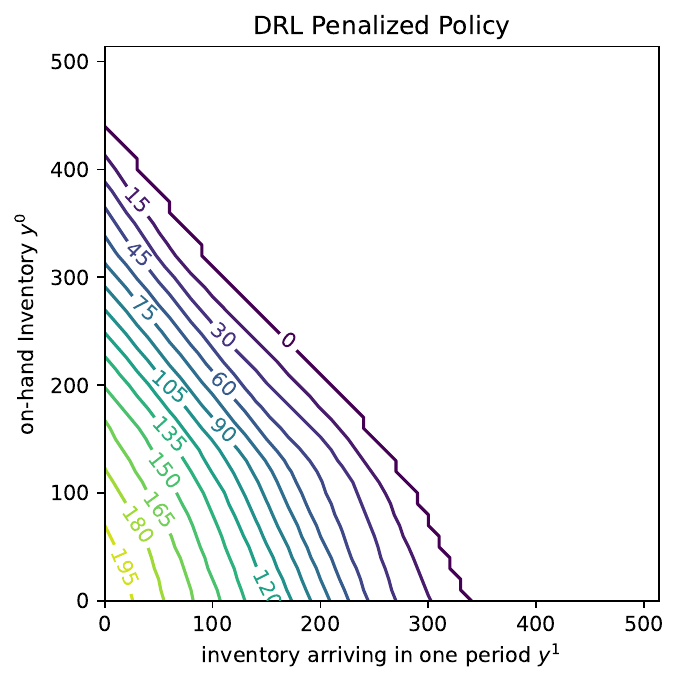}
\includegraphics[scale=0.33]{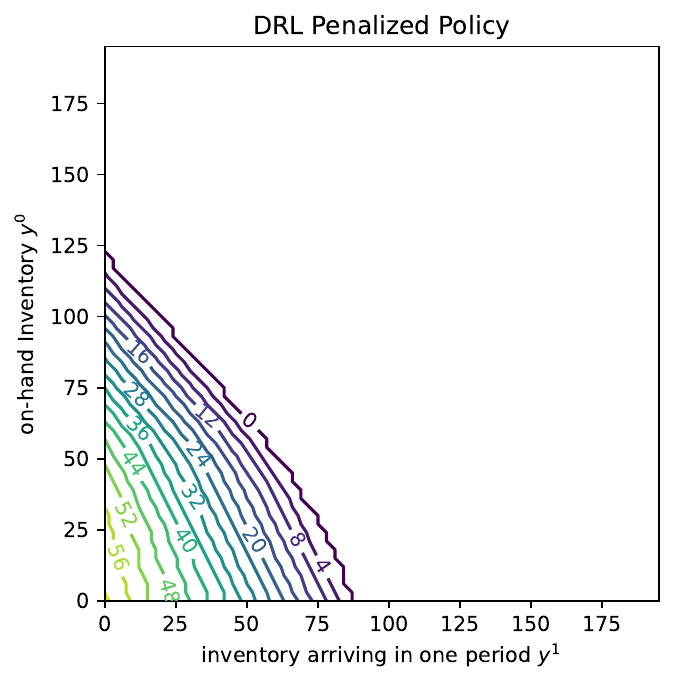}
\includegraphics[scale=0.33]{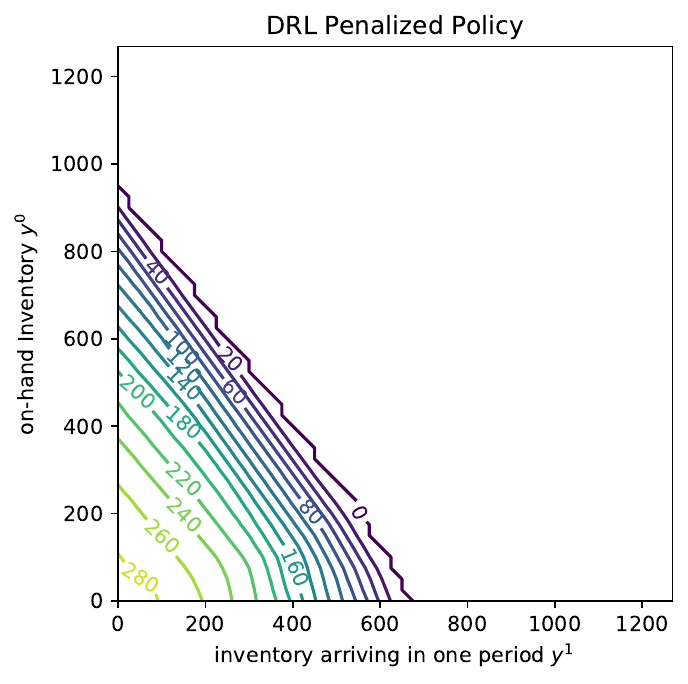}
\caption{Example contour plots of the unpenalized (top), and penalized with $\lambda=1e5$ (bottom) DRL policies for a few given product in the case of $L=2$ as a function of the endogenous state $\mathbf{y}=(y^0,y^1)$ at time $t=0$.}\label{fig:leadtime_penalized_L_2}
\end{figure}

\begin{figure}[htbp]
\centering
\includegraphics[scale=0.33]{figs/leadtime_DRL_L_5_asin_1_nonpenalized.pdf}
\includegraphics[scale=0.33]{figs/leadtime_DRL_L_5_asin_2_nonpenalized.pdf}
\includegraphics[scale=0.33]{figs/leadtime_DRL_L_5_asin_3_nonpenalized.pdf}
\includegraphics[scale=0.33]{figs/leadtime_DRL_L_5_asin_8_nonpenalized.pdf}

\includegraphics[scale=0.33]{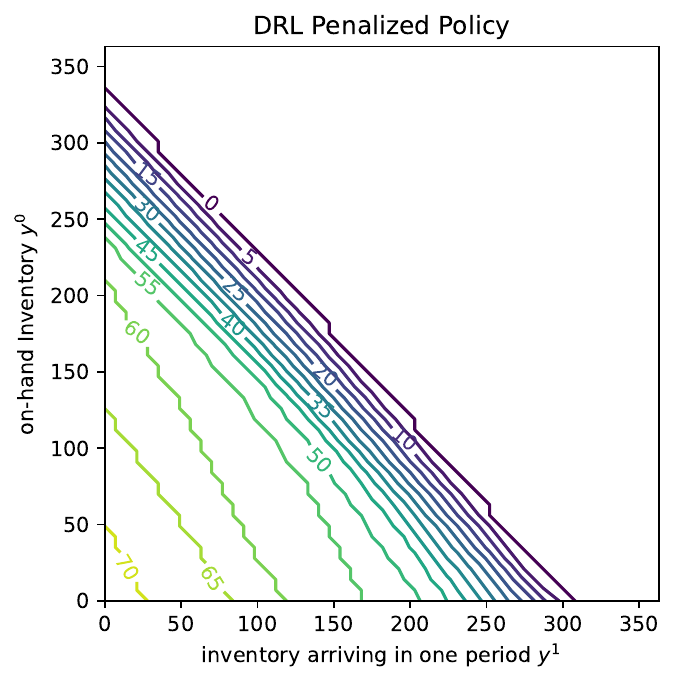}
\includegraphics[scale=0.33]{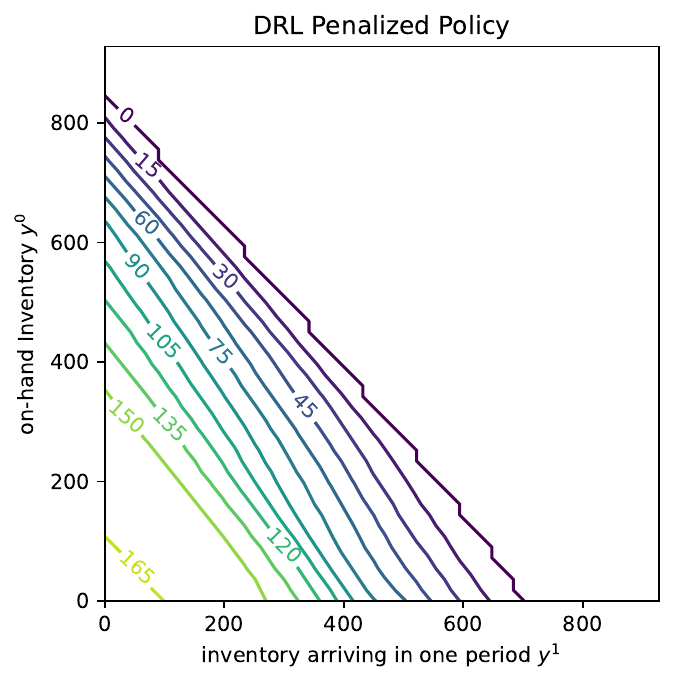}
\includegraphics[scale=0.33]{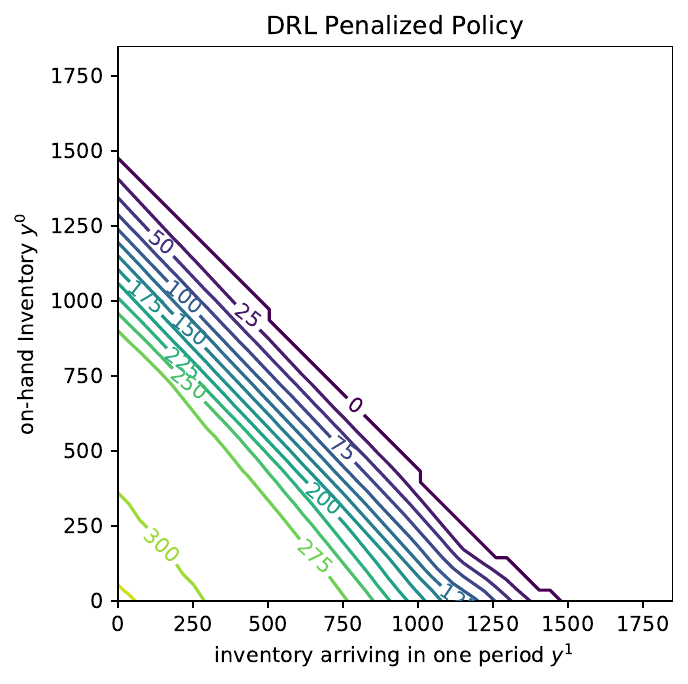}
\includegraphics[scale=0.33]{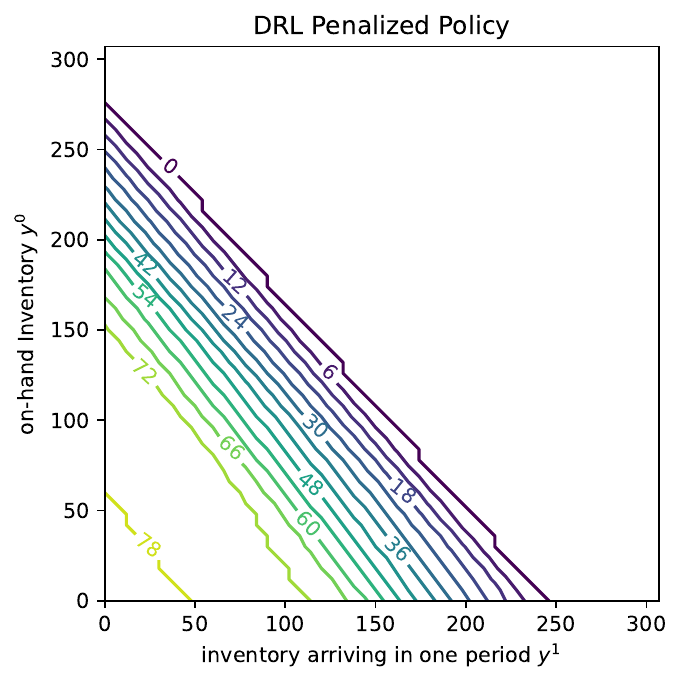}
\caption{Example contour plots of the unpenalized (top), and penalized with $\lambda=1e5$ (bottom) DRL policies for a few given product in the case of $L=5$ as a function of the endogenous state $\mathbf{y}=(y^0,y^1,0,0,0)$ at time $t=0$.}\label{fig:leadtime_penalized_L_5}
\end{figure}

\paragraph{Policies}
Part of the motivation for the insertion of structural information in the reward function was the observation that while the DRL policies performed well and exhibited overall structures that would be expected of an optimal policy, they still failed to uniformly  display that behavior. We return to the few examples plotted in Figure~\ref{fig:leadtime_drl_examples_L_5} and Figure~\ref{fig:leadtime_drl_examples_L_5} for $L=2$ and $L=5$, respectively, in order to check the impact that the penalization had on the learned policy. Figures~\ref{fig:leadtime_penalized_L_2} and~\ref{fig:leadtime_penalized_L_5} compare the unpenalized (top)  and penalized (bottom) results for the same products, and we observe that the policies have been stretched into  better ones from the perspective of satisfying the structural constraints.

\paragraph{Extrapolation}
One of the desired behaviors of the structured-informed policy is that it would yield better decisions in states that are outside of the range usually visited in training, and materially translate into better results on test data. We consider this effect using realistic demand data in the next section.

\subsection{Extrapolation Example}\label{sec:favorita}

\subsubsection{Setting}

\begin{figure}[htbp]
\centering
\includegraphics[scale=0.5]{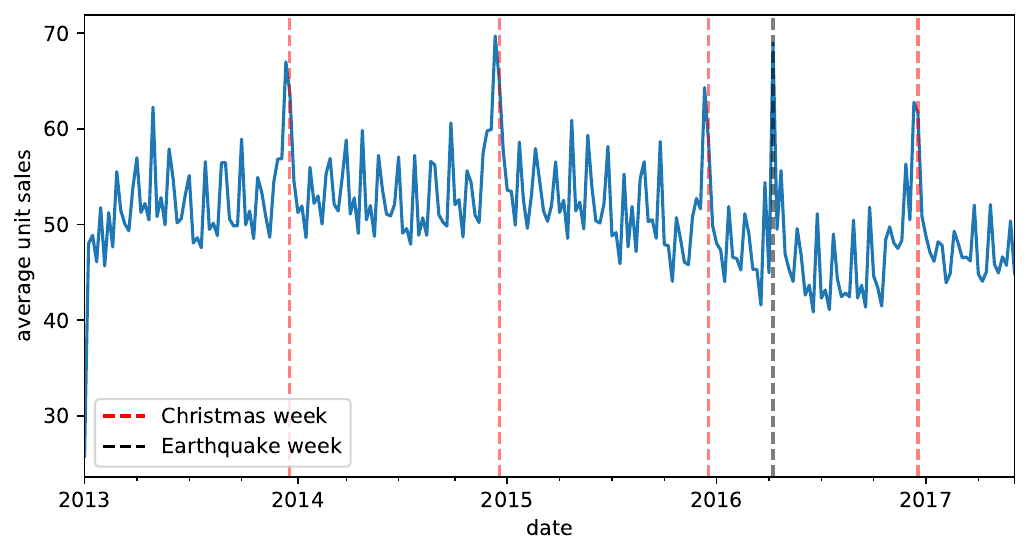}
\caption{Average unit sales of the products in the Favorita data.}\label{fig:favorita_mean}
\end{figure}

We aim in this section to illustrate the effect of infusing the DRL policy with structural information as described in Section~\ref{sec:sinn_implementation}. To do so, we consider in some realistic non-stationary  grocery sales data shared by the Corporacion Favorita, an Ecuadorian retailer. This data was originally shared as part of a Kaggle competition \cite{favorita-grocery-sales-forecasting}, and further processed and used in \cite{alvo2023neural} who made their version of the data available online\footnote{\url{https://github.com/MatiasAlvo/Neural_inventory_control}}. The data consists of about 33k demand traces for weekly sales of products across 54 stores from January 2013 to  August 207. The average unit sales of the products over this period is plotted in Figure~\ref{fig:favorita_mean} where we observe major spikes around Christmas, as well as around the earthquake that hit Ecuador in April 2016. Additionally, we also observe some form of periodicity with minor spikes about every 4-5 weeks, which correspond to the weeks marking the end/beginning of months, presumably explained by people receiving their salaries.

To test the potential benefits of structure-informed policies, we still consider the data generating process described in Section~\ref{sec:overview}, and in particular in Table~\ref{tab:distributions}, but instead of generating demand traces from stylized Gamma distributions, we directly use the realistic ones, albeit at times censored, of the Favorita dataset. 

The benefits of the principled structural regularization are likely to be more pronounced in cases of limited training data, which may not be sufficient for the DRL agent to capture the full sense of the optimal policy. To observe this behavior, we train a policies on the Favorita data, assuming a lead time of 5 periods, and using a history window of $H=16$ weeks on an increasing number of products (100,500,1k,5k,10k) on 26 weeks (6 months) worth of data, excluding $H$ weeks to initialize the endogenous variables, corresponding to the first 42 weeks of 2013. For each value of the sample size, we sample 200 training sets, and thus learn 200 policies, with and without structural penalty, using a penalty parameter $\lambda=1e6$).  We then test the learned policies on the data for 30k products using the demand data starting in January 2014. Note that the training data does not include any Christmas period, and we do not provide any information as to the distance to Christmas, or any other time reference. The purpose is for the Christmas periods, as well as the earthquake, to represent ``shocks'' the policy needs to recover from.

\begin{table}[htbp]
\centering
\caption{Datasets used for training and testing in the Favorita example.}\label{tab:favorita_data}
\begin{tabular}{lcc}
\toprule
Set & $N$ & time period \\
\midrule
Training & 100,500,1k,5k,10k & 16 + 26 weeks (starting 2013-01-01)\\
Testing & 30k & 16 + 162 weeks (starting 2014-01-01)\\
\bottomrule
\end{tabular}
\end{table}

We use the same neural architecture and parameters as the ones in Table~\ref{tab:hyperparams}, except for a history window of $H=16$ periods, and a learning rate of 0.003.

We further note that the aim is not here to assess the DRL policy against a benchmark, but merely evaluate the impact of the structural regularization, so we compare penalized DRL policies against unpenalized ones.

\subsubsection{Results}

\begin{figure}[htbp]
\centering
\includegraphics[scale=0.42]{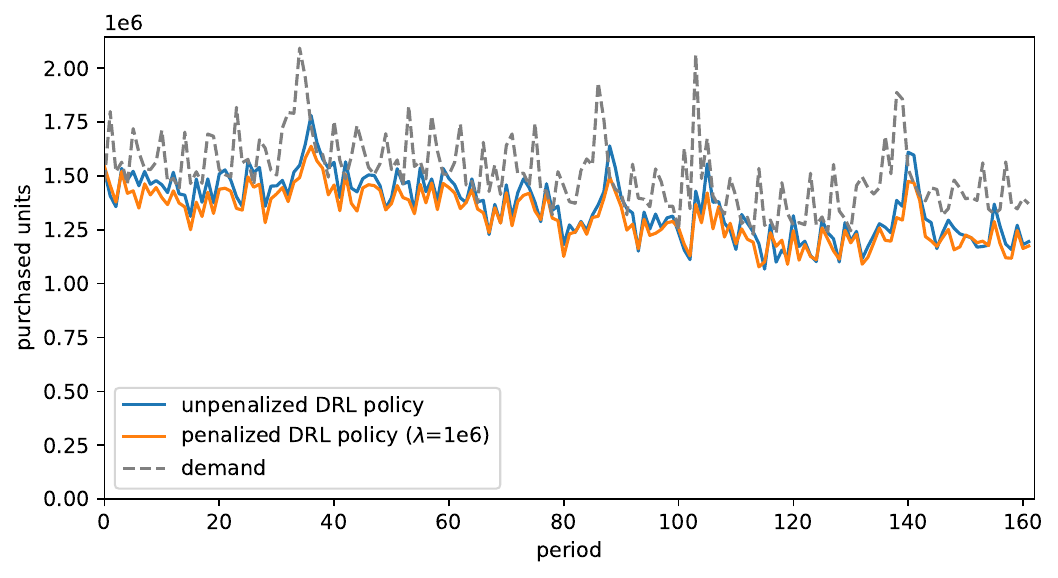}
\includegraphics[scale=0.42]{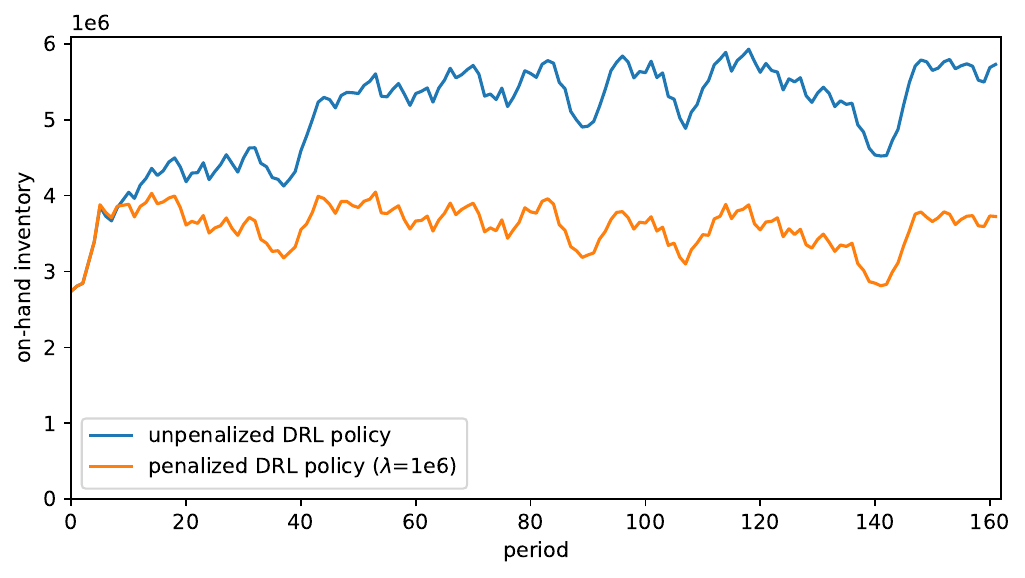}
\caption{Aggregate purchased units (left) and on-hand inventory (right) on the test set for the unpenalized and penalized DRL policies trained on one of the 200 sets with $N=5k$ products.}\label{fig:favorita_5k}
\end{figure}

For each value of the sample size $N\in[100,500,1k,5k,10k]$, we evaluate the 200 learned policies, with and without structural penalty, on the testing set. To check whether the learned policies are sensible, we can plot the aggregate order quantities and aggregate on-hand inventory over time of the DRL policies. Figure~\ref{fig:favorita_5k} shows those statistics for the policies learned on one of the 10k products training set. The left plot shows the aggregate purchased units for each of the two policies, and we observe that their levels make sense, and that they react to the Christmas and earthquake spikes by placing large purchases immediately after those deplete inventory. However, we observe on the right plot that the unpenalized policy operates at a higher inventory level. We also recall that the lead time is of 6 periods to explain why their inventory levels are identical for the first 6 periods. For this particular training set, the penalized policy did indeed outperform the unpenalized one on the test set by about 1.8\%.

\begin{figure}[htbp]
\centering
\includegraphics[scale=0.5]{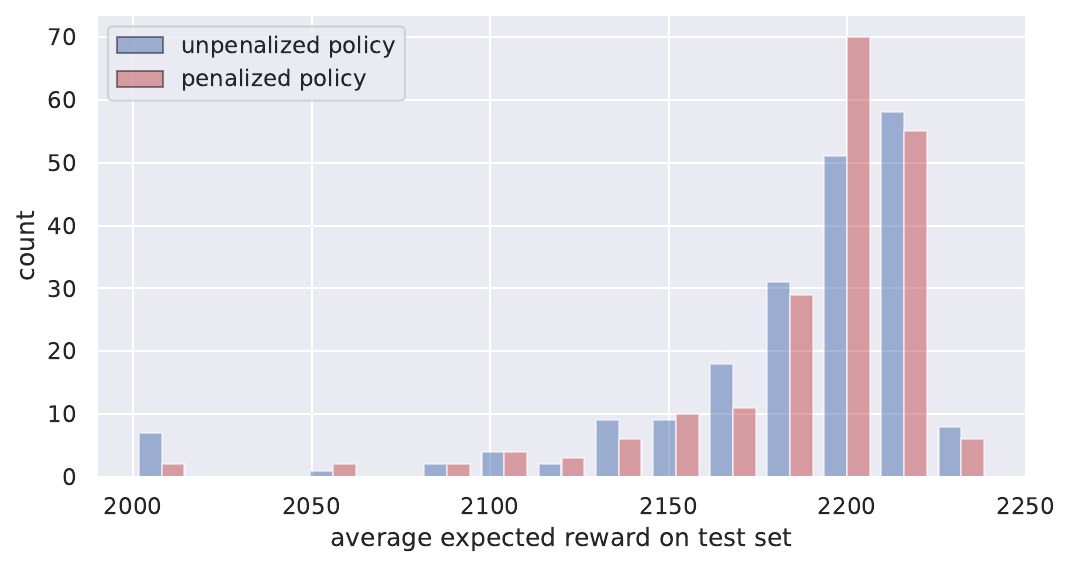}
\caption{Histogram of the average expected rewards on the test set for the unpenalized and penalized DRL policies trained on 200 sets of randomly generated 1k products. The left tail was clipped at 2,000.}\label{fig:favorita_hist}
\end{figure}

In order to make a better assessment as to the benefits, or lack thereof, of the penalization procedure, we consider the distribution of average expected reward over the 200 sampled test sets for each of the sample sizes. Figure~\ref{fig:favorita_hist} shows the histogram of average expected returns on the test set of the learned policies on the 200 randomly generated training sets with 1k products, where we clipped the left tail at 2,000.

We observe that the distribution of test set average expected returns for the penalized policies is tighter around its mode than that of the unpenalized policy. Crucially, we clipped the left tail of the distributions at 2,000 for display purposes, but we see that the left tail (indicating poorer test performance) of the unpenalized policy is heavier. Not only do we have more instances of poorer performance, but also instances of ``catastrophic'' failure, where the policy gets in a self-defeating loop. 

\begin{figure}[htbp]
\centering
\includegraphics[scale=0.42]{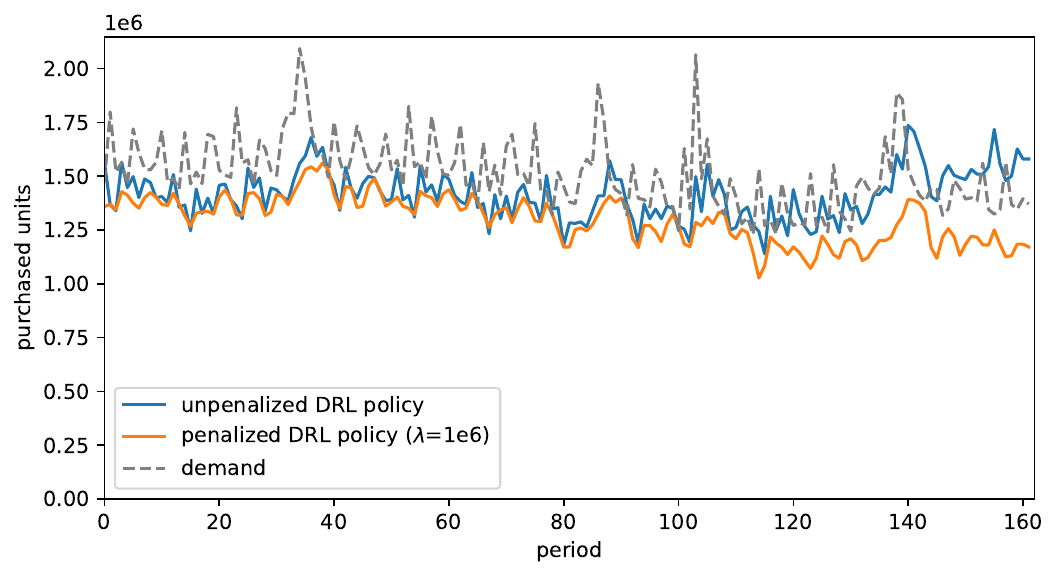}
\includegraphics[scale=0.42]{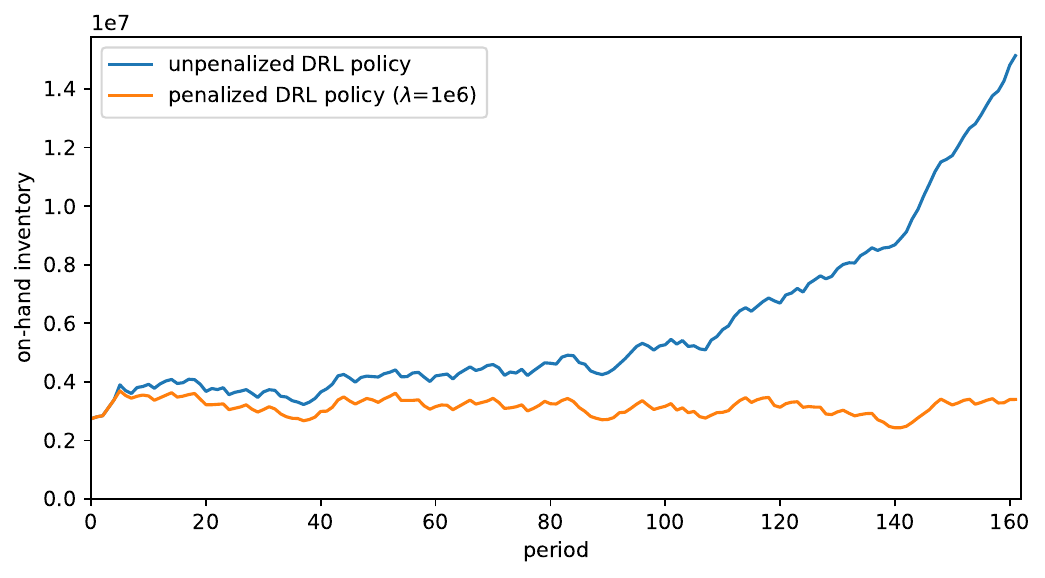}
\caption{Aggregate purchased units (left) and on-hand inventory (right) on the test set for the unpenalized and penalized DRL policies trained on one of the 200 sets with $N=1k$ products.}\label{fig:favorita_catastrophic_failure}
\end{figure}

An example of such behavior is presented in Figure~\ref{fig:favorita_catastrophic_failure}, where we show the aggregate unit purchases and on-hand inventory of penalized and unpenalized DRL policy trained on the same 1k product set, over the test set data. The unpenalized policy initially operates at a higher inventory level than the penalized one, but is then further thrown out of balance after the earthquake demand spike, leading to a loop of ever-increasing purchases and inventory.

\begin{figure}[htbp]
\centering
\includegraphics[scale=0.42]{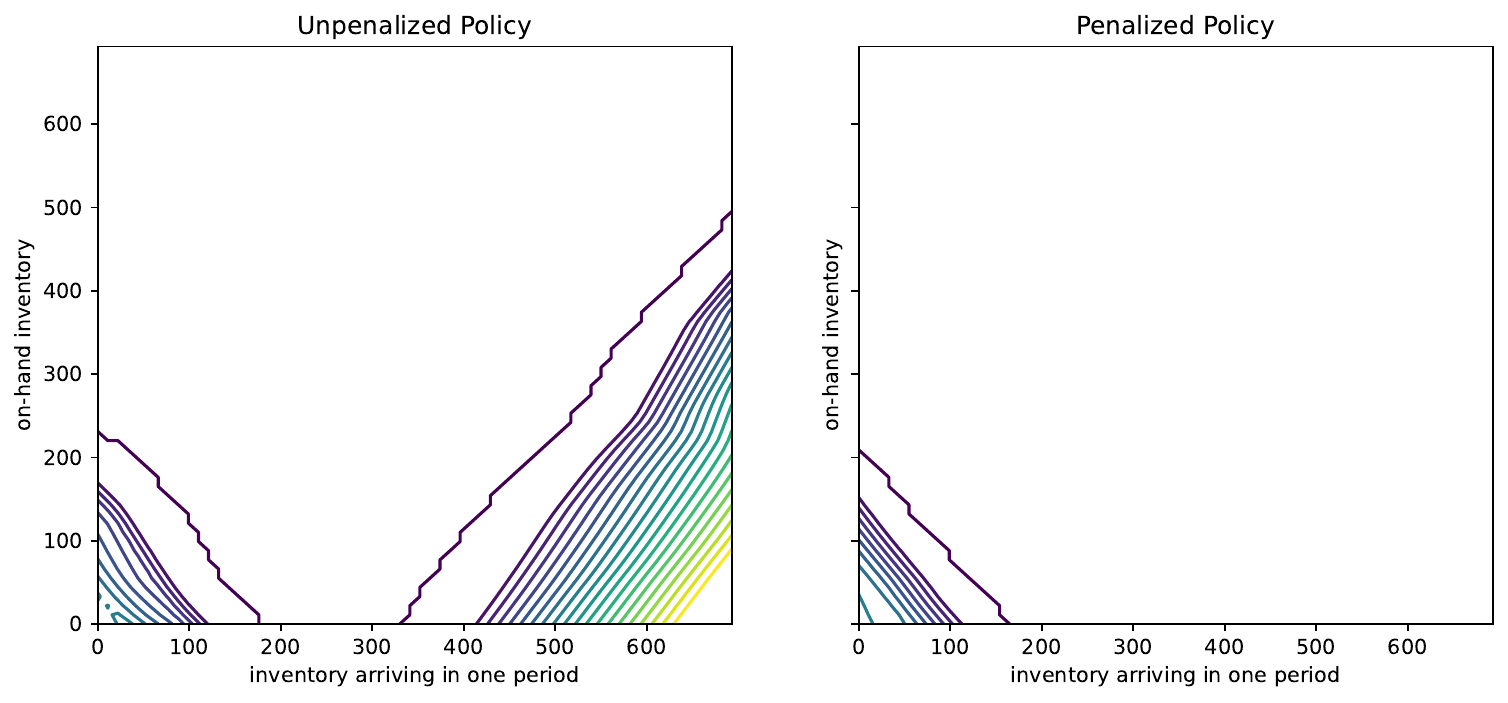}
\caption{Example contour plots of the unpenalized and penalized ($\lambda=1e6$) DRL policies trained on 1k products, for a given product in the Favorita data, as a function of the endogenous state $\mathbf{y}=(y^0,y^1,0,0,0)$ at $t=17$.}\label{fig:favorita_policy}
\end{figure}

To understand this behavior, it can be useful to look at individual product policies. Figure~\ref{fig:favorita_policy} shows a slice of the buying policy for a given product at a given time during the test rollout for both the unpenalized and penalized policies, in the form of a contour plot describing the order quantity as a function of only the on-hand inventory, and inventory arriving in one period, setting the other inventory-related state variables to 0. We observe that the policies learned in what it is the usual range of operation (say 0-300 in the context of this product) are very similar. However, beyond that region, the decisions generated by the unpenalized policy are essentially ``noise'' since it has never encountered that region and never had to worry about extrapolating there, leading to spurious behaviors where in spite of having a very high inventory position, the agent keeps buying, entering a vicious cycle of continued purchasing. The penalized policy on the other hand, by nudging the polciy to be monotonous, does not present this type of insidious effect and is thus more robust to extrapolation errors.

A more quantitative assessment can be obtained by looking at the mean and p50 (50th quantile) of the average expected reward across the 200 policies learned policies on the test set, for each of the training sample set size. The mean informs on the robustness of the learned policies, since a single poor out-of-sample performance from one of the policies would drag the metric down, and the p50 value on the other hand is more representative of the ``typical'' performance one would expect by learning a policy with that large a training set. Tables~\ref{tab:favorita_results_mean} and \ref{tab:favorita_results_p50} show these results.

\begin{table}[htbp]
\centering
\caption{Mean average expected reward of 200 learned policies on the test set for different training set sample sizes.}\label{tab:favorita_results_mean}
\begin{tabular}{llllll}
\toprule
 & \multicolumn{5}{c}{training set size $N$}\\
policy & 100 & 500 & 1k & 5k & 10k\\
\midrule
unpenalized & $-4.37\times 10^9$ & 1,229,83 & -734,986.85 & 1,606.76 & 2,232.40\\
penalized & 2,119.28 & 2,180.35 & 2,198.36 & 2,229.50 & 2,240.17\\
\bottomrule
\end{tabular}
\end{table}

\begin{table}[htbp]
\centering
\caption{P50 average expected reward of 200 learned policies on the test set for different training set sample sizes.}\label{tab:favorita_results_p50}
\begin{tabular}{llllll}
\toprule
 & \multicolumn{5}{c}{training set size $N$}\\
policy & 100 & 500 & 1k & 5k & 10k\\
\midrule
unpenalized &  2,119.28 	&	 2,180.35 	&	 2,198.36 	&	 2,229.50 	&	 2,240.17 	\\
penalized & 2,133.85 	&	 2,183.64 	&	 2,199.62 	&	 2,229.74 	&	 2,240.19 	\\
\midrule
improvement & 0.69\%	&	0.15\%	&	0.06\%	&	0.01\%	&	0.00\%	\\
\bottomrule
\end{tabular}
\end{table}

What we observe is that the regularization procedure yields much more robust policies than their unregularized counterpart. Looking at the mean average expected reward, we see that unregularized policies are sensitive, especially at smaller training set sizes, and prone to poor extrapolation, leading to possibly catastrophic performance out of sample. The regularized policies on the other hand, even for very small training set sizes, help apply the kind of structural properties that prevent the policies from poorly extrapolating out-of-sample. What is also notable is that there isn't necessarily a trade-off in the added robustness in that the regularization doesn't seem to harm the overall performance of the learned policies. In fact, for lower sample set sizes, the typical performance of regularized policies, as measured by the p50, is better than that of unpenalized policies, and that gap closes as the training set size increases, but doesn't tilt towards the unpenalized policies, at least up to $N$=10k in this example.

\section{Non-Stationary Setting Illustration}\label{sec:nonstationary}

\subsection{Overview}

Our DRL approach is one that doesn't make any assumption about the demand distributions, whether in terms of stationarity, temporal correlation, or marginal distribution. Indeed, the neural architecture we presented in Section~\ref{sec:policy_architecture} is inspired by the MQ-CNN architecture used for demand forecasting \cite{wen2017multihorizon,eisenach2020mqtransformer}, and is thus well-suited for a wide variety of applications. By providing, as part of the inputs, the same inputs that would be used in a pure forecasting problem, the DRL agent implicitly constructs a forecast picture, but one that is blended with additional information such as economics of the products and inventory picture internally, to produce an action that directly aims to maximize the long-term reward, rather than producing intermediate values that would be post-processed and then used as inputs in a downstream system.

Nonetheless, assumptions about the demand distributions, among other things, are often necessary to derive structural results about optimal policies in traditional inventory management  problems. As a result, the examples we considered in Section~\ref{sec:experiments} did not fully capture the benefits of DRL by restricting it to simple stationary scenarios.  In this section, we consider the same realistic demand data as in Section~\ref{sec:favorita}, originating from the Corporacion Favorita, and use two different approaches:
\begin{enumerate}
\item A \emph{Predict-then-Optimize} (PtO) approach similar to what would traditionally be done in practice, where a forecasting system is first trained to generate forecasts consumed by a downstream buying agent,
\item An end-to-end DRL approach that directly produces a buying quantity.
\end{enumerate}

\subsection{Data}

We generate data in the same way we did in Section~\ref{sec:favorita}, using 33k demand traces from the Corporacion Favorita data, and generating economic parameters as was described in Section~\ref{sec:overview}, and in particular in Table~\ref{tab:distributions}. We use the first 32+52=84 weeks as initialization and training data. The initialization is so that we have $H=32$ periods of prior observations for the time series  variables. We then evaluate the PtO and DRL policies first on the following 85 weeks, so as to stop right before the earthquake, but also on all remaining 156 weeks in the data, going through the earthquake, to observe how the DRL agent behaves.

\subsection{DRL Agent}

\subsubsection{MDP Formulation}

We consider a simple scenario still, with null lead times but lost sales, similar to the one considered in Section~\ref{sec:lost_sales} in the stationary case. Because the demand is non-stationary and Christmas seems to have an outsize impact on sales, we additionally introduce a time-series feature that indicates the number of weeks left until the next Christmas. As a result, the MDP formulation of the problem takes the following form:

\paragraph{State} The endogenous state is simply given by the inventory level $y_t$, while the exogenous time series variables $\mathbf{x}_t$ consists of the previous $H$ demand realizations and distances to Christmas:
\begin{align*}
\mathbf{x}_t=\begin{bmatrix}
d_{t-H} & \ldots & d_{t-1} \\
\delta_{H-1} & \ldots & \delta_{t-1}
\end{bmatrix},
\end{align*}
where $\delta_{s}$ is the number of weeks left before the next Christmas at time $s$. The exogenous static vector contains the economic parameters of a product, $\mathbf{s}=(p,c,h,b)$, where $p$ is the price, $c$ the purchasing cost, $h$ the holding cost, and $b$ the penalty for lost sale.

\paragraph{Action} The action $a_t=q_t$ at time $t$ is given by the number of units to purchase at time $t$, which will also arrive at the beginning of the period.

\paragraph{Transition Function} The transition function is simply given by $y_{t+1} = \max(y_t + q_t - d_t , 0)$, meaning that we sell up to $d_t$ units out of the $y_t+q_t$ units that are available for sale, and any unit left over is carried on to the next period.

\paragraph{Reward Function} In each period, we receive $p$ for each sold unit, and incur unit costs of $c$ for any purchased unit, $b$ for any missed sale, and $h$ for any leftover unit, yielding:
\begin{align*}
R_t &= p \min(d_t, y_t+q_t) - c q_t - b \max(d_t - y_t - q_t, 0) - h \max(y_t + q_t - d_t, 0).
\end{align*}

\subsubsection{Training}

\begin{figure}[htbp]
\centering
\includegraphics[scale=0.4]{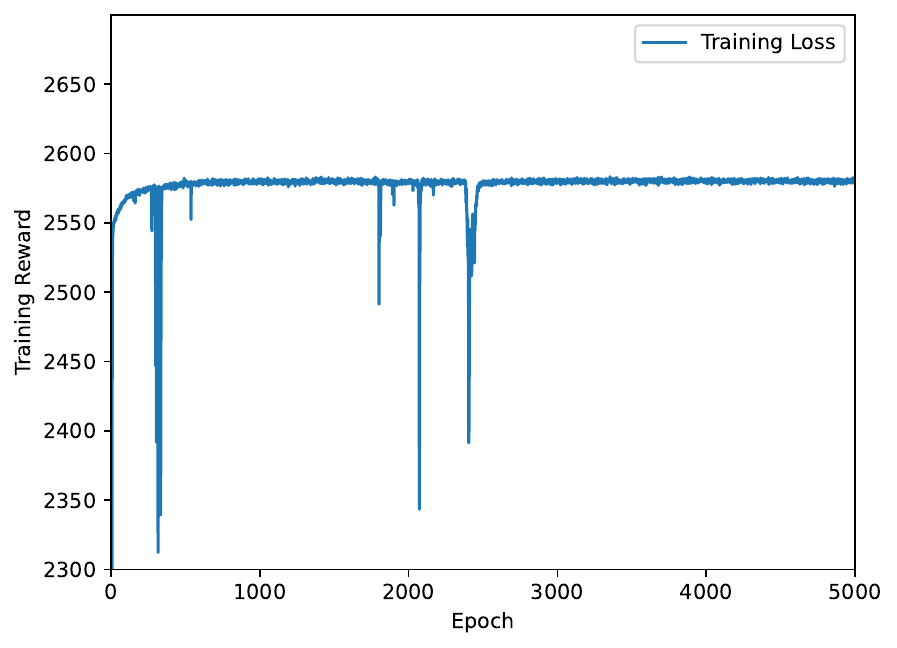}
\caption{Training reward for the training of the DRL agent.}\label{fig:favorita_drl_training}
\end{figure}

The DRL agent implements the same architecture we used for the experiments in Sections~\ref{sec:experiments} and \ref{sec:favorita}, and represented in Figure~\ref{fig:policy}. We also use the same hyperparameters, as presented in Table~\ref{tab:hyperparams}, except for the batch size, which is now 33k instead of 40k, and the number of epochs, now 5k. The training reward is presented in Figure~\ref{fig:favorita_drl_training}.

\subsection{Predict-then-Optimize Approach}

While considering the same MDP dynamics as the DRL agent,  the PtO approach consists of three steps in each period $t$:
\begin{enumerate}
\item Generate a demand forecast $D_t$,
\item Find the demand quantile corresponding to the product's critical ratio to yield an order-up-to level $s_t=F_{D_t}^{-1}\left(\frac{c_u}{c_u+c_o}\right)$ (where $c_u$ and $c_o$ are the underage and overage costs, respectively, and $F_D^{-1}$ is the inverse of the cumulative distribution function of $D$),
\item Generate an order quantity as $q_t=\max(s_t - y_t, 0)$.
\end{enumerate}

This approach implements a myopic policy, which we already mentioned in Section~\ref{sec:lost_sales} are is in the stationary setting. It is actually optimal for a wider range of demand processes, in particular when the critical order-up-to levels are nondecreasing over time \cite{veinott1965optimal}, for which a sufficient condition is that the  demand distributions be stochastically decreasing \cite{veinott1965optimal_single}. \cite{johnson1975optimality} further extend the result to the case where the demand is generated by some ARMA process.

\subsubsection{Forecasting Agent}

\begin{figure}[htbp]
\centering
\includegraphics[scale=0.5]{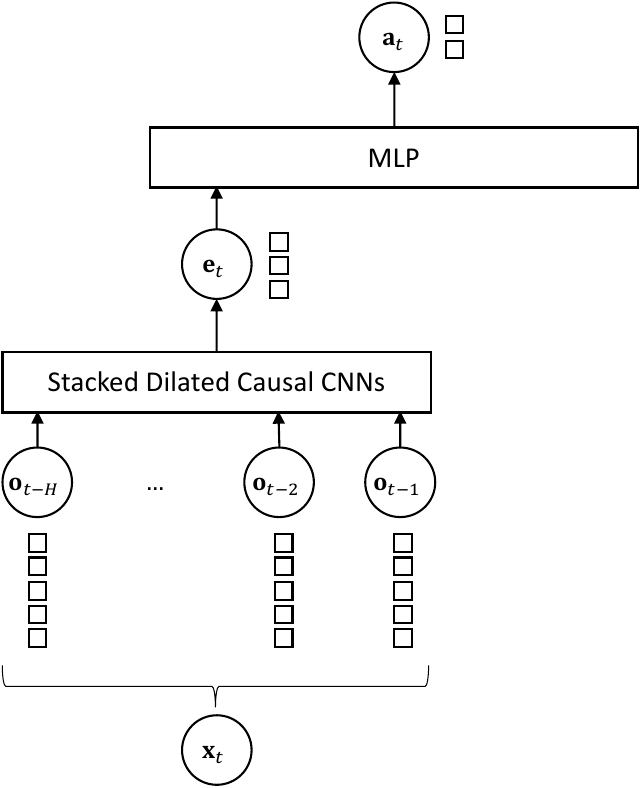}
\caption{Schematic representation of the forecasting neural network architecture.}\label{fig:mqcnn}
\end{figure}

To generate stochastic demand forecasts, we first train a simple MQ-CNN architecture that consumes the same amount of demand information as the DRL agent does. This architecture is in a way the restriction of the DRL architecture to just this information, as shown in Figure~\ref{fig:mqcnn}. The network is fed the past $H$ (uncensored) demand realizations, as well as distances to the next Christmas.

The ``actions'' of this forecasting agent are quantiles for the distribution of the demand in the next period, and it is trained by minimizing the empirical total quantile loss, or equivalently maximizing its opposite:
\begin{align*}
J_T(\bm\theta) = \mathbb{E}\left[\sum_{t=0}^{T-1} \sum_{q\in Q} - L_q(D_t, \hat{d}_t^q)\right],
\end{align*}
where $Q$ is the set of considered quantiles, $\hat{d}_t^q$ is the predicted quantile at level $q$, and $L_q(d, \hat{d}^q)$ is the quantile loss defined as:
\begin{align*}
L_q(d, \hat{d}^q)=q (d- \hat{d}^q)^+ + (1-q) (\hat{d}^q - d)^+.
\end{align*}

\subsubsection{Forecasting Agent Training}

\begin{figure}[htbp]
\centering
\includegraphics[scale=0.4]{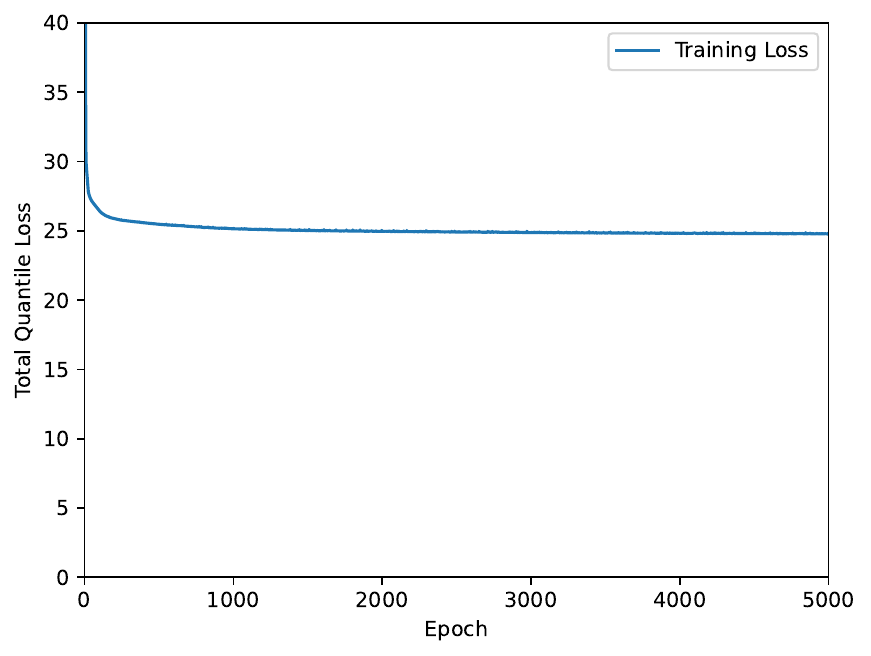}
\caption{Training loss for the training of the forecasting agent.}\label{fig:forecasting_training}
\end{figure}

We train this forecasting system on the same data as the DRL agent, i.e. on 32+52 weeks worth of data. The neural architecture is also the same, but limited to just ingesting the time series data (past $H$ demand realizations and distances to the next Christmas), but otherwise using the same stacked dilated causal CNNs, and same number of MLP layers (2) with the same number of neurons, using the same activation functions (ELU). We set $Q=[0.01,0.1,0.25,0.5,0.75,0.9,0.99]$, so that it produces $|Q|$ ``actions''. We further use the same hyperparmeters as for the DRL agent and presented in Table~\ref{tab:hyperparams}, except for the sample size, which is now just the 33k products in the dataset, and 5k epochs. Figure~\ref{fig:forecasting_training} shows the average (per-period) total quantile loss.

\begin{figure}[htbp]
\centering
\includegraphics[scale=0.35]{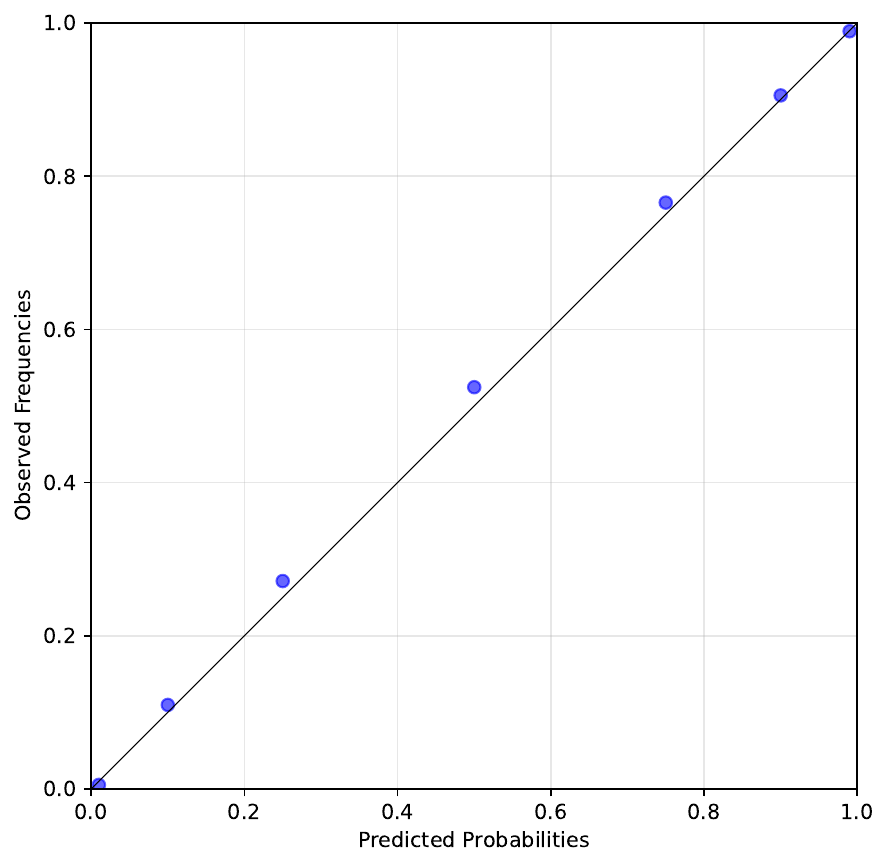}
\caption{Calibration plot of the forecasting agent in the evaluation period.}\label{fig:forecasting_calibration}
\end{figure}

To check that the trained forecasting agent is reasonable, we plot in Figure~\ref{fig:forecasting_calibration} the calibration plots obtained on the evaluation period, where we observe a decent calibration considering the simplicity of the model.

\begin{figure}[htbp]
\centering
\includegraphics[scale=0.4]{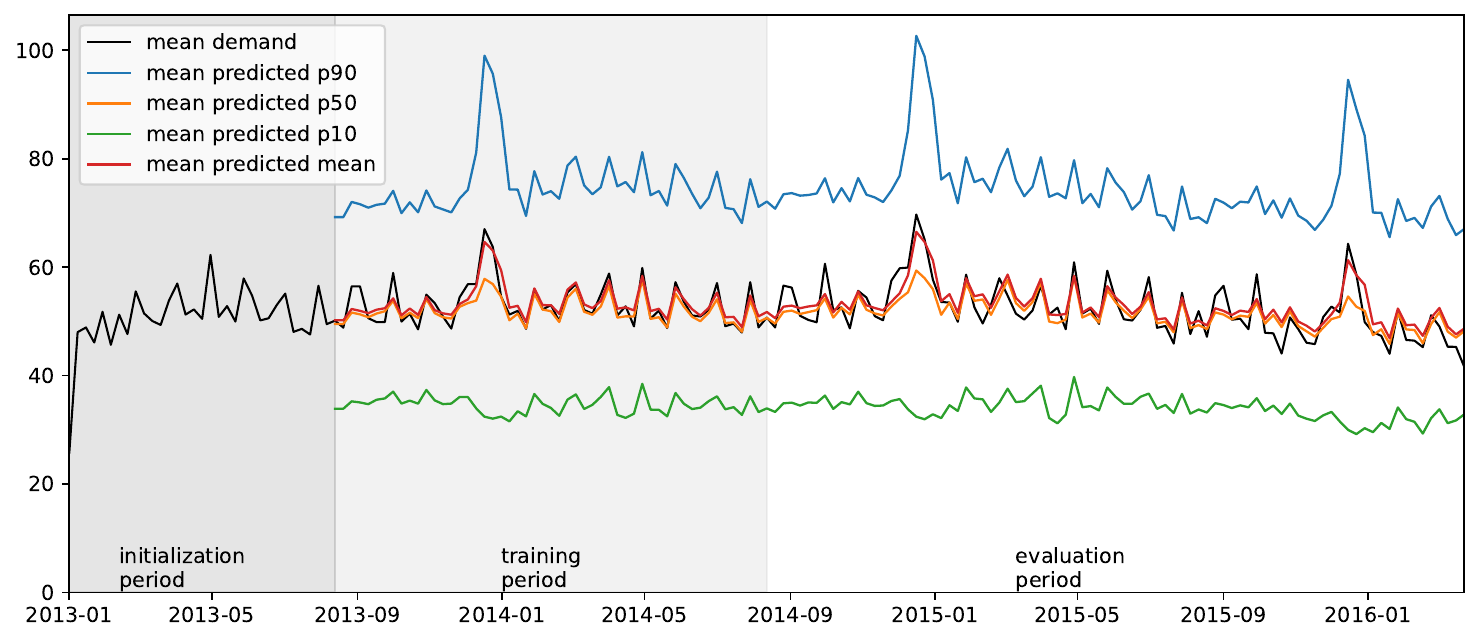}
\caption{Mean forecasted p10, p50, and p90, and  mean forecasted mean demand generated by the trained forecasting agent.}\label{fig:forecasting_agent}
\end{figure}

\begin{figure}[htbp]
\centering
\includegraphics[scale=0.4]{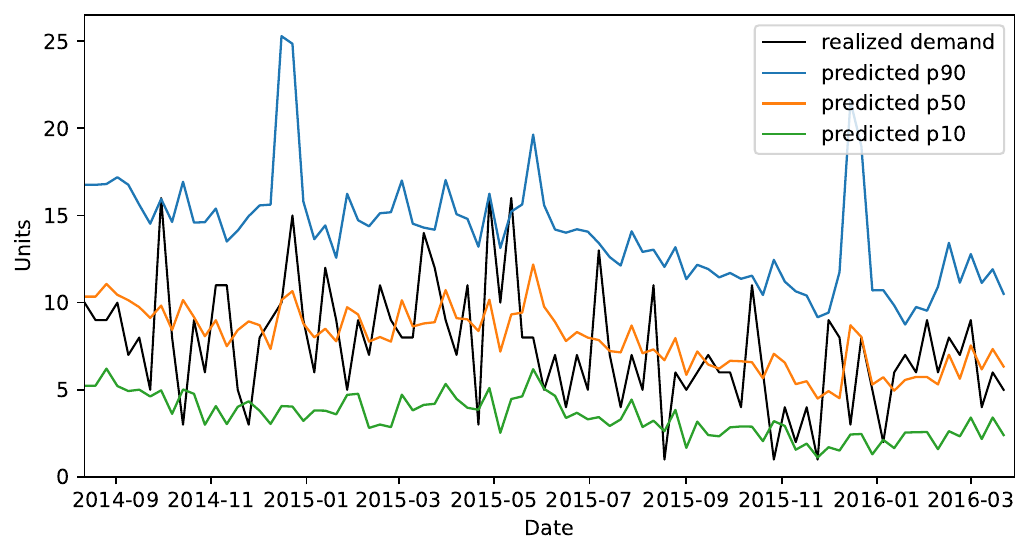}
\includegraphics[scale=0.4]{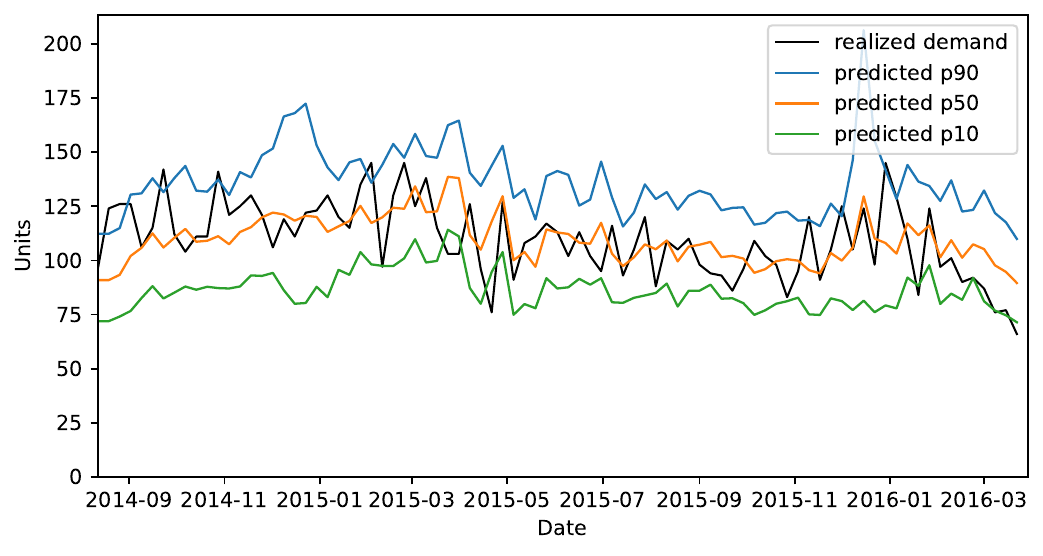}
\caption{Example of forecasted p10, p50, and p90, and  realized demand for a couple of products.}\label{fig:forecasting_examples}
\end{figure}

As a means of illustration, we also plot in Figure~\ref{fig:forecasting_agent} the mean product demand over the first 169 weeks, as well as the mean forecasted p10, p50, and p90, and the mean forecasted mean demand obtained from the forecasted quantiles (note that the  mean forecasted p10, p50, and p90 don't have any concrete meaning and don't represent quantiles for the total demand). We observe that the forecasting agent is able to pick up the roughly 4 week periodicity, as well as the impact of Christmas on demand. Figure~\ref{fig:forecasting_examples} also shows the forecasted p10, p50, and p90, along with the realized demands, for a couple of products.

\subsubsection{Order-up-to Level and Buying Quantity Computation}

We follow a common industry practice to predict a set of quantiles for the distribution of the demand, and interpolate in between to get any quantile. As a result, in any period $t$, and for a given product, we compute an order-up-to level $s_t$ as:
\begin{align*}
s_t =F_{D_t}^{-1}\left(\frac{c_u}{c_u+c_o}\right),
\end{align*}
where the underage and overage costs are given by:
\begin{align*}
c_u &= p-c+b, & c_o&=h,
\end{align*}
and where we recall that $p$ is the selling price, $c$ the purchasing cost, $b$ the penalty for lost sale, and $h$ the per-unit holding cost.

The buying quantity is then directly computed as:
\begin{align*}
q_t &= \max(s_t - y_t, 0).
\end{align*}

\subsection{Results}

After training both the DRL and Forecasting agents for 5k epochs, we rollout the corresponding DRL and PtO policies on the evaluation data corresponding to the time periods following the training data. We consider the average per-period reward first in the 85 weeks following the training period, ending right before the earthquake that hit Ecuador, and then on the whole 156 remaining weeks that do include the earthquake. The results are presented in Table~\ref{tab:nonstationary_results}, where we see that the DRL agent outperforms the PtO one, and that the improvement increases on the larger evaluation period.

\begin{table}[htbp]
\caption{Average per-period reward of the DRL and PtO agents on the evaluation sets using the Favorita data.}\label{tab:nonstationary_results}
\centering
\begin{tabular}{lrr}
\toprule
 & \multicolumn{2}{c}{Period}\\
 Agent & 85 weeks & 156 weeks\\
 \midrule
 PtO agent & 2,412.70 & 2,281.11\\
 DRL agent& 2,427.75 & 2,300.59 \\
 \midrule
 Improvement(\%) & 0.62\% & 0.85\% \\
 \bottomrule
\end{tabular}
\end{table}

\begin{figure}[htbp]
\centering
\includegraphics[scale=0.4]{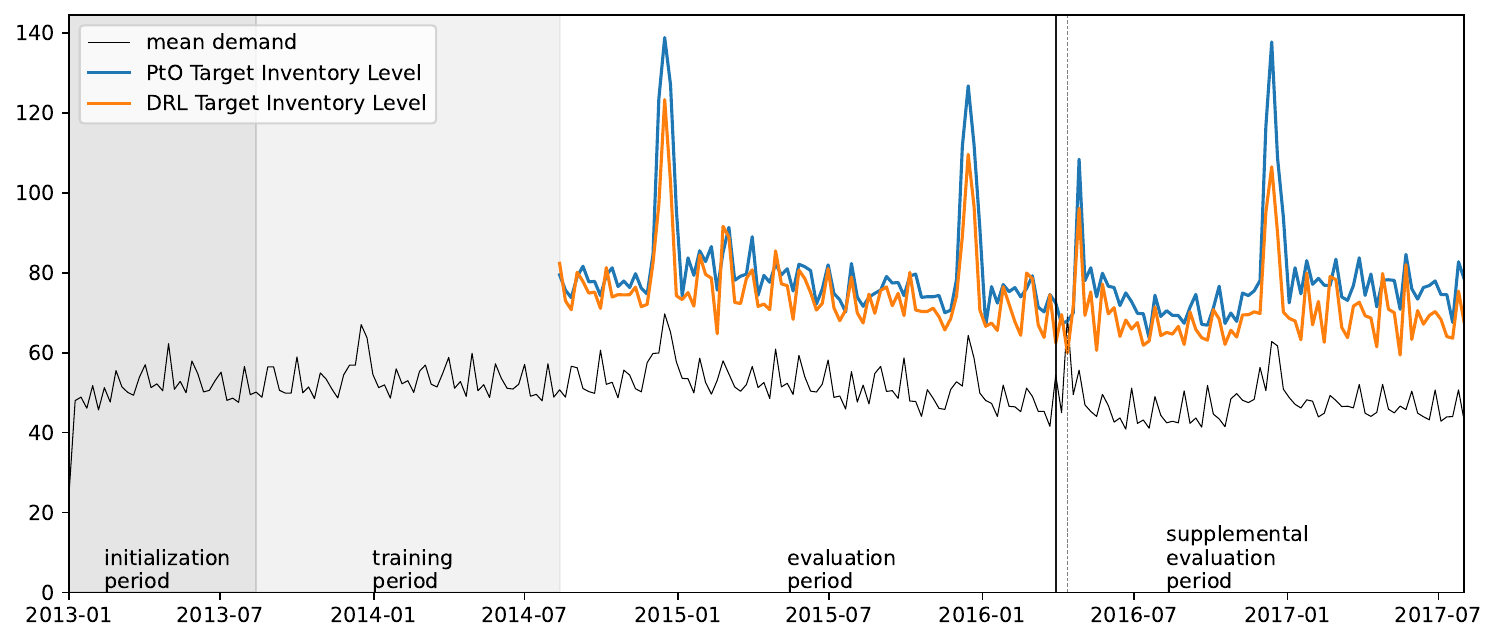}
\caption{Mean target inventory levels for the PtO and DRL agents over time.}\label{fig:favorita_tips}
\end{figure}

Figure~\ref{fig:favorita_tips} shows the mean target inventory levels in each period for both agents, which we recall are the sum of the on-hand inventory and purchases quantities. 

\begin{figure}[htbp]
\centering
\includegraphics[scale=0.4]{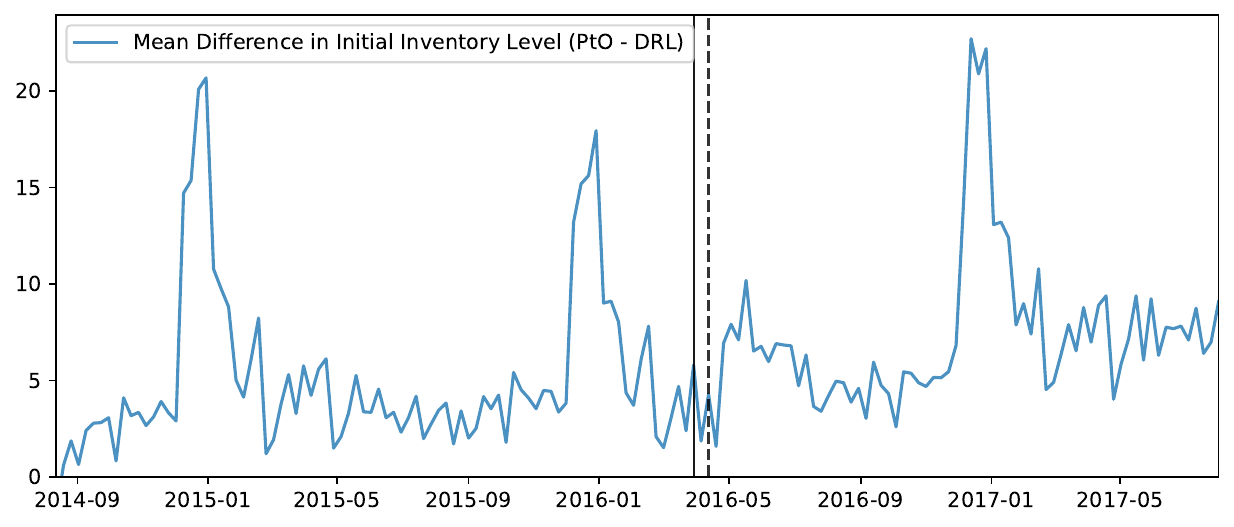}
\caption{Mean difference in initial inventory levels for the PtO and DRL agents over time.}\label{fig:diff}
\end{figure}

We observe that the DRL agent consistently operates at a lower target level than the PtO agent, which is reflected in Figure~\ref{fig:diff} where we plot the mean (per-product) difference in inventory levels between the two agents. The inventory levels held by the DRL agent are consistently lower than those held by the PtO agent, and the gap is especially pronounced around the Christmas peak, where the myopic nature of the PtO policy ceases to be close to optimal. The DRL agent on the other hand takes into account the longer term impact of potentially holding on to larger stocks of inventory and incurring higher holding costs.

To understand whether and how the differences in policy behavior affect the customer experience, we consider the Type 1 and Type 2 service levels, both product and demand weighted. We recall that the Type 1 service level $\alpha$ measures the probability of not going out-of-stock ($\alpha=\mathbb{P}[D_t \leq y_t]$), while the Type 2 service level $\beta$ is a measure of fill rate, i.e. the fraction of demand that is satisfied ($\beta = 1 - \frac{\mathbb{E}[(D_t - y_t)^+]}{\mathbb{E}[D_t]}$). We compute the averages of these metrics for both agents using two different weighing schemes, one simply averaging product level values, and the other weighing by the demand in each period. The results are presented in Table~\ref{tab:favorita_service_levels}. The table shows a consistent behavior in that the PtO agent shows better results on product weighted metrics, while the DRL agent outperforms it on the demand weighted versions of these metrics.

\begin{table}[htbp]
\centering
\caption{Average Type 1 ($\alpha$) and Type 2 ($\beta$) service levels, both product and demand (subscript $d$) weighted for the PtO and DRL agents.}\label{tab:favorita_service_levels}
\begin{tabular}{lcccccccc}
\toprule
& \multicolumn{8}{c}{Period}\\
\cmidrule(l){2-9}
& \multicolumn{4}{c}{85 weeks} & \multicolumn{4}{c}{156 weeks} \\
\cmidrule(l){2-5} \cmidrule(l){6-9}
Agent & $\alpha$ & $\alpha_d$ & $\beta$ & $\beta_d$ & $\alpha$ & $\alpha_d$ & $\beta$ & $\beta_d$\\
\midrule
PtO Agent & \textbf{85.93\%} & 79.31\% &\textbf{ 96.92\%} & 95.44\% & \textbf{85.70\%} & 77.88\% & \textbf{96.71\%} & 94.40\%\\
DRL Agent & 85.65\% & \textbf{79.60\%} & 96.91\% & \textbf{95.70\%} & 85.36\% & \textbf{78.11\%} & 96.68\% & \textbf{94.65\%}\\
\bottomrule
\end{tabular}
\end{table}

\section{Conclusion}

This paper makes several contributions to the literature on reinforcement learning applications in inventory management. First, we demonstrate that a generic DRL implementation using DirectBackprop can effectively handle diverse inventory management scenarios while requiring minimal parameter tuning. Our approach performs competitively against, or outperforms, established benchmarks across multiple classical settings including multi-period systems with lost sales (with and without lead times), perishable inventory management, dual sourcing, and joint inventory procurement and removal.

Second, we show that the DRL approach naturally learns many of the structural properties of optimal policies that have been analytically derived through traditional operations research methods. This finding helps bridge the gap between data-driven learning and theoretical insights. The learned policies exhibit expected behaviors like monotonicity and appropriate sensitivity to state variables, even without explicitly encoding these properties.

Third, we introduce the Structure-Informed Policy Network technique that explicitly incorporates analytically-derived characteristics of optimal policies into the learning process. We show on some realistic demand data how this approach can help with extrapolation and provide robustness on out-of-sample data.

From a practical perspective, our work shows that DRL can be effectively implemented in realistic settings where only historical information is available, avoiding unrealistic assumptions about demand distributions or access to distribution parameters. The approach learns policies across products rather than at an instance level, making it more applicable to real-world implementations.

Several directions for future research emerge from this work. First, extending the Structure-Informed Policy Network approach to incorporate additional types of structural properties could further improve performance and robustness. Second, investigating how the approach scales to larger networks with multiple echelons or more complex constraints would be valuable. Finally, developing theoretical guarantees for the convergence and optimality of structure-informed policies represents an important open question.

In conclusion, this paper demonstrates that combining deep reinforcement learning with structural insights from classical inventory theory yields practical and robust solutions for inventory management problems.

\bibliographystyle{alpha}
\bibliography{bib}

\appendix
\section{DirectBackprop Algorithm (from \cite{Madeka2021})}\label{app:dbp}

\begin{algorithm}
\caption{Direct Backpropagation for Inventory Control}
\begin{algorithmic}[1]
\Require $\mathcal{D}$ (a set of products), batch size $M$, initial policy parameters $\theta_0$
\State $b \gets 1$
\While{not converged}
    \State Sample mini-batch of products $\mathcal{D}_M$
    \State $J_b \gets 0$
    \For{$i \in \mathcal{D}_M$}
        \State $J_i \gets 0$, initialize $\mathbf{z}^i_0=(\mathbf{x}_0^i,\mathbf{s}^i,\mathbf{y}_0^i)$
        \For{$t = 0, \ldots, T$}
            \State $\mathbf{a}_t^i = \pi_{\theta_b,t}(\mathbf{z}^i_t)$
            \State $\mathbf{z}_{t+1}^i \gets f(\mathbf{z}_t^i,\mathbf{o}_t^i,\mathbf{a_t^i})$
            \State $J_i \gets J_i + \gamma^t R(\theta_b;\mathbf{z}_t)$
        \EndFor
        \State $J_b \gets J_b + J_i$
    \EndFor
    \State $\theta_b \gets \theta_{b-1} + \alpha \nabla_\theta J_b$
    \State $b \gets b + 1$
\EndWhile
\end{algorithmic}
\end{algorithm}

\end{document}